\documentclass[lettersize,journal]{IEEEtran}
\usepackage{physics}
\usepackage{graphicx}
\usepackage{float}
\usepackage{amsmath}
\usepackage{subfigure}
\usepackage{amsmath}
\usepackage{longtable}
\usepackage{booktabs}
\usepackage{float}
\usepackage{threeparttable}
\usepackage{xcolor}
\usepackage{diagbox}
\usepackage{cite}
\usepackage{hyperref}
\usepackage{pifont}
\usepackage{multirow}
\usepackage{xcolor}
\usepackage{colortbl} 
\usepackage{booktabs} 
\usepackage{array} 

\newcommand{\comment}[1]{}
\begin{document}

\title{Subjective and Objective Quality Assessment of Banding Artifacts on Compressed Videos}

\author{Qi~Zheng,
        Li-Heng~Chen,
        Chenlong He, 
        Neil~Berkbeck,
        Yilin~Wang,
        Balu~Adsumilli, \\Alan~C.~Bovik,~\IEEEmembership{Life Fellow,~IEEE},
        Yibo~Fan$^\star$,
        Zhengzhong~Tu
\thanks{This work was supported in part by the China NSF under Grant 62427801, in part by the National Key R\&D Program of China (2023YFB4502802), in part by the China NSF under Grant 62031009, in part by Fudan-ZTE joint lab, in part by Alibaba Research Fellow (ARF) Program.}
\thanks{Qi Zheng, Chenlong He, and Yibo Fan are with Fudan University, Shanghai 200000, China (e-mail: qzheng21@m.fudan.edu.cn; clhe22@m.fudan.edu.cn; fanyibo@fudan.edu.cn).}
\thanks{Li-Heng Chen is with Video Algorithms Team, Netflix, Los Gatos, CA, 95032, USA. (email: lhchen@utexas.edu)}
\thanks{N. Birkbeck, Y. Wang, and B. Adsumilli are with YouTube Media Algorithms Team, Google LLC, Mountain View, CA, 94043, USA. (email: birkbeck@google.com, yilin@google.com, 
badsumilli@google.com)}
\thanks{Alan C. Bovik are with the Laboratory for Image and Video Engineering (LIVE), Department of Electrical and Computer Engineering, The University of Texas at Austin, Austin, TX 78712 USA (e-mail: bovik@utexas.edu).}
\thanks{Zhengzhong Tu is with the Department of Computer Science and Engineering, Texas A\&M University, College Station, TX 77840
(e-mail: tzz@tamu.edu). This work was done prior to the employment of Zhengzhong Tu by Texas A\&M University, and he was not
supported by any grant.}
\thanks{This paper has supplementary downloadable material available at http://ieeexplore.ieee.org., provided by the author. The material includes supplementary experimental results. Contact qzheng21@m.fudan.edu.cn for further questions about this work.}
\thanks{$^\star$Corresponding author.}}

\markboth{IEEE TRANSACTIONS ON IMAGE PROCESSING, VOL. 34, 2025}%
{Shell \MakeLowercase{\textit{et al.}}: A Sample Article Using IEEEtran.cls for IEEE Journals}


\maketitle
\begin{abstract}
\label{abstract}
Although there have been notable advancements in video compression technologies in recent years, banding artifacts remain a serious issue affecting the quality of compressed videos, particularly on smooth regions of high-definition videos. Noticeable banding artifacts can severely impact the perceptual quality of videos viewed on a high-end HDTV or high-resolution screen. Hence, there is a pressing need for a systematic investigation of the banding video quality assessment problem for advanced video codecs. Given that the existing publicly available datasets for studying banding artifacts are limited to still picture data only, which cannot account for temporal banding dynamics, we have created a first-of-a-kind open video dataset, dubbed LIVE-YT-Banding, which consists of 160 videos generated by four different compression parameters using the AV1 video codec. A total of 7,200 subjective opinions are collected from a cohort of 45 human subjects. To demonstrate the value of this new resources, we tested and compared a variety of models that detect banding occurrences, and measure their impact on perceived quality.
Among these, we introduce an effective and efficient new no-reference (NR) video quality evaluator which we call CBAND. CBAND leverages the properties of the learned statistics of natural images expressed in the embeddings of deep neural networks. Our experimental results show that the perceptual banding prediction performance of CBAND significantly exceeds that of previous state-of-the-art models, and is also orders of magnitude faster. Moreover, CBAND can be employed as a differentiable loss function to optimize video debanding models. The LIVE-YT-Banding database, code, and pre-trained model are all publically available at \url{https://github.com/uniqzheng/CBAND}.

\end{abstract}
\section{Introduction}
\label{Introduction}
\IEEEPARstart{I}{n} recent years, the rapid evolution of streaming media technologies and platforms has led to video content dominating Internet traffic.
This shift, particularly prominent since the explosion of user-generated content, has made video an integral part of billions of people's daily lives.
For video service providers, a significant and persistent challenge is to enhance the efficiency of cloud-based video transcoding techniques, while ensuring satisfying quality of experience (QoE) of customers being served over varying network bandwidths~\cite{liu2025frequency}.
Video compression or transcoding often leads to annoying distortions that may seriously impair perceptual quality~\cite{8653951,9043584,9035388,zheng2024unicorn,8772137,zuo20254kagent}, including issues such as block effects, banding artifacts, ringing, blurring, mosquito effects, ghosting, and jerkiness, among others.
Banding artifacts continue to impact the perceptual quality of originally high-quality, high-bitrate videos, especially when displayed on large high-definition displays~\cite{tu2020bband}.
%
To understand human perception of, and reactions to banding, a comprehensive study of the perception of banding artifacts arising in compressed videos is needed.
Such a data resource holds the potential to serve as a foundation towards developing perceptually optimal banding detection/prediction methods as well as post-processing debanding procedures~\cite{tu2020adaptive,sole2023debanding,peng2021bilateral,zhou2022deep}. Success in this direction can lead to enhanced quality and efficiency of streaming video coding and transcoding processes, thereby augmenting the performance of multimedia applications.

Banding, also known as false contouring, arises from the quantization operation (ubiquitous in modern video encoders)~\cite{yuen1998survey}. Banding often occurs on smoothly textured areas of frames containing gradual transitions of color and/or luminance. Banding artifacts can be attributed to excessively coarse quantization of DC coefficients or low frequency AC coefficients in compressed DCT-domain encodes. Prevalent video compression standards featuring block-based transform coding strategies, including H.264/AVC~\cite{wiegand2003overview}, H.265/HEVC~\cite{sullivan2012overview}, VP9~\cite{mukherjee2013latest}, AVS3~\cite{zhang2019recent}, AV1~\cite{han2021technical}, and even the most advanced H.266/VVC~\cite{bross2021overview}, AV2~\cite{av2}, are all susceptible to noticeable banding artifacts~\cite{sole2023debanding}. Since a goal of streaming video platforms is to deliver perceptually optimized, artifact-free video while still compressing the data as much as possible, the development of accurate and efficient perceptual banding prediction models is greatly desired.

As mentioned, there is a dearth of perceptual video quality databases focused on banding artifacts.
Subjective video quality databases are the basic tools for the development, calibration, and benchmarking of perceptual video quality models~\cite{min2020study}. However, among the few existing banding databases, most are either not publically available~\cite{wang2016perceptual,tandon2021cambi} or only provide binary banding labels(i.e. banding is present or absent)~\cite{kapoor2021capturing}, which are too coarse to model the perception of suprathreshold banding artifacts and their impact on predicted quality. To the best of our knowledge, there does not yet exist an open-source video quality database dedicated to studying the perceptual aspects of banding artifacts induced by video compression.
This paucity of scientific data poses a great barrier to the development of perceptual banding measurement methods. Since these kinds of tools are needed to be able to conduct banding-oriented optimization in high-end streaming video workflows, we have been strongly motivated to create such a perceptual data resource.


The development of objective video quality prediction models and algorithms generally involves training learning machines to map "distortion-aware" video features (in the form of neuroscience-based statistics and/or deep embeddings) to human perceptual judgments of visual quality. While there has been interest in the topic of banding artifacts for some time~\cite{wang2016perceptual, chen2023fs}, most existing algorithms have been based on heuristic handcrafted features~\cite{bhagavathy2009multiscale, baugh2014advanced,wang2016perceptual,5712203, 7032274}, making them susceptible to errors or misclassification. Recently, advances in deep neural networks (DNN) ~\cite{simonyan2014very,he2016deep, huang2017densely,dosovitskiy2020image,tu2022maxvit,tu2022maxim} have made possible highly effective approaches to the general problem of video quality assessment. These most often deploy end-to-end fine tuning of models pre-trained on high-level recognition tasks~\cite{zhang2018unreasonable,li2019quality,wu2022neighbourhood}. Two models have been developed that exploit DNN modules for the detection of banding artifacts: the DBI model proposed by Kapoor et al.~\cite{kapoor2021capturing} and the FS-BAND model advanced by Chen et al.~\cite{chen2023fs}. However, the considerable computational load and the large number of network parameters likely limit the deployment of these models in practical video delivery applications. 

Here we seek to make progress towards addressing these challenges in two ways.
Firstly, we built a large-scale open-source video quality database dedicated to the study of perceptual aspects of banding artifacts induced by video compression. This new data resource, which we call \textbf{LIVE-YT-Banding}, is to the best of our knowledge, the first of its kind.
We began by employing the open source video codec AV1~\cite{han2021technical}, to generate test video sequences, on which we conducted a controlled subjective study involving 45 volunteer subjects, yielding a total of 7.2K human judgments of video quality.
Second, we used this data resource to create an efficient and effective blind video banding quality evaluation engine dubbed \textbf{CBAND}. Our contributions can be summarized as follows:
\begin{itemize}
    \item {We built the first large-scale open-source perceptual video quality database dedicated to the study of the perceptual impacts of banding artifacts arising from video compression. It is called the \textbf{LIVE-YT-Banding} Database. It encompasses 160 video sequences, including 40 different reference videos, along with four versions of each processed by different levels of AV1 compression. A total of 7,200 opinion scores were collected from 45 volunteer subjects in a controlled laboratory environment.}
    \item {We designed \textbf{CBAND}, a lightweight yet highly effective blind video banding quality evaluator. CBAND conducts banding-aware feature extraction in the form of activation maps produced by the early stages of pre-trained CNNs. We cast these early/deep features against parametric neurostatistical models of distortion perception.}
    \item {We evaluate the efficacy of CBAND metric and other leading VQA models on the LIVE-YT-Banding Database. The experimental results show that CBAND significantly outperforms the prior state-of-the-art in terms of accurate banding quality prediction, with orders-of-magnitude faster inference speed.}
    \item {Additionally, we demonstrate the additional usefulness of CBAND as a differentiable optimization objective on the video frame debanding task. We demonstrate the effectiveness of this approach to perceptual video enhancement on practical transcoding scenarios.}
\end{itemize}

The rest of the paper is organized as follows. Section~\ref{sec:related_work} provides an overview of related work, including existing banding databases and objective video banding quality measurement algorithms. Section~\ref{sec:database_creation} explains the construction of the LIVE-YT-Banding database. Details regarding the protocol and execution of the subjective video banding quality study are presented in Section~\ref{sec:subjective_study}. Section~\ref{sec:objective_metric} proposes the new
no-reference video banding evaluator, called CBAND. Section~\ref{sec:experiment} gives experimental results and comparative analysis of video quality models on the new subjective database. Section~\ref{sec:conclusion} concludes the paper with final remarks.

\section{Related Work}
\label{sec:related_work}


\begin{table*}[!t]
\centering
\fontsize{6pt}{7pt}\selectfont
\setlength{\tabcolsep}{1.5pt}
\caption{Metadata describing existing picture and video banding quality databases.}
\vspace{-8pt}
\label{table:metadata_datasets}
    \begin{tabular}{lcccccc}
    \toprule
    \textbf{Metadata} & \textbf{Wang et al.~\cite{wang2016perceptual}} & \textbf{Tandon et al.~\cite{tandon2021cambi}}   & \textbf{Kapoor et al.~\cite{kapoor2021capturing}}   & \textbf{Xue et al.~\cite{xue2023large}}  & \textbf{BAND-2k~\cite{10438477}}  & \textbf{LIVE-YT-Banding}\\ \midrule
    Year  &   2016                                                           &    2021                                               &  2021 & 2023 & 2023 & 2024                                                \\  
    Source content &   YouTube                                                        & Netflix catalogue &       unknown &       PGC    &  PGC, UGC, CG      &  PGC, UGC, CG  \\ 
    Stimuli type & Video                                                             & Video & Image           &        Image        &  Image   &                                            Video                                                            \\ 
    Number of source contents & 7                                                           &  9 & 600         &        72          &  873  &                                              40                                                           \\ 
    Number of all stimuli & 21                                                             & 86 & 1,440          &       310         &  2,000 &                                               160                                                           \\ 
    Resolution & $1280\times720$                                                             & $3840\times2560$ & $1920\times1080$                           & $1920\times1080$ & $1920\times1080$ & $1920\times1080$                                                           \\ 
    Distortion type & VP9 compression                                                            & \begin{tabular}[c]{@{}c@{}}AV1 compression, resolution \\  and bit-depth downsampling\end{tabular} & bit-depth downsampling  & AVC compression & \begin{tabular}[c]{@{}c@{}}H.264, H.265, \\ VP9, bit-depth\end{tabular} &  AV1 compression                                                  \\
    Number of distortion levels & 3  & 9 & 6 & unknown & 3,3,3,6 & 4   \\
    Experiment & laboratory  & crowdsourcing & -   & laboratory &laboratory &  laboratory \\
    Rating scale & Continuous rating 0-100 & Continuous rating 0-100 & - & Discrete rating 1-5 & Continuous rating 0-100 &  Continuous rating 0-100 \\
    Number of subjects & 25  & 23 & - & 56 & 23 &   45   \\
    Number of ratings & \textgreater1,000  & - & -  & 2900 & 44,371 &   7,200   \\
    MOS & \textcolor{green}{\ding{52}}  & \textcolor{green}{\ding{52}} & \textcolor{red}{\ding{56}} & \textcolor{green}{\ding{52}} & \textcolor{green}{\ding{52}}   & \textcolor{green}{\ding{52}}   \\
    Open-sourced  &   \textcolor{red}{\ding{56}}                                                          &     \textcolor{red}{\ding{56}}                                               & \textcolor{green}{\ding{52}}  & \textcolor{red}{\ding{56}} & \textcolor{green}{\ding{52}}  & \textcolor{green}{\ding{52}}                                             \\ 
    \bottomrule
    \end{tabular}
    \vspace{-10pt}
\end{table*}

\begin{figure*}[t]
\centering
\includegraphics[width=0.85\textwidth]{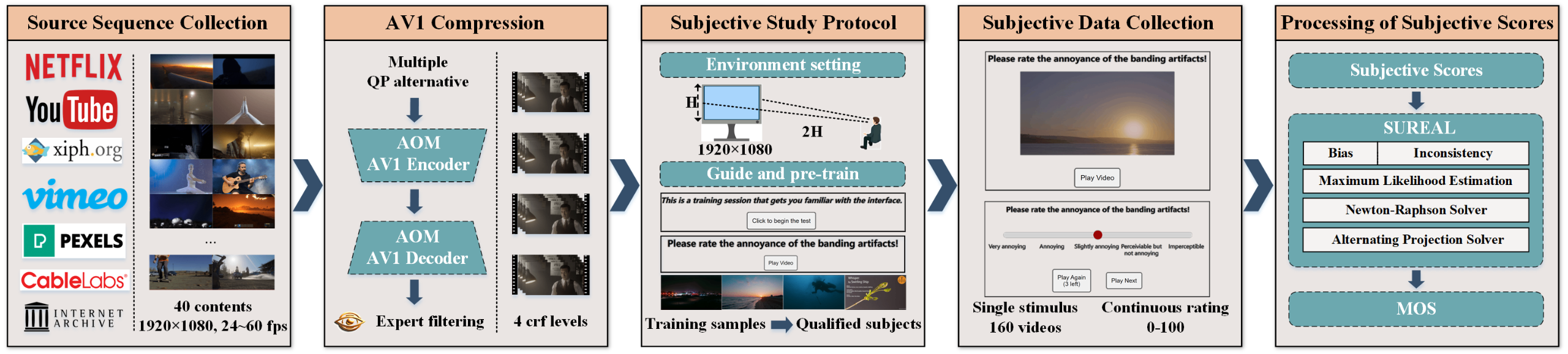}
\vspace{-10pt}
\caption{Workflow of the LIVE-YT-Banding database construction. The five main elements are: 1) source sequence collection; 2) AV1 compression; 3) subjective study protocol; 4) subjective data collection; 5) processing of subjective scores.}
\vspace{-15pt}
\label{fig:workflow_database_creation}
\end{figure*}

\subsection{Banding Quality Assessment Databases}
We are aware of only five prior subjective quality studies of banding artifacts: three still picture datasets~\cite{kapoor2021capturing, xue2023large, 10438477} and two video databases~\cite{wang2016perceptual,tandon2021cambi}, with characteristics summarized in Table~\ref{table:metadata_datasets}.
Among these, Kapoor et al.~\cite{kapoor2021capturing} describe a banding picture database containing 1,440 images of $1920\times1080$ resolution, sampled from 600 high-definition videos. Bit-depth reduction followed by bit-depth expansion was applied on all the pictures, using six different quantization levels. All of the pictures were segmented into patches of size $235\times235$, each of which was automatically labeled as banded (distorted by banding artifacts) or as non-banded. The patch-level banding database that was thereby obtained was then used to train learning-based picture banding models. Xue et al.~\cite{xue2023large} collected 72 1080p standard dynamic range (SDR) source videos on which they applied 8-bit AVC compression with a target bitrate of 8,500 kbps. A total of 150 pairs of distorted video frames were selected, along with 10 uncompressed frames. Both frame-level human opinion scores and patch-level two-forced choice (2FC) scores were collected in a subjective study. Chen et al.~\cite{10438477} created so far the largest banding IQA database, comprising 2,000 images generated from 15 compression and bit-depth quantization schemes across 873 source videos. The subjective IQA experiment involved 23 workers, producing over 214,000 patch-level banding labels and 44,371 reliable image-level quality ratings. 
Wang et al.~\cite{wang2016perceptual} proposed the first banding-relevant true video quality database, wherein seven clips of time/space resolutions 30fps/720p were transcoded using 3 levels of quantization of VP9 compression, yielding a total of 21 distorted video sequences. Tandon et al.~\cite{tandon2021cambi} leveraged nine 10bit, 4K source video clips having durations between 1 and 5 seconds to construct an 86-clip database. One clip had no banding, serving as a reference against the rest of the clips, which were subjected to varying degrees of distortion, including spatial downsampling, bit-depth reduction, and AV1 compression using three different quantization parameters. 
Importantly, none of the above-mentioned video banding databases are publicly accessible. The only open-source picture banding image database~\cite{kapoor2021capturing} provides patch-level banding labels, yet lacks picture-level quality scores.

\subsection{Prior Banding Quality Assessment Models}




Early work on banding detection mainly focused on detecting false contours~\cite{huang2016understanding,lee2006two,daly2004decontouring} or false segments~\cite{wang2016perceptual,bhagavathy2009multiscale,baugh2014advanced,5712203,7032274}.
The false contour detection algorithms in~\cite{huang2016understanding,lee2006two,daly2004decontouring} measured the degree of monotonicity local gradients, contrasts, or entropies, to measure potential banding edge statistics. False segment detection methods, utilized segmentation at the pixel level~\cite{bhagavathy2009multiscale, baugh2014advanced,wang2016perceptual} or block level~\cite{5712203, 7032274}. Bhagavathy et al.~\cite{bhagavathy2009multiscale} proposed to detect the possible presence and scale of banding around each pixel by calculating the likelihood of banding via a multi-scale analysis. Baugh et al.~\cite{baugh2014advanced} sought to measure the presence of banding based on the distribution of blocks, which they defined as groups of connected pixels having the same RGB color. Wang et al.~\cite{wang2016perceptual} observed that the areas of bands, and the contrasts across banding contours are two essential factors affecting the visibility of banding. They proposed a banding detector incorporating both edge length and contrast. These above-described algorithms are limited in assessing the severity of video ~\cite{huang2016understanding,lee2006two,daly2004decontouring}, are sensitive to edge noise~\cite{bhagavathy2009multiscale, baugh2014advanced,wang2016perceptual}, and often misclassifying blocks correctly where banding and textures coexist~\cite{5712203, 7032274}.


More recent banding detection algorithms have attempted to address these problems by accounting for human perception. For example, Tu et al.~\cite{tu2020bband} built a completely blind video banding detector based on edge detection techniques and various models of human vision. In their approach, a pixel-wise banding visibility map is first generated, based on which spatiotemporal importance pooling is applied, yielding frame-level and video-level banding scores. Tandon et al.~\cite{tandon2021cambi} model the human Contrast Sensitivity  Function (CSF) to account for the possible presence of multiple contrast steps and their spatial frequency expressions on banding visibility. Kapoor et al.~\cite{kapoor2021capturing} trained a Deep Neural Network (DNN) model to classify picture patches into `banded' or `non-banded' categories, followed by aggregation of patch-level labels to yield overall picture banding predictions. A banding map can also be generated based on the patch-level labels, to capture spatial variations of banding. Krasula et al.~\cite{krasula2022banding} proposed a banding-aware video quality metric as a simple linear combination of VMAF~\cite{li2016toward} and CAMBI~\cite{tandon2021cambi}. Similar to~\cite{kapoor2021capturing}, Chen et al.~\cite{chen2023fs} developed a no-reference picture banding detection model capable of generating pixel-wise banding maps, as well as overall banding scores. Each analyzed picture is pre-processed into patch-level frequency maps, which are fed into dual-CNN model that classifies the patches as either banded or non-banded. Lastly, a spatial frequency masking module yields a banding map and a whole-picture banding score. 

    \begin{figure*}[t]
\centering
\includegraphics[width=0.8\textwidth]{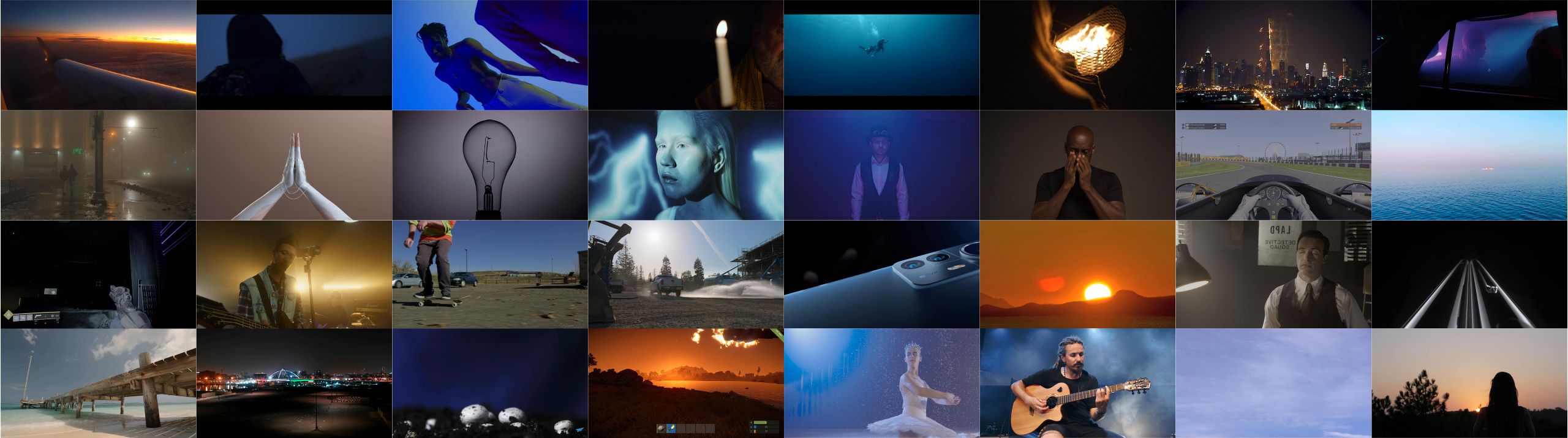}
\vspace{-5pt}
\caption{Examples of source contents included in the LIVE-YT-Banding dataset. Top three rows: thumbnails of original high quality video contents. Left four contents of bottom row: thumbnails of UGC content. Right four contents of bottom row: thumbnails of negative samples.}
\vspace{-15pt}
\label{fig:overview_source_content}
\end{figure*}

\section{LIVE-YT-Banding Dataset Creation}
\label{sec:database_creation}

\subsection{Source Sequence Curation}
An overall diagram of the database curation is given in Figure~\ref{fig:workflow_database_creation}. Banding artifacts mostly manifest on smooth, low-gradient areas of video frames, often associated with projections of sky, water, fog, sunsets, and night scenes, and similar areas dominated by low frequencies. 
Banding is also more of an issue on high-resolution, high-quality videos, where it is often quite visible, than on low-quality clips.
We collected a dozen candidate videos prone to banding from compression, from various open video corpuses, including Waterloo1K~\cite{duanmu2020modeling,duanmu2020characterizing}, YouTube-UGC~\cite{wang2019youtube}, Xiph AV2 Test Sequences~\cite{montgomery1994xiph}, Netflix Open Content~\cite{netflixopen}, Mitchimartinez~\cite{mitchmartinez}, VIMEO~\cite{xue2019video}, Pexels~\cite{pexels}, the Internet Archive~\cite{Internetarchive}, Cablelabs4K~\cite{CableLabs}, and various web repositories. 
We also conducted a pilot encoding study of each content, to ensure that it exhibits visible banding after compression.
To account for UGC use cases, we also included 10 additional clips that exhibit visible banding effects of different degrees (without additional compression). 
Our dataset contains more professionally generated content (PGC) than UGC since banding artifacts are more prevalent in high-resolution PGC, which undergoes heavy compression in streaming. While UGC typically has lower resolution and varied compression, we included clips with visible banding to ensure a balanced and realistic representation of real-world streaming scenarios.
We also curated five negative samples that did not exhibit any perceivable banding either before or after compression, but could be ``expected'' to exhibit banding because they contain large areas having smooth gradients.

The data curation process yielded a total of 40 source videos representing diverse banding-prone scenarios, including positive and negative samples. A representative video subset is shown in Figure~\ref{fig:overview_source_content}. 
Our target use cases are consumers viewing content on laptops or mobile devices with a maximum 1080p resolution. Since much streamed content is at least 4K, but downscaled on the smaller displays in our target use case, we selected source videos having resolutions of at least 1080p. All the videos larger than 1080p were thus downsized to 1080p. This has the additional benefit of excluding resolution factors, while allowing for varied framerates.

\subsection{Encoding Settings}
\begin{figure*}[t]
\centering
\footnotesize
\begin{tabular}{c}
\includegraphics[width=0.65\textwidth]{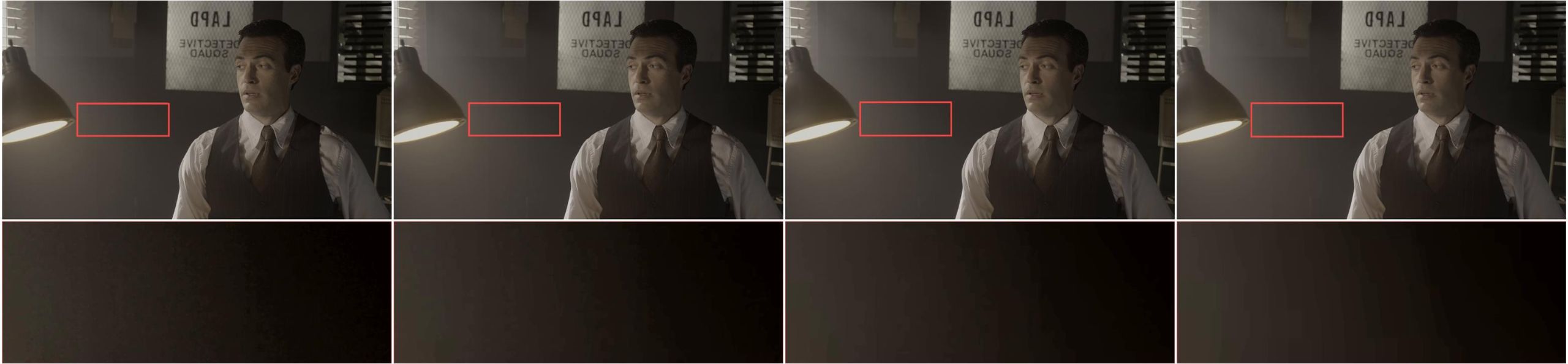} \\
(a) Examples of a high quality video content encoded with different AV-1 crf levels (from left to right crf= 0, 15, 23, 39)\\
\includegraphics[width=0.65\textwidth]{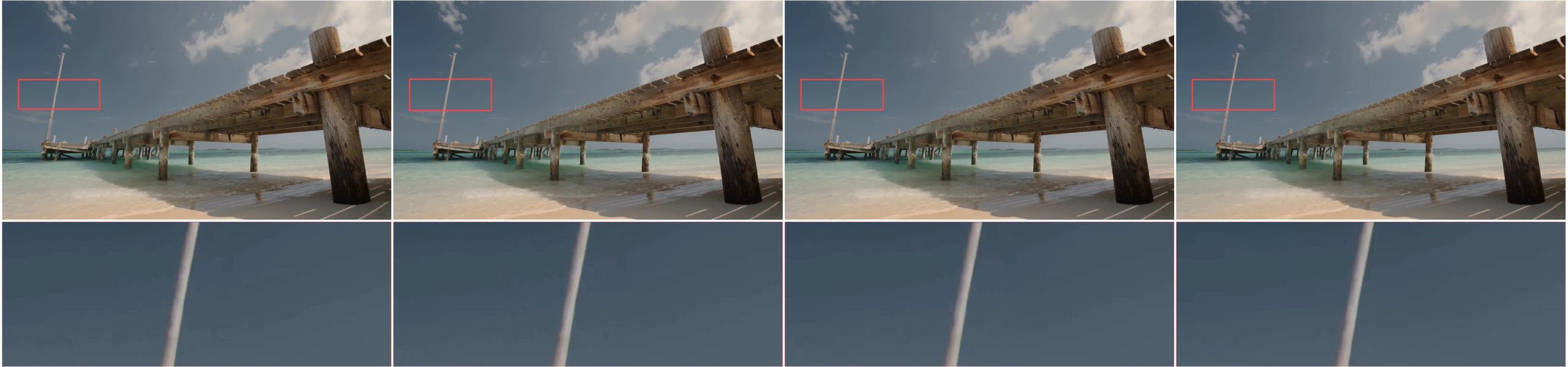}\\
(b) Examples of a UGC content encoded with different AV-1 crf levels (from left to right crf= 0, 11, 23, 39)\\
\includegraphics[width=0.65\textwidth]{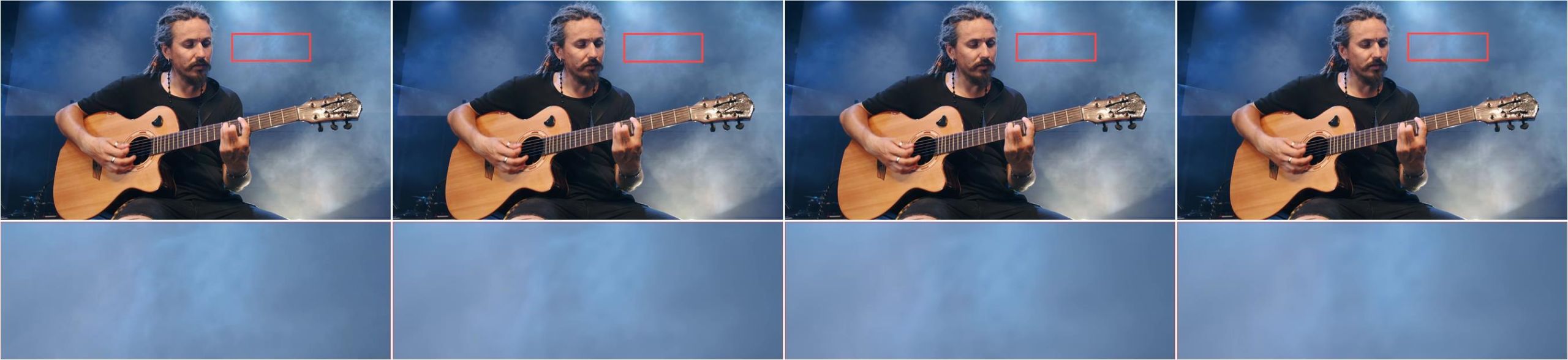}\\
(c) Examples of a negative sample encoded with different AV-1 crf levels (from left to right crf= 0, 15, 25, 39)\\
\end{tabular}
\vspace{-5pt}
\caption{Examples of banded video frames generated by AV1 compression using different crf levels. The crops have been contrast-enhanced for better visualization. Since these distortions are subtle, the reader should zoom in for better visibility.} 
\vspace{-15pt}
\label{fig:visual_compressed_content}
\end{figure*}

Since banding is a subtle distortion, it is important to carefully design the distortion space.
We ensured that included banding artifacts would present a wide range of severities allowing for perceptual separablility to enable better model learning.
%
%
Towards this end, we conducted a pre-screening procedure to generate a ladder of compressed versions of each video using different constant rate factors (crf): crf=(11, 15, 19, 23, 27, 31, 35, 39, 47, 55, 63) of AV1 compression, then asked a few knowledgeable video experts to manually select three different levels of crf so that each adjacent pair of crf levels (including the reference) exceeded one just-noticeable-difference (JND)~\cite{wang2017videoset} of banding visibility, determined by video quality experts. This expert-driven selection ensures that each chosen compression level introduces a perceptible change in banding artifacts. Larger crf values indicate higher compression levels.
After examining these results, we selected three levels of crf (11, 23, and 37), roughly corresponding to little banding, moderate banding, and heavy banding, respectively, with examples shown in Figure~\ref{fig:visual_compressed_content}.
The crf levels of a couple of sequences containing special content (e.g., electronic games) were separately chosen, yielding slightly different crf values. It is worth mentioning that we do not deploy very high crf values since encoding with more extreme settings tended to produce a preponderance of blocking artifacts, rather than banding.

\begin{figure}[t]
\centering
\footnotesize
\includegraphics[width=0.8\linewidth]{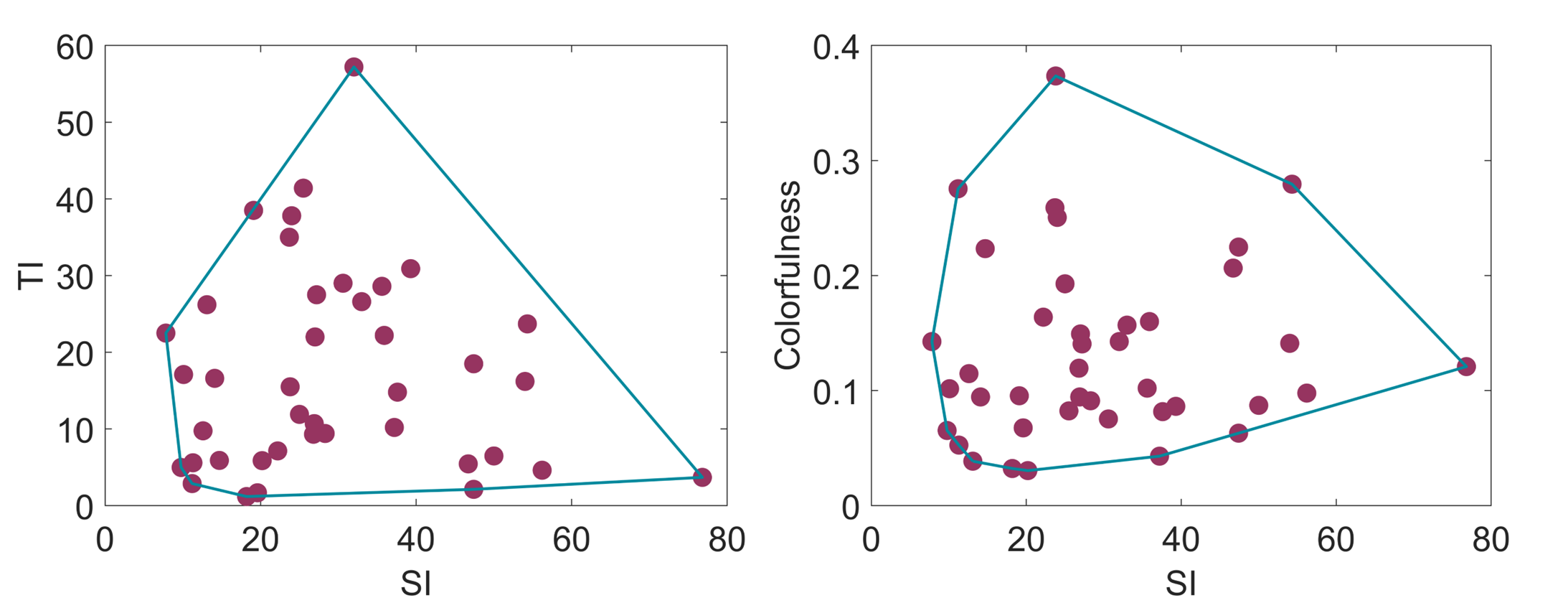} 
\vspace{-8pt}
\caption{Source content (red dots) distribution in paired feature space with corresponding convex hulls (blue boundaries).} 
\vspace{-14pt}
\label{fig:si-colorfulness-ti}
\end{figure}


\subsection{Content Diversity}
As suggested by Winkler~\cite{winkler2012analysis}, spatial activity, temporal activity, and colorfulness can be measured to characterize the content diversity of videos in a database. We calculated the following features on the 40 source contents we selected: colorfulness~\cite{WOS:000184179500009}, spatial information (SI)~\cite{rec2008p}, and temporal information (TI)~\cite{rec2008p}. We calculated each feature on each frame, then averaged them over frames to obtain overall scores. Scatter plots and convex hulls of pairs of these features computed on each video are depicted in Figure~\ref{fig:si-colorfulness-ti}, which shows that the source videos include a diverse range of visual content.

\begin{figure}[t]
\centering
\includegraphics[width=0.75\linewidth]{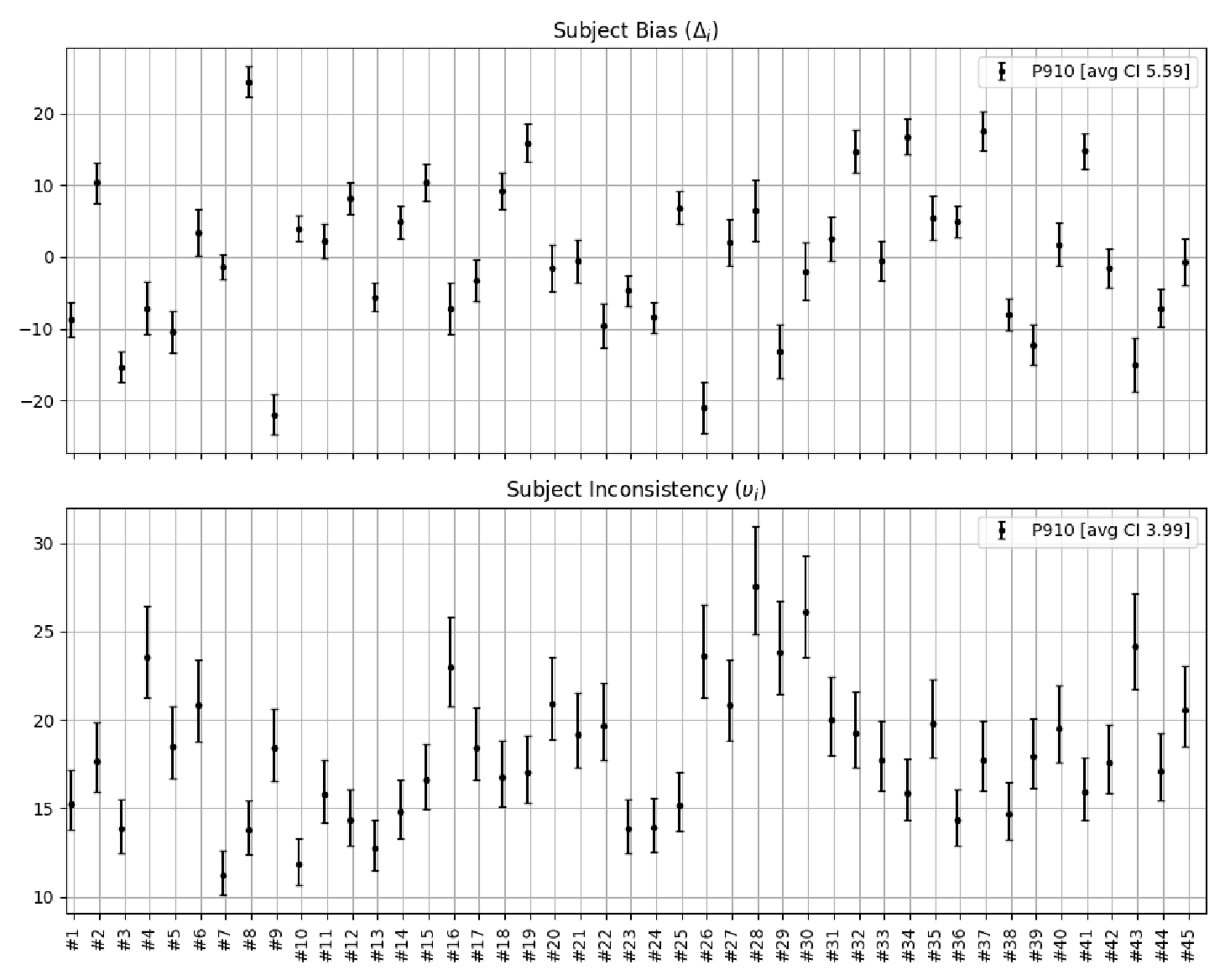}
\vspace{-10pt}
\caption{Bias and inconsistency of each participant in the subjective experiment.}
\vspace{-18pt}
\label{fig:sureal_bias_inconsistency}
\end{figure}

\begin{figure}[t]
\centering
\footnotesize
\includegraphics[width=0.7\linewidth]{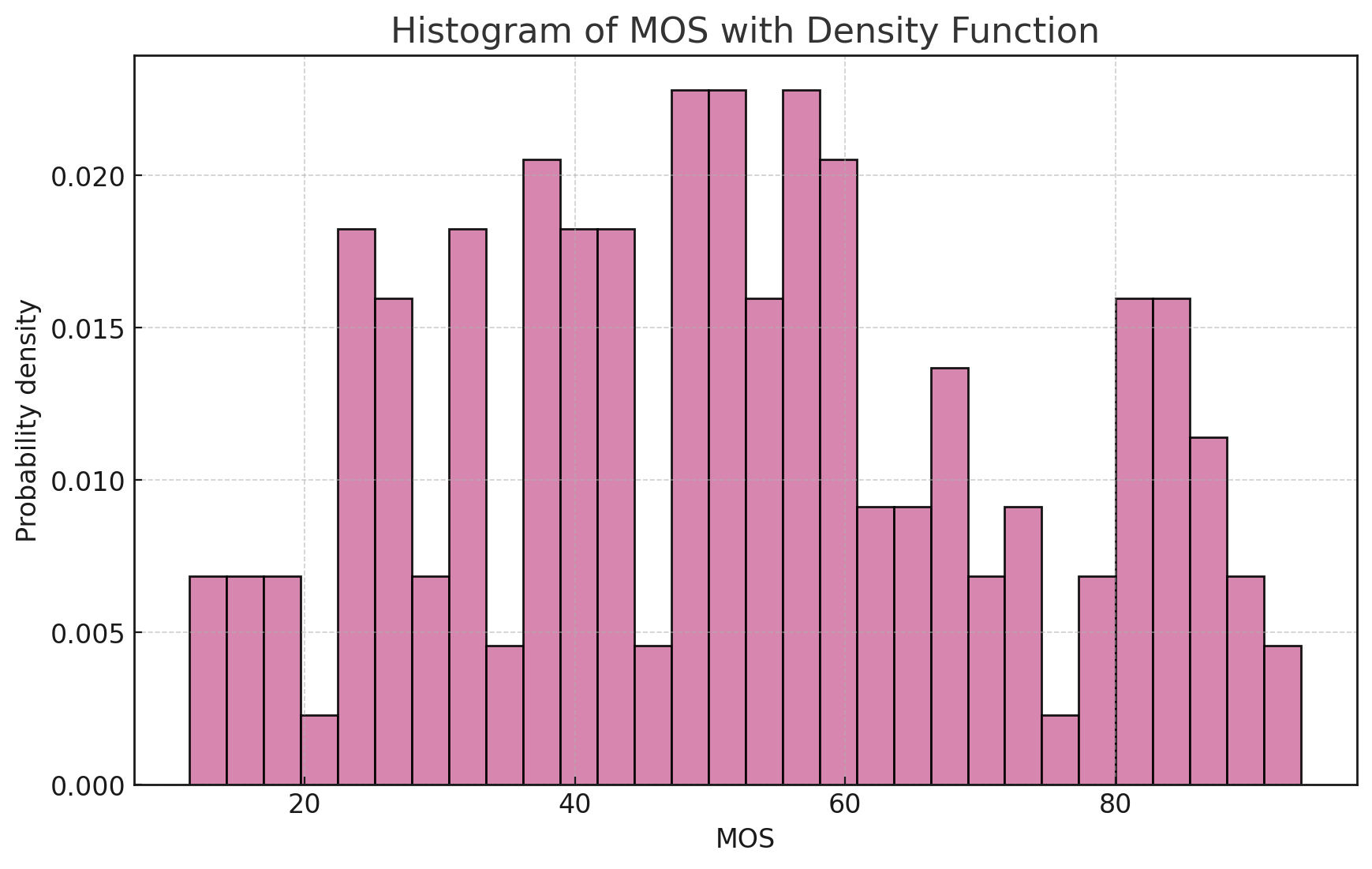} 
\vspace{-10pt}
\caption{Histogram of MOS on the LIVE-YT-Banding Database.}
\vspace{-12pt}
\label{fig:mos_hist}
\end{figure}

\begin{figure}[!tb]
\centering
\subfigure[Interface for playing videos.]{\includegraphics[width=0.35\textwidth]{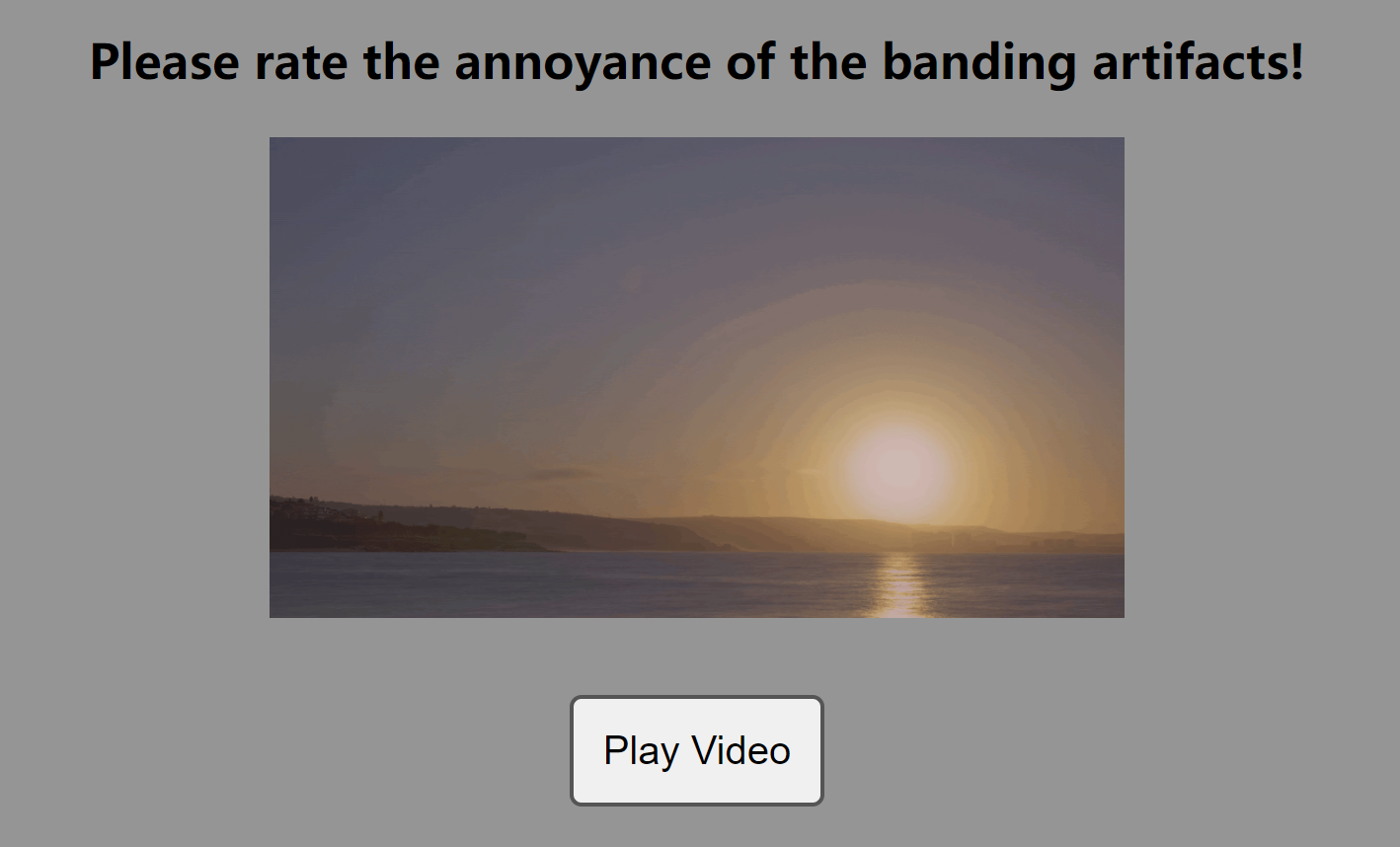}}
\\ [-1ex]
\subfigure[Interface for applying subjective ratings.]{\includegraphics[width=0.35\textwidth]{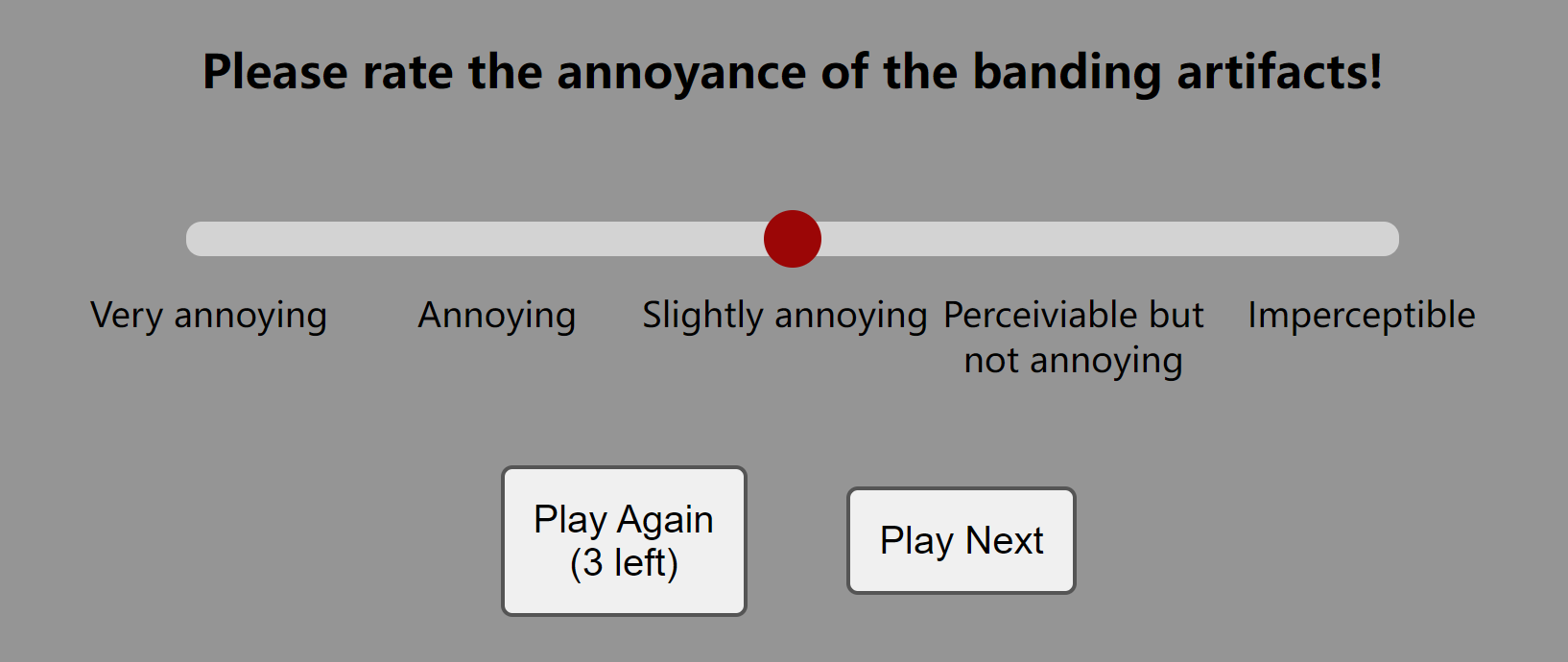}}
\\ [-2ex]
\caption{The human study interface used to record subjective assessments.}
\vspace{-20pt}
\label{fig:gui_subjective_assessment}
\end{figure}

\section{Subjective Video Quality Study}
\label{sec:subjective_study}

\subsection{Subjective Experiments}
We then conducted a controlled human study of the perceptual quality of the 160 curated videos. The study was conducted using 
a single-stimulus continuous absolute category rating (ACR) protocol.

\textbf{Subject Training.} Banding is a subtle distortion, yet can be quite annoying, especially since a single banding artifact may traverse a large portion of a video frame, and may be active across frames. We designed an initial instruction phase, where each participant familiarized themselves with the perceivability or ``annoyance'' of banding. Specifically, each subject was asked to view five training samples exhibiting different levels of banding, so they could visually experience and consider a variety of banding artifacts and severities.

\begin{figure*}[t]
\centering
\includegraphics[width=0.8\textwidth]{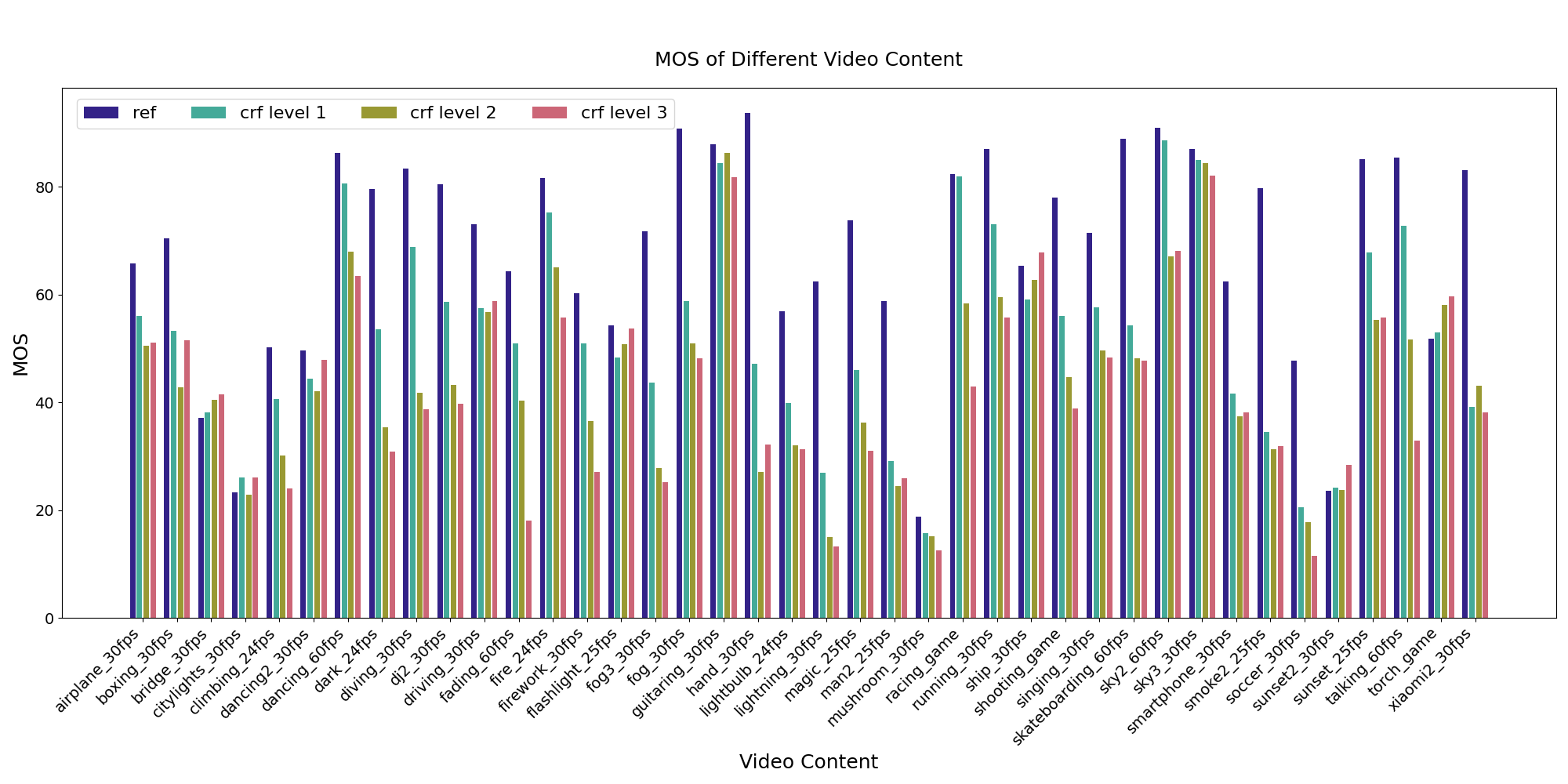}
\vspace{-12pt}
\caption{MOS distribution for video sequences of various content under different crf levels.} 
\vspace{-15pt}
\label{fig:mos_seq}
\end{figure*}

\textbf{Experimental Design.} All of the videos were displayed in full-screen on a 1080p monitor. We followed the standard single stimulus procedure described in ITU-T P.910~\cite{rec2008p}, whereby the videos were displayed to each subject in a different randomized order. The user interface, shown in Figure~\ref{fig:gui_subjective_assessment}, was designed using a web app as a lightweight solution that was compatible with different operating systems. After each video to be rated was played, a rating bar was displayed. To help the subjects understand the range of ratings they could apply, five Likert markers ranging from ``very annoying" to ``imperceptible" are included on the scale, as shown in Figure~\ref{fig:gui_subjective_assessment}(b). Each continuous score collected was quantized to an integer value in the range [1, 100].

\textbf{Subject Rating.} A total of 45 human subjects were recruited from the student population at The University of Texas at Austin. Each subject was asked to rate all 160 videos, yielding 45 x 160 = 7200 human opinion scores. When rating each video, the subjects used a mouse to apply their ratings. After deciding on a quality score, the subject could either ``Play Next'' to proceed to the next test video, or ``Play Again" (up to three times), since it could help them rate subtle banding distortions more accurately.

\subsection{Post-Processing of Subjective Scores}
\label{ssec:post_processing_mos}
There are several ways in which subjective scores can be converted into Mean Opinion Scores (MOS).
The recommendations in ITU-R BT.500~\cite{bt2002methodology}, ITU-T P.910~\cite{rec2008p} and ITU-T P.913~\cite{union2016methods}
standardize the types of post-processing procedures that can be applied on raw opinion scores to conduct subject outlier rejection and bias removal. However, a statistically optimal method called SUREAL~\cite{li2020simple} has recently emerged, which computes Maximum Likelihood (ML) estimates of the mean opinion scores (MOS) under a simple noise model, while also estimating the subjective quality of each impaired stimulus (true score), along with the bias and inconsistency of test subjects, and the overall ambiguity of the visual contents.

Formally, opinion scores $Q_{e,s}$ are represented as random variables:
\vspace{-0.6em}
\begin{equation}
\vspace{-0.6em}
    Q_{e,s} = q_e + T_{e,s} + R_{e,s},
\end{equation}
\vspace{-0.6em}
\begin{equation}
\vspace{-0.6em}
    T_{e,s} \sim N(t_s, v^2_s),
\end{equation}
\vspace{-0.6em}
\begin{equation}
\vspace{-0.6em}
    R_{e,s} \sim N(0, a^2_c), 
\end{equation}
$Q_{e,s}$ is a raw opinion score, $q_e$ is the true quality score of the stimulus $e$, and $T_{e,s}$ is the noise factor of subject $s$ when rating stimulus $e$. $T_{e,s}$ is assumed to follow a Gaussian distribution, where the mean $b_s$ represents the subject's bias, and the variance $v^2_s$ represents the subject's inconsistency. $R_{e,s}$ refers to the source $c$ corresponding to stimulus $e$, and $a^2_c$ represents the ambiguity of $c$. The estimate of each parameter $(q_e, b_s, v_s, a_c)$ is associated with a 95\% confidence interval. The estimated subject biases and inconsistencies of each participant are plotted in Figure~\ref{fig:sureal_bias_inconsistency}. It may be observed that both the subject biases and inconsistencies are quite dispersed. By accounting for the noise and unreliability of each subject, we consider the quality scores recorded by SUREAL as the ground truth MOS in the database. Figure~\ref{fig:mos_hist} plots the histogram of MOS over the entire database, showing a broad range of recorded qualities. Figure~\ref{fig:mos_seq} plots the MOS distribution of all of the video contents, at four different crf levels (`ref' indicates crf=0, while crf levels from 1 to 3 indicate increasing compression.) It is instructive to observe from Figure~\ref{fig:mos_seq} the nonlinear correlation between the perceptual impact of banding artifacts and the applied crf levels. For many of the video contents, the recorded MOS does not consistently decrease with increased crf levels. This suggests that depending on the video content, banding artifacts may appear and predominate as the crf is increased, yet many disappear or have reduced perceptual relevance as it is increased further, e.g., because of flattening. The lack of monotonicity might also be attributed to noise in the perceptual measurements.

\begin{figure*}[t]
\centering
\includegraphics[width=0.8\textwidth]{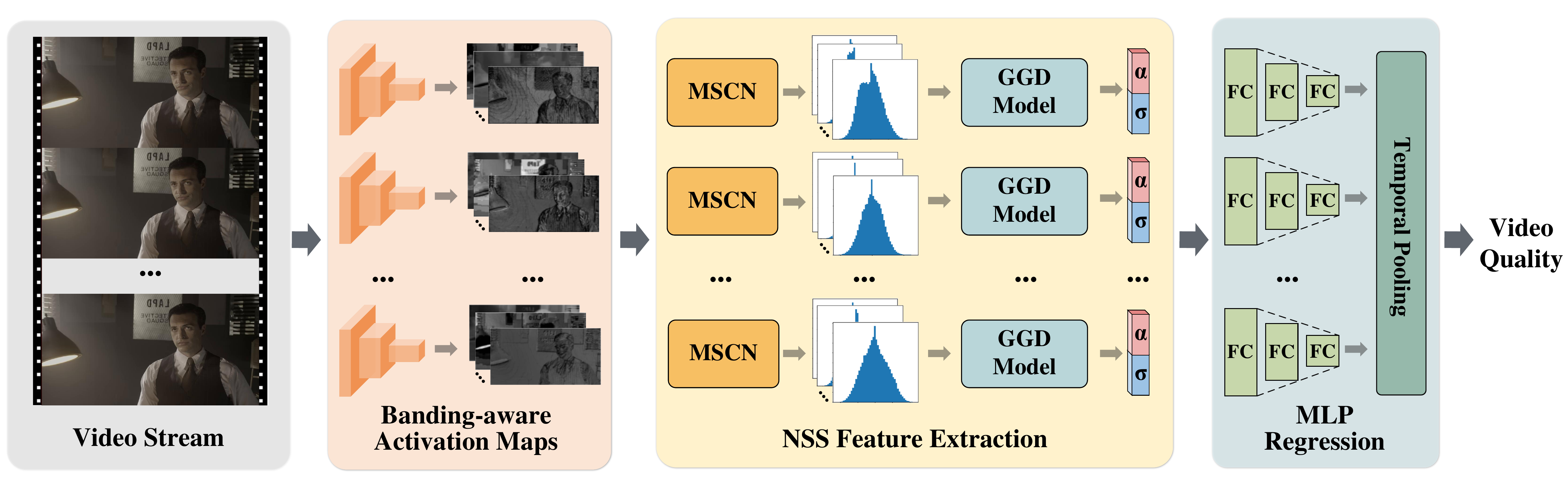}
\vspace{-10pt}
\caption{Schematic diagram of the CBAND video quality metric. Given an input (possibly banded) video, frame-level banding-aware feature maps are derived from the early stages of a pre-trained image classification model. The MSCN transform is performed on each feature map, based on which a set of NSS features are fitted by a GGD model. The extracted NSS feature vector is then fed to three MLP layers which regress the statistical features into frame-level quality scores, which are finally averaged over frames into an overall video banding-quality score.} 
\vspace{-15pt}
\label{fig:objective_framework}
\end{figure*}

\section{CBAND: A CNN-feature-based Banding Metric}
\label{sec:objective_metric}



Recent years have witnessed notable strides in convolutional neural networks (CNN), propelling learning-based methodologies to the forefront of video analysis along with a shift away from reliance on handcrafted features. However, CNN-based solutions suffer from significant computational requirements~\cite{kapoor2021capturing,chen2023fs}, which hampers their wider adoption on real-time video streaming applications. Accordingly, we have crafted efficient and compact NR VQA models tailored to the nuances of compressed banding artifacts. We will refer to this family of models as CBAND. As depicted in Figure~\ref{fig:objective_framework}, the CBAND model mainly consists of three modules: 1) banding-aware activation map extraction; 2) natural scene statistics (NSS) based feature modeling; 3) MLP regression. The following subsections discuss these modules in detail.


\subsection{Banding-aware Activation Maps}
\label{ssec:banding_aware_activation_maps}
\begin{figure*}[t]
\centering
\footnotesize
\begin{tabular}{c}
\includegraphics[width=0.8\textwidth]{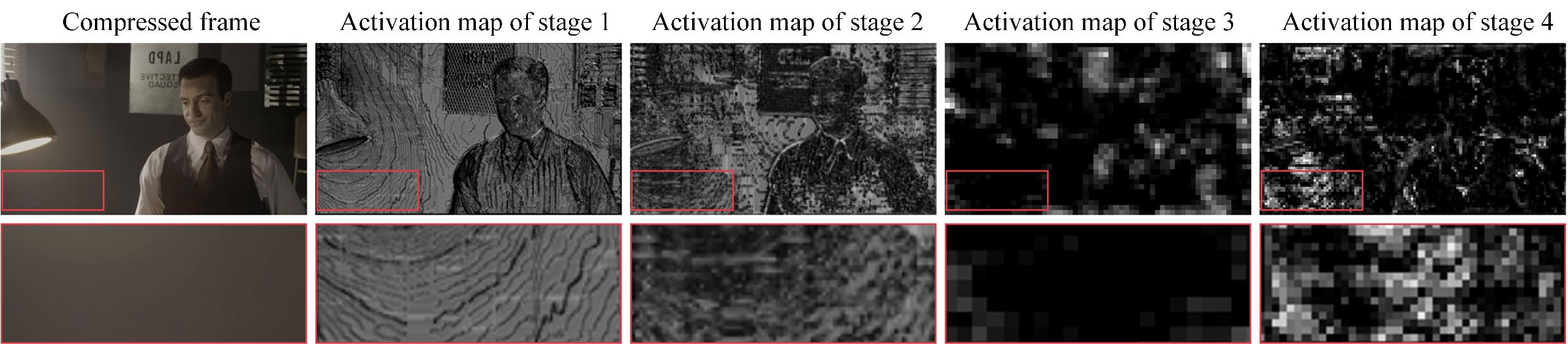} \\
(a) Depiction of information expressive of banding artifacts in the activation maps of increasingly deep stages of a pretrained Resnet50\\
\includegraphics[width=0.8\textwidth]{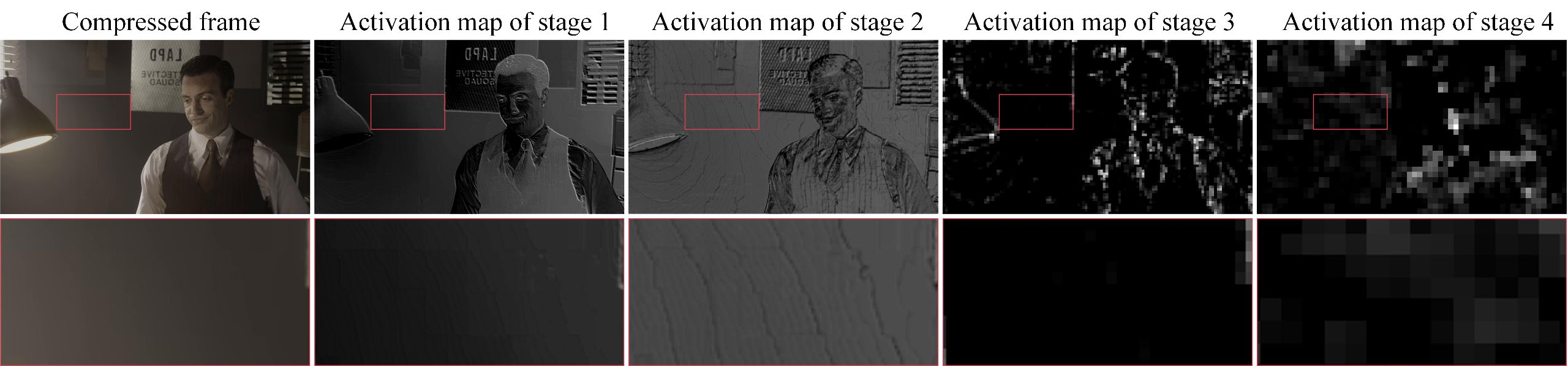}\\
(b) Depiction of information expressive of banding artifacts in the activation maps of increasingly deep stages of a pretrained VGG16\\
\end{tabular}
\vspace{-5pt}
\caption{Visual comparison of the expression of banding artifacts by different stages of a Resnet50 and VGG16.}
\vspace{-15pt}
\label{fig:different_layers_activationmap}
\end{figure*}

It is generally agreed that CNN models learn hierarchical features to represent multiple and increasingly abstract (with depth) levels of image representations. Early stages mostly learn low-level features, while deeper stages learn higher-level semantic embeddings. 
To obtain a deeper understanding of this in the content of banding, we began by conducting a pilot study to analyze how the learned intermediate activation maps from the popular image classification models ResNet50~\cite{he2016deep} and VGG16~\cite{simonyan2014very}), that are pre-trained on ImageNet~\cite{5206848}, respond to banding in videos. 
In other words, we investigated how low-level banding artifacts are encoded at different stages of pre-trained CNN architectures.
More specifically, consider the network structures of Resnet50 and VGG16 shown in Table~\ref{table:network_architecture}, where we define a new stage whenever the resolution is reduced. 
To understand how intermediate activations respond to banding effects, we fed a $1920\times1080$ video frame compressed at $\text{crf}=39$ into the pre-trained Resnet50 and VGG16 networks, then visualized the activation maps from different network stages. We observed that for both architectures, the activation maps from early stages were capable of capturing banding artifacts more accurately than were those of deeper stages, as illustrated in Figure~\ref{fig:different_layers_activationmap}.
%
%
We clarify that early-stage features in ResNet50 and VGG16 are most effective for banding detection, as they capture fine-grained gradient discontinuities without being influenced by high-level semantics. Our observations in Figure~\ref{fig:different_layers_activationmap} shows that deeper layers become less sensitive to banding, aligning with the consensus that early-stage features encode low-level spatial patterns, while deep layers focus on semantic content. This choice also improves computational efficiency, ensuring practicality for real-world banding assessment.

Thus, assuming a video has $T$ frames, the video frames $F_t(t=1,2,...,T)$ are fed into a pre-trained CNN model yielding feature maps $M_t$, denoted as
\vspace{-0.55em}
\begin{equation}
\vspace{-0.55em}
    M_t = CNN_s(F_t),
\end{equation}
where $M_t$ contains a total of $C$ feature maps. $CNN_s$ is the network stack of the first $s$ stages of the pre-trained model. In our implementation, we found that $s=2$ yielded the best results for both the Resnet50 and the VGG16 architectures. Hence, in the following, we utilize the activation maps delivered by stage $2$ when deploying both the Resnet50 and the VGG16 networks. The number $C$ of feature maps $M_t$ for the Resnet50 and the VGG16 was $512$ and $128$, respectively.  

\begin{table}[!htb]
\centering
\fontsize{6.5pt}{6.5pt}\selectfont
\setlength{\tabcolsep}{0.8pt}
\vspace{-7pt}
\caption{Architecture and stages of Resnet50 and VGG16 networks.}
\vspace{-8pt}
\label{table:network_architecture}
    \begin{tabular}{cc|cc|c}
    \toprule
    \multicolumn{2}{c|}{\textbf{Resnet50~\cite{he2016deep}}} & \multicolumn{2}{c|}{\textbf{VGG16~\cite{simonyan2014very}}} &   \\
    ConvNet & \#Output channel & ConvNet & \#Output channel & Stage name\\ \midrule
    \begin{tabular}{@{}c@{}}
$7\times7$, 64 \\
Maxpool \\
\raisebox{-.5\height}{$\begin{bmatrix}
1\times1, 64 \\
3\times3, 64 \\
1\times1, 256
\end{bmatrix}$}\raisebox{-1.5\height}{$\times 3$} \\
\end{tabular}& 256 
& \begin{tabular}{@{}c@{}}
{$\begin{bmatrix}
3\times3, 64 \\
\end{bmatrix}$}\raisebox{0.1\height}{$\times 2$}\\
Maxpool \end{tabular} & 64 & Stage 1\\
\midrule

\raisebox{-.5\height}{$\begin{bmatrix}
1\times1, 128 \\
3\times3, 128 \\
1\times1, 512
\end{bmatrix}$}\raisebox{-1.5\height}{$\times 4$} & 512 & \begin{tabular}{@{}c@{}}
{$\begin{bmatrix}
3\times3, 128 \\
\end{bmatrix}$}\raisebox{0.1\height}{$\times 2$}\\
Maxpool \end{tabular} & 128 & Stage 2 \\
\midrule

\raisebox{-.5\height}{$\begin{bmatrix}
1\times1, 256 \\
3\times3, 256 \\
1\times1, 1024
\end{bmatrix}$}\raisebox{-1.5\height}{$\times 6$} & 1024 & \begin{tabular}{@{}c@{}}
{$\begin{bmatrix}
3\times3, 256 \\
\end{bmatrix}$}\raisebox{0.1\height}{$\times 3$}\\
Maxpool \end{tabular} & 256 & Stage 3 \\
\midrule

\raisebox{-.5\height}{$\begin{bmatrix}
1\times1, 512 \\
3\times3, 512 \\
1\times1, 2048
\end{bmatrix}$}\raisebox{-1.5\height}{$\times 3$} & 2048 & \begin{tabular}{@{}c@{}}
{$\begin{bmatrix}
3\times3, 512 \\
\end{bmatrix}$}\raisebox{0.1\height}{$\times 3$}\\
Maxpool \end{tabular} & 512 & Stage 4 \\
\midrule

\begin{tabular}{@{}c@{}}
Average pool \\
FC layers \\
Softmax  \end{tabular} & 1000 & \begin{tabular}{@{}c@{}}
{$\begin{bmatrix}
3\times3, 512 \\
\end{bmatrix}$}\raisebox{0.1\height}{$\times 3$}\\
Maxpool \end{tabular} & 512 & Stage 5 \\
\midrule

- & - & \begin{tabular}{@{}c@{}}
Maxpool \\
FC layers \\
Softmax  \end{tabular} & 1000 & Stage 6 \\
    
    \bottomrule
    \end{tabular}
    \vspace{-15pt}
\end{table}
\subsection{Natural Scene Statistical Feature Extraction}
High-quality, natural images and video frames reliably exhibit certain statistical regularities that are predictably perturbed by various types and degrees of visual distortions~\cite{ruderman1994statistics,sheikh2006image}.
This empirical observation has fostered a number of blind image/video quality assessment (BIQA/BVQA) models that utilize this regularity in various perceptual domains~\cite{6353522,mittal2012no,xue2014blind,zheng2022completely,kundu2017no,FRIQUEE,zheng2022blind,6705673,li2016spatiotemporal,tu2021rapique,zheng2024faver,zheng2022no}. However, to the best of our knowledge, no prior work has analyzed the NSS properties of the activation maps of pre-trained classification models in the context of banding analysis. We do so as follows.

Consider a feature map $M_t^c(i,j) (c = 1, 2,...C)$. Then, the mean-subtracted contrast-normalized  coefficients (MSCN~\cite{mittal2012no}) of $M_t^c(i,j) (c = 1, 2,...C)$ are defined as
\vspace{-0.6em}
\begin{equation}
\vspace{-0.6em}
\label{eq:mscn}
    \widehat{M_t^c(i,j)} = \frac{M_t^c(i,j) - \mu_t^c(i,j)}{\sigma_t^c(i,j) + C_1},
\end{equation}
where $i \in 1,2,...P$, $j \in 1,2...Q$, are spatial indices of the feature map $M_t^c$, and $C_1=1$ is a saturation constant that reduces numerical instabilities. As in (\ref{eq:mscn}), $\mu_t^c(i,j)$ and $\sigma_t^c(i,j)$ are local weighted sample means and standard deviations given by 
\vspace{-0.7em}
\begin{equation}
\vspace{-0.7em}
    \mu_t^c(i,j) = \sum_{k=-K}^K \sum_{l=-L}^L \omega_{k,l} M_t^c(i,j)_{k,l},
\end{equation}
and
\vspace{-0.7em}
\begin{equation}
\vspace{-0.7em}
    \sigma_t^c(i,j) = \sqrt{\sum_{k=-K}^K \sum_{l=-L}^L \omega_{k,l} (M_t^c(i,j)_{k,l} - \mu_t^c(i,j) )^2},
\end{equation}
where $\omega={\omega_{k,l}|k = -K,...,K, l = -L,...,L}$ is a 2D
circularly-symmetric Gaussian weighting function sampled out to 3 standard deviations and rescaled to unit volume. In our implementation, $K = L = 3$.

The first-order statistics of the MSCN coefficients of high quality natural images/videos strongly tends towards decorrelated Gaussianity. Visual distortions/degradations generally alter this statistical regularity in ways that can be used to accurately predict perceived quality~\cite{mittal2012no, 6353522}. The basic feature extractor is based on simple natural image statistics models. The first basic model is the zero-mean generalized Gaussian distribution (GGD):
\vspace{-0.6em}
\begin{equation}
\vspace{-0.3em}
    f(x;\alpha,\sigma^2) = \frac{\alpha}{2\beta \varGamma(1/\alpha)}\exp(-(\frac{\left\lvert x \right\rvert}{\beta})^\alpha),
\end{equation}
\vspace{-0.3em}
\begin{equation}
\vspace{-0.6em}
    \beta = \sigma \sqrt{\frac{\varGamma(1/\alpha)}{\varGamma(3/\alpha)}},
\end{equation}
where the model parameters $\alpha$ and  $\sigma$ control the shape and variance respectively, and $\varGamma(\cdot)$ is the gamma function. These two parameters are estimated using a popular moment-matching-based method~\cite{sharifi1995estimation}.

\begin{figure}[!tb]
\centering
\subfigure[Histograms of MSCN coefficients.]{\includegraphics[width=0.3\textwidth]{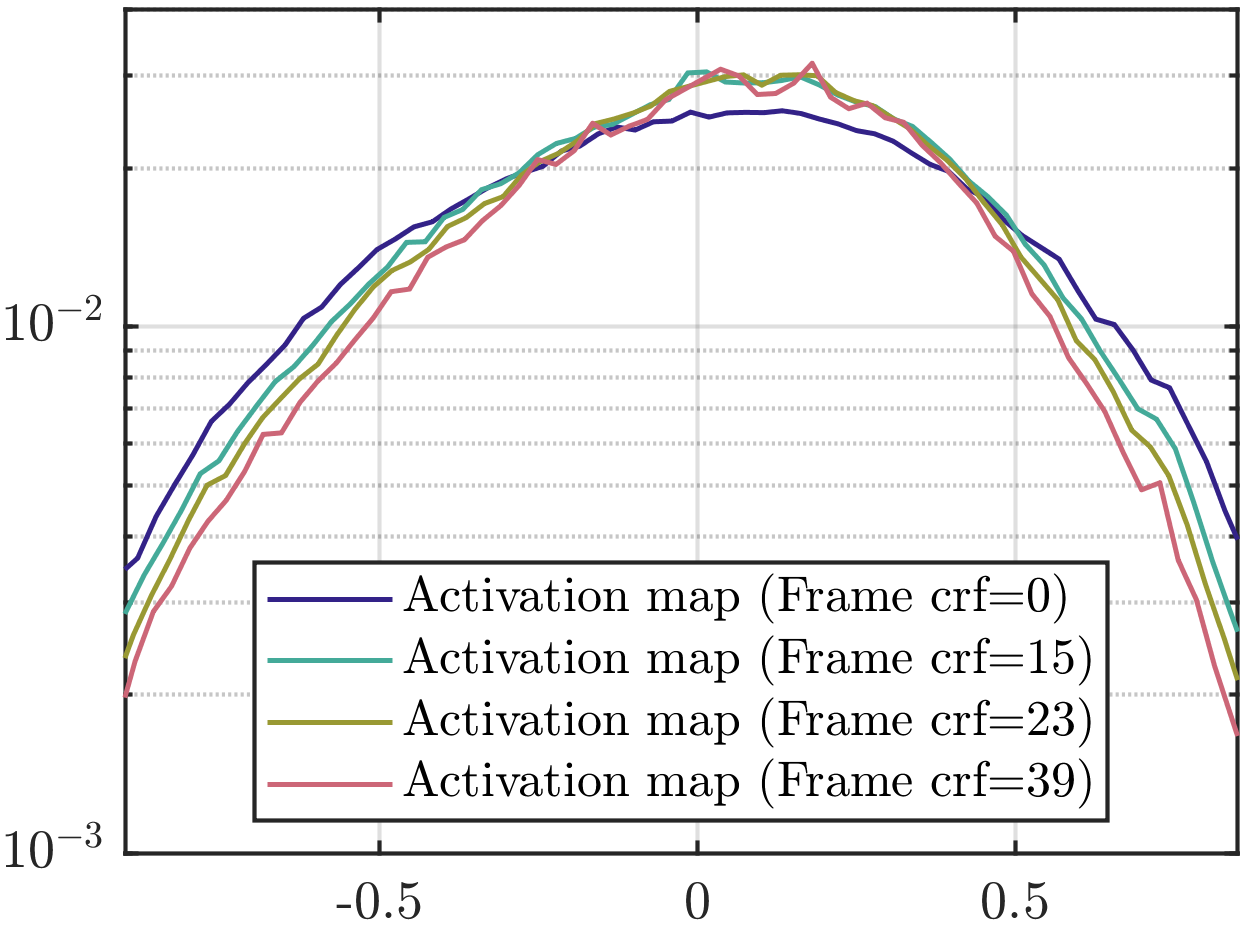}}
\hfil
\vspace{-5pt}
\subfigure[Four activation maps derived from stage 2 of a pretrained VGG16 with input frames individually compressed at crf=(0,15,23,39) (from left to right, respectively).]{\includegraphics[width=0.35\textwidth]{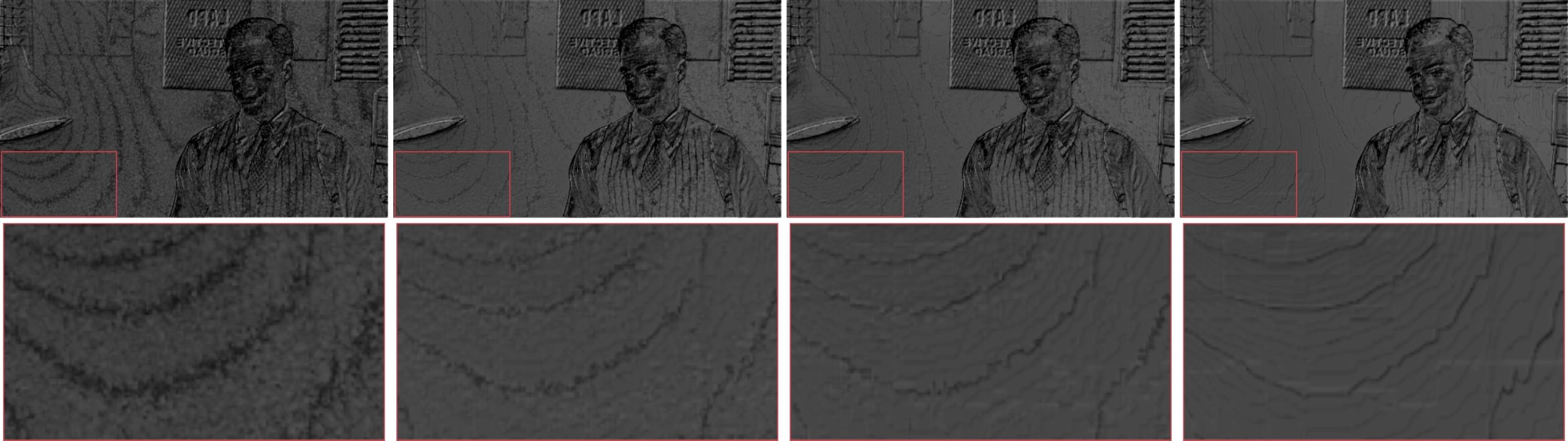}}
\vspace{-5pt}
\caption{Visualization of the effectiveness of modeling banding artifacts using NSS features derived from the MSCN coefficients of the activation maps of a VGG16 network processing a video.}
\label{fig:different_qp_mscn}
\vspace{-16pt}
\end{figure}

To illustrate how banding impairments affect NSS features derived from MSCN coefficients, we extracted frames from four variants of a same source content compressed at $\text{crf}=0,15,23,39$, respectively. These four compressed frames were fed into the pre-trained VGG16, and the representative activation maps derived from stage 2 are shown in Figure~\ref{fig:different_qp_mscn}(b). Figure~\ref{fig:different_qp_mscn}(a) plots the histograms of the MSCN coefficients of these activation maps as a function of the crf. 
The figure clearly suggests that parameters estimated from the GGD distributions fitted by banding-sensitive activation maps are reliable indicators of perceptual banding quality on videos distorted by compression banding artifacts.


We then modeled the MSCN distributions of activation maps derived from early stages of image classification networks using the GGD model, thereby extracting banding-aware statistical features. Given a set of activation maps $M_t^C$ at frame $t$ with $C$ channels, a $2$-dim parameter set $(\alpha, \sigma)$ is obtained for each activation map $M_t^c (c=1,2,...,C)$, thus yielding a $2\times C$-dim feature vector $V_t$ for each frame indexed $t$. Denote the overall NSS feature extraction process as follows:
\vspace{-0.6em}
\begin{equation}
\vspace{-0.6em}
    V_t^c = GGD(MSCN(M_t^c)),
\end{equation}
\vspace{-0.6em}
\begin{equation}
\vspace{-0.6em}
    V_t = V_t^1 \oplus V_t^2 \oplus V_t^3...V_t^{C-1}\oplus V_t^C,
\end{equation}
wherein $\oplus$ is the concatenation operator. In our implementation, the dimension of $V_t$ is $1024$ and $256$ for the Resnet50 and the VGG16, respectively.

Unlike traditional NSS-based methods operating on pixel-level~\cite{mittal2012no,xue2014blind,kundu2017no} or frequency-domain~\cite{zheng2024faver} representations, CBAND innovatively applies NSS directly to shallow pre-trained CNN feature maps identified as sensitive to subtle banding artifacts. And we deliberately select only two NSS features after rigorous experimentation, ensuring high efficiency and robustness. This NSS-CNN hybrid approach significantly advances perceptual quality assessment, extending readily to advanced models (see Section~\ref{exp:alternative_pretrained_architectures}).

\subsection{MLP Regression Head}
We applied three MLP layers with $ReLU$ as the activation function to conduct quality regression, yielding $1$-dim frame-level quality scores, which are denoted as:
\vspace{-0.6em}
\begin{equation}
\vspace{-0.6em}
    q_t = MLP(V_t),
\end{equation}
wherein the MLP layers have a dropout rate of 0.2. We used the $L_1$ loss as the objective function when training the MLP layers. The $L_1$ loss computes the mean absolute error (MAE) between a batch of predicted quality scores and MOS:
\vspace{-0.5em}
\begin{equation}
\vspace{-0.5em}
    L_{1} = \frac{1}{B}\sum_{t=1}^{B}\abs{q_t-\hat{q_t}},
\end{equation}
where $B$ is the batch size, $q_t$ and $\hat{q_t}$ are the predicted score and ground truth score of the $t$-th video frame in the batch, respectively. Lastly, overall video-level quality scores are obtained by average-pooling frame-level quality scores.
\section{Experimental Results}
\label{sec:experiment}


\subsection{Experimental Settings}
\subsubsection{Implementation details}
We will denote our two proposed models as CBAND-RN50 and CBAND-VGG16, depending on which pretrained network is used.
In our implementation, frozen ResNet50~\cite{he2016deep} and VGG16~\cite{simonyan2014very} backbones are used to extract feature maps at the end of the second stage, yielding features across $512$ channels and $128$ channels, respectively.
Afterward, NSS features are computed by fitting a GGD model on the MSCN coefficients of each feature map yielding a feature vectors of $1024$ or $256$ dimensions (RN50 or VGG16).
We employed a three-layer fully connected MLP with a dropout rate of 0.2 to gradually reduce the feature dimensions to final scalar quality predictions.
The CBAND models were implemented in PyTorch~\cite{paszke2017automatic}, trained using L1 loss and optimized using Adam~\cite{kingma2014adam} with an initial learning rate of 0.0001 and a batch size of 32. Each CBAND model was trained over 100 epochs.

\subsubsection{Benchmark Settings}
Note that prior to our work, there exists no open-source banding video quality database equipped with MOS labels, as shown in Table~\ref{table:metadata_datasets}.
Therefore, we conducted all of the experiments on the new LIVE-YT-Banding Database.
We randomly split the database into training and test sets, each containing approximately 80\% and 20\% of the source videos, respectively.
We ensured that the training and test sets shared no versions (distorted or otherwise) of any of the same original contents.
All of the experiments conducted on the LIVE-YT-Banding database were repeated 50 times with different random splits, after which we reported the mean performance metrics.

\subsubsection{Baseline Models}
We included a variety of representative IQA/VQA algorithms as baseline models in our evaluation:
\begin{itemize}
    \item General-purpose FR IQA/VQA models: PSNR, SSIM~\cite{wang2004image}, LPIPS~\cite{zhang2018unreasonable}, and VMAF~\cite{vmaf}. These algorithms are widely used for image and video coding, image reconstruction, and image enhancement.
    \item General-purpose NR IQA models: BRISQUE~\cite{mittal2012no}, GM-LOG~\cite{xue2014blind}, HIGRADE~\cite{kundu2017no}, NIQE~\cite{6353522}, FRIQUEE~\cite{FRIQUEE}, HOSA~\cite{7501619}, and CORNIA~\cite{6247789}. Five NR VQA models were also included: VIDEVAL~\cite{9405420}, TLVQM~\cite{korhonen2019two}, FAVER~\cite{zheng2024faver}, RAPIQUE~\cite{tu2021rapique}, and deep learning-based models: VSFA~\cite{li2019quality}, FAST-VQA~\cite{10.1007/978-3-031-20068-7_31}, FasterVQA~\cite{10264158}, DOVER~\cite{Wu_2023_ICCV}, SAMA~\cite{Liu_Quan_Xiao_Li_Wu_2024}, and ModularBVQA~\cite{Wen_2024_CVPR}.
    \item Banding-specific IQA/VQA models: given the limited research available on banding quality assessment, we selected one deep learning-based banding model called DBI~\cite{kapoor2021capturing}, and three other video banding models, BBAND~\cite{tu2020bband}, CAMBI~\cite{tandon2021cambi}, and $\text{VMAF}_{\text{BA}}$~\cite{krasula2022banding}. Among these three only $\text{VMAF}_{\text{BA}}$ is an FR model, while the other two are NR models.
\end{itemize}

Following the conventions used in~\cite{9405420, tu2021rapique}, we computed features or scores using each of these IQA models at a rate of one frame per second, then averaged the features or scores across all frames to obtain final video-level features or scores.
For all the other video quality models, we simply employed them in their unaltered original forms.

\subsubsection{Evaluation Metrics}
We employ classic metrics for comparing video quality models: Spearman’s rank-order correlation coefficient (SROCC), Kendall
rank-order correlation coefficient (KROCC), Pearson’s linear correlation coefficient (PLCC), and root mean squared error (RMSE).
SROCC and KROCC evaluate the monotonicity of prediction performance, while PLCC and RMSE measure the prediction accuracy.
Note that PLCC and RMSE were computed after performing a nonlinear four-parametric logistic regression to linearize objective predictions to be on the same scale as MOS~\cite{5404314}:
\vspace{-0.7em}
\begin{equation}
\vspace{-0.7em}
\label{eq:logistic}
f(x)=\beta_2+\frac{\beta_1-\beta_2}{1+\exp{(-x+\beta_3/|\beta_4|})}.
\end{equation}

\begin{figure*}[!t]
\centering
\def\xheight{0.24}  
\footnotesize
\subfigure[PSNR]{\includegraphics[width=2.96cm,height=1.7cm]{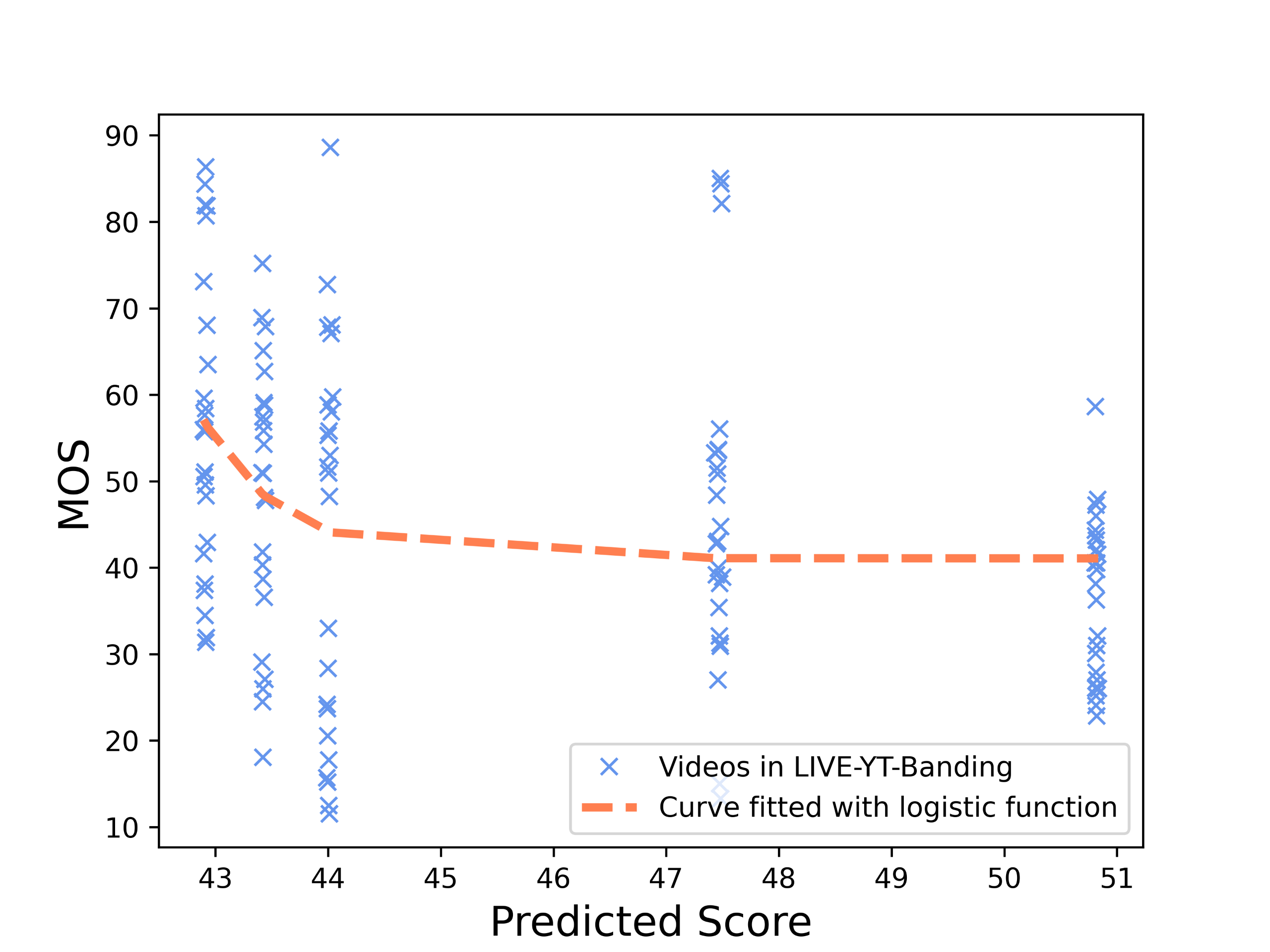}}
\hspace{-3mm} 
\subfigure[SSIM]{\includegraphics[width=2.96cm,height=1.7cm]{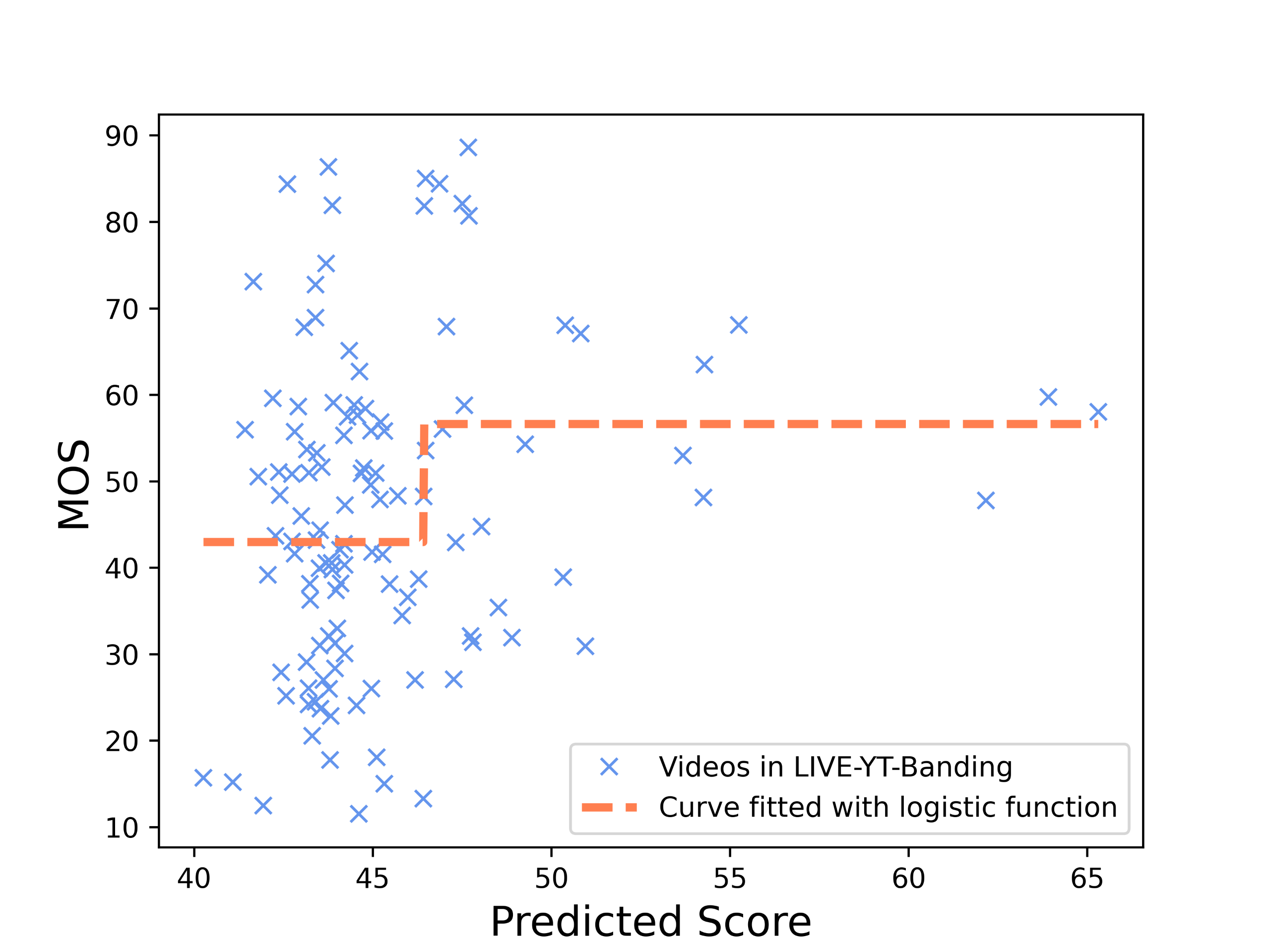}}
\hspace{-3mm} 
\subfigure[LPIPS]{\includegraphics[width=2.96cm,height=1.7cm]{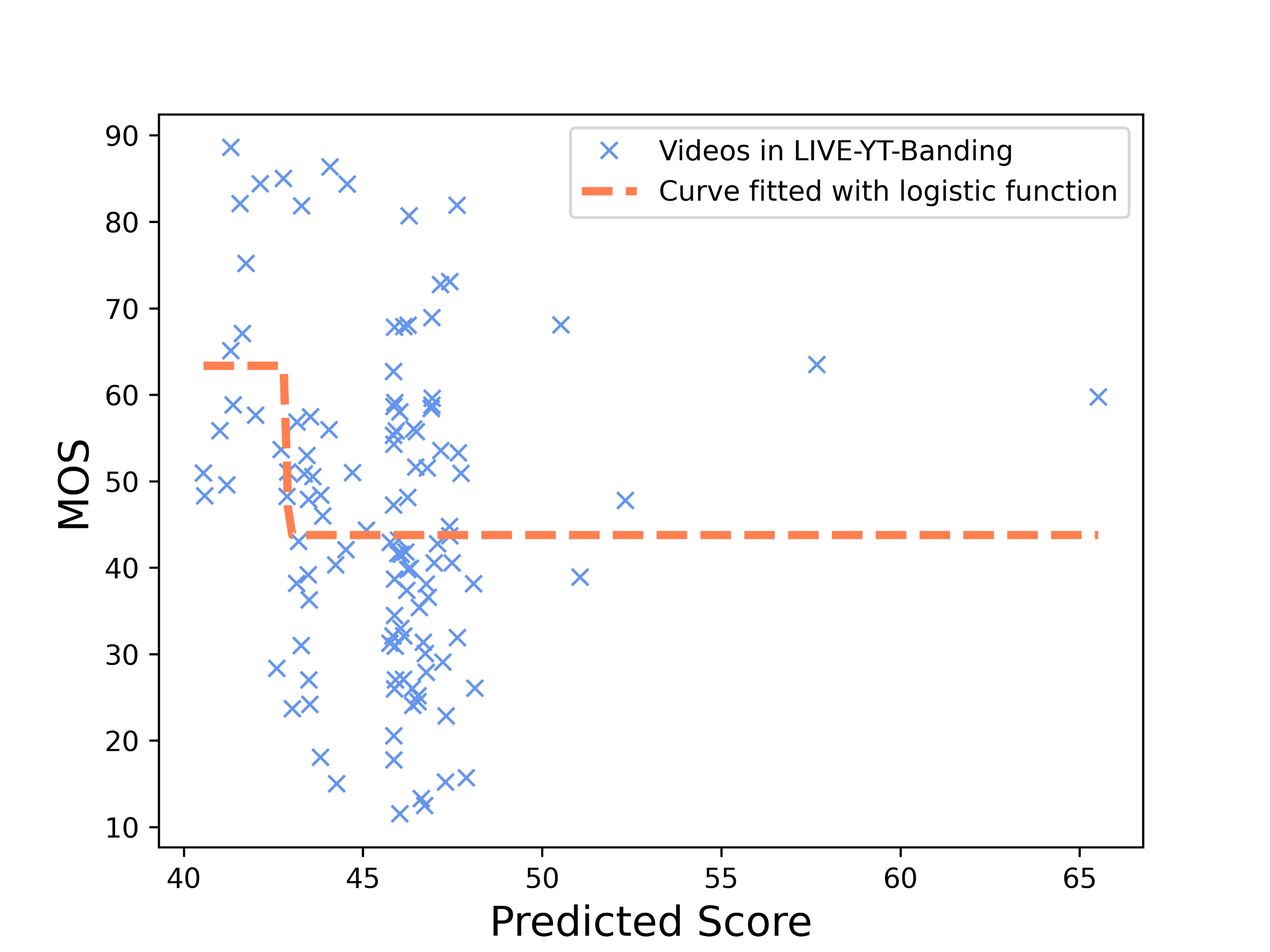}}
\hspace{-3mm} 
\subfigure[VMAF]{\includegraphics[width=2.96cm,height=1.7cm]{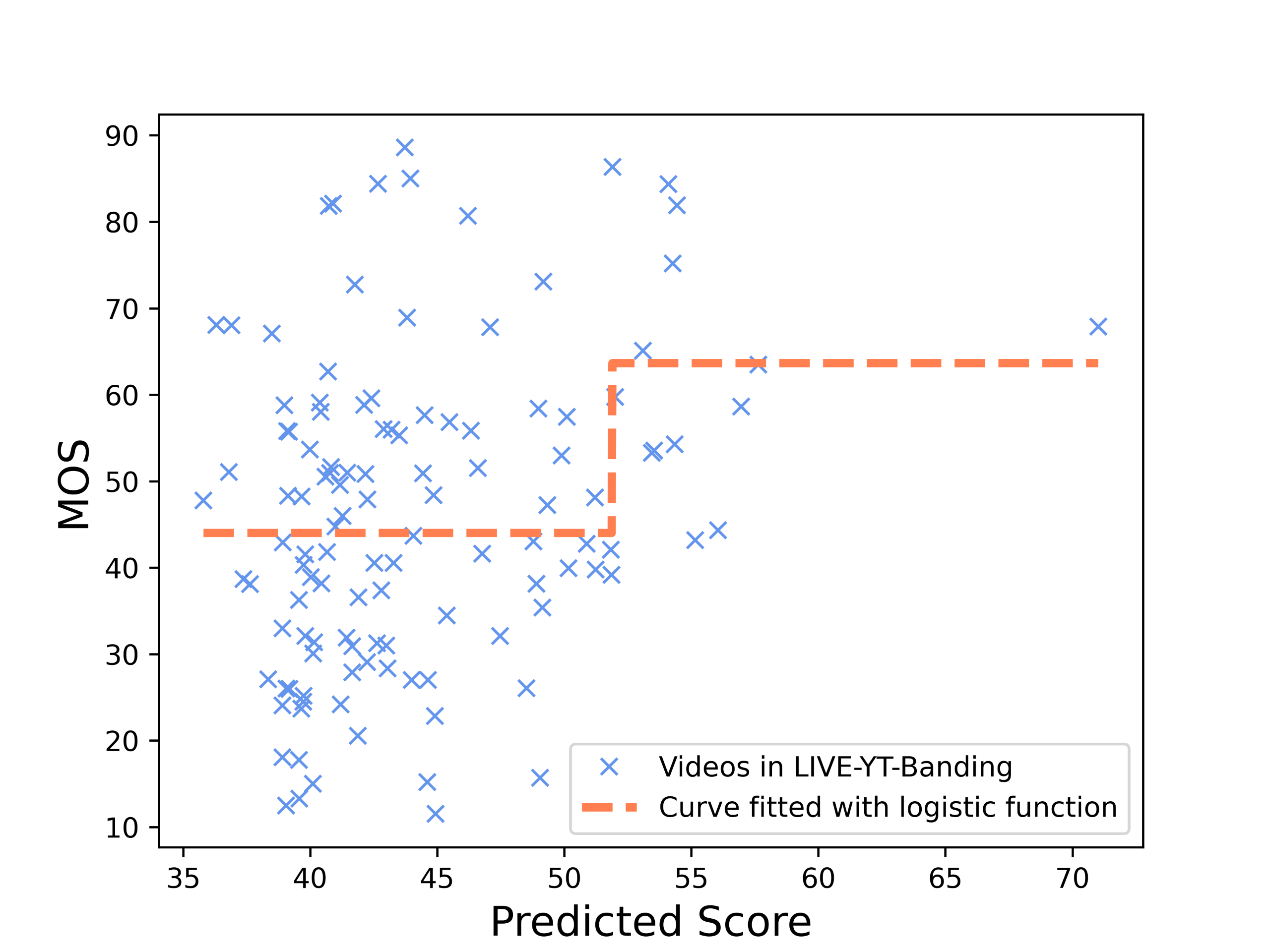}}
\hspace{-3mm} 
\subfigure[BRISQUE]{\includegraphics[width=2.96cm,height=1.7cm]{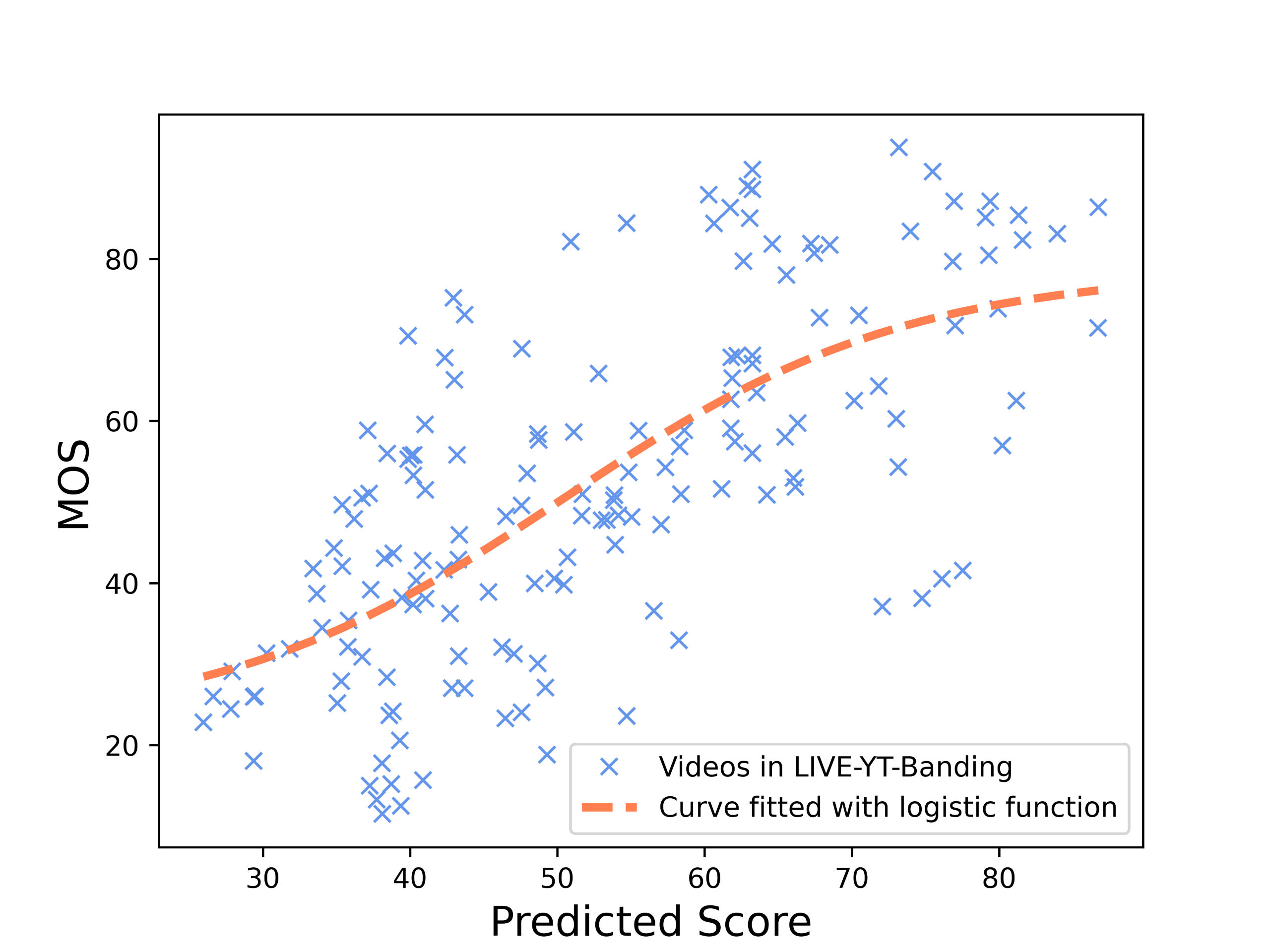}}
\\ [-3ex]
\subfigure[GM-LOG]{\includegraphics[width=2.96cm,height=1.7cm]{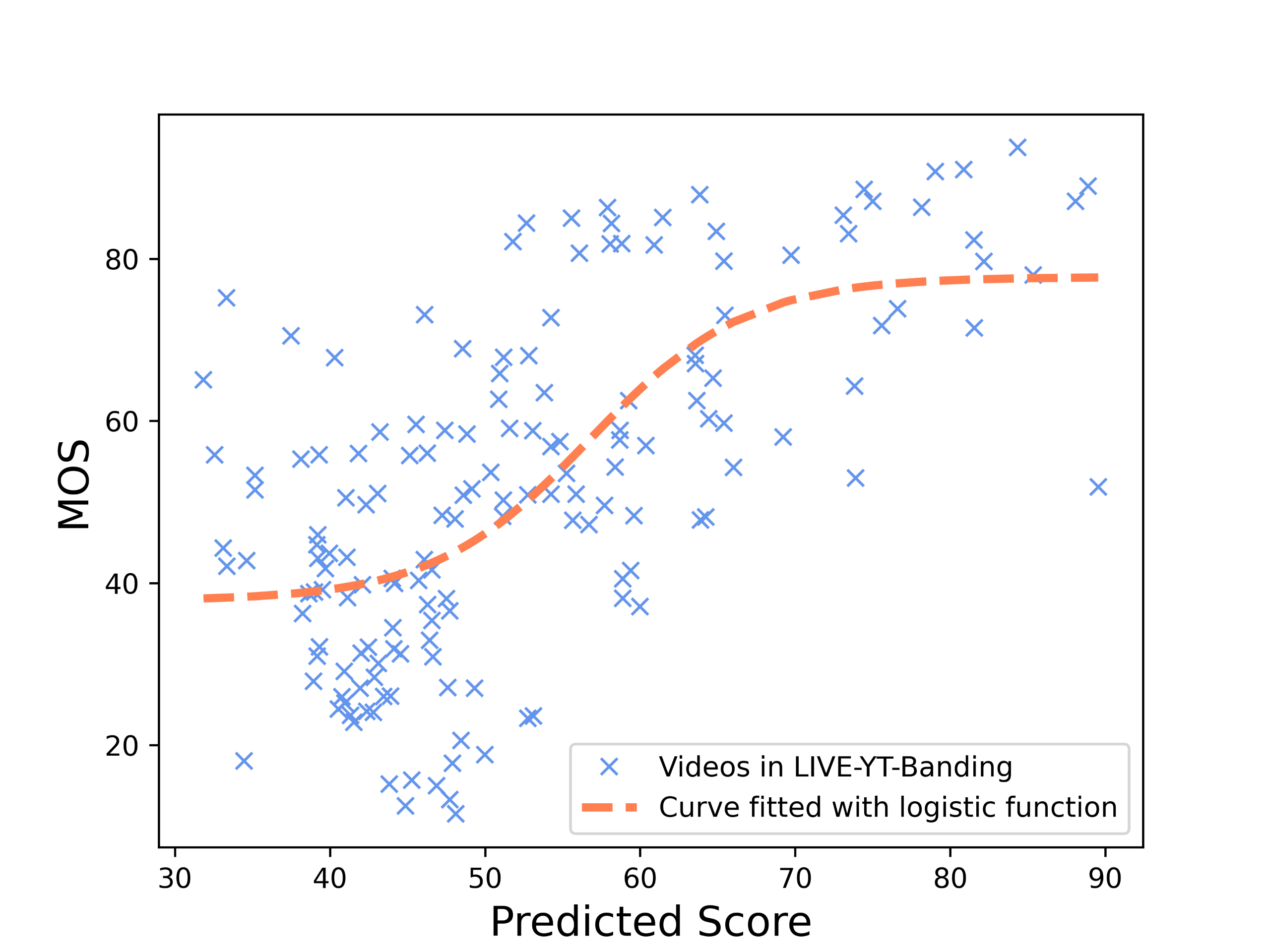}}
\hspace{-3mm} 
\subfigure[HIGRADE]{\includegraphics[width=2.96cm,height=1.7cm]{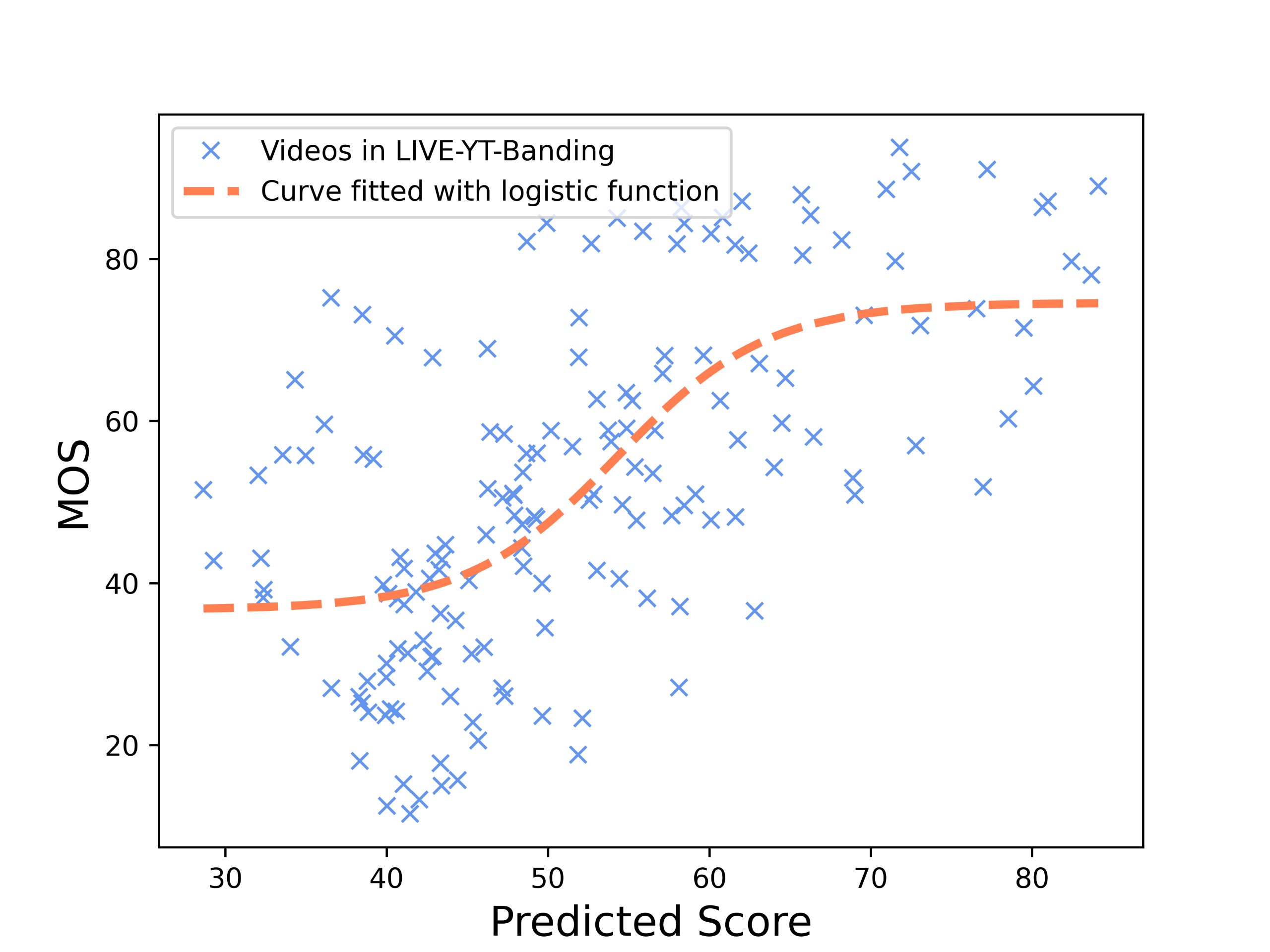}}
\hspace{-3mm} 
\subfigure[NIQE]{\includegraphics[width=2.96cm,height=1.7cm]{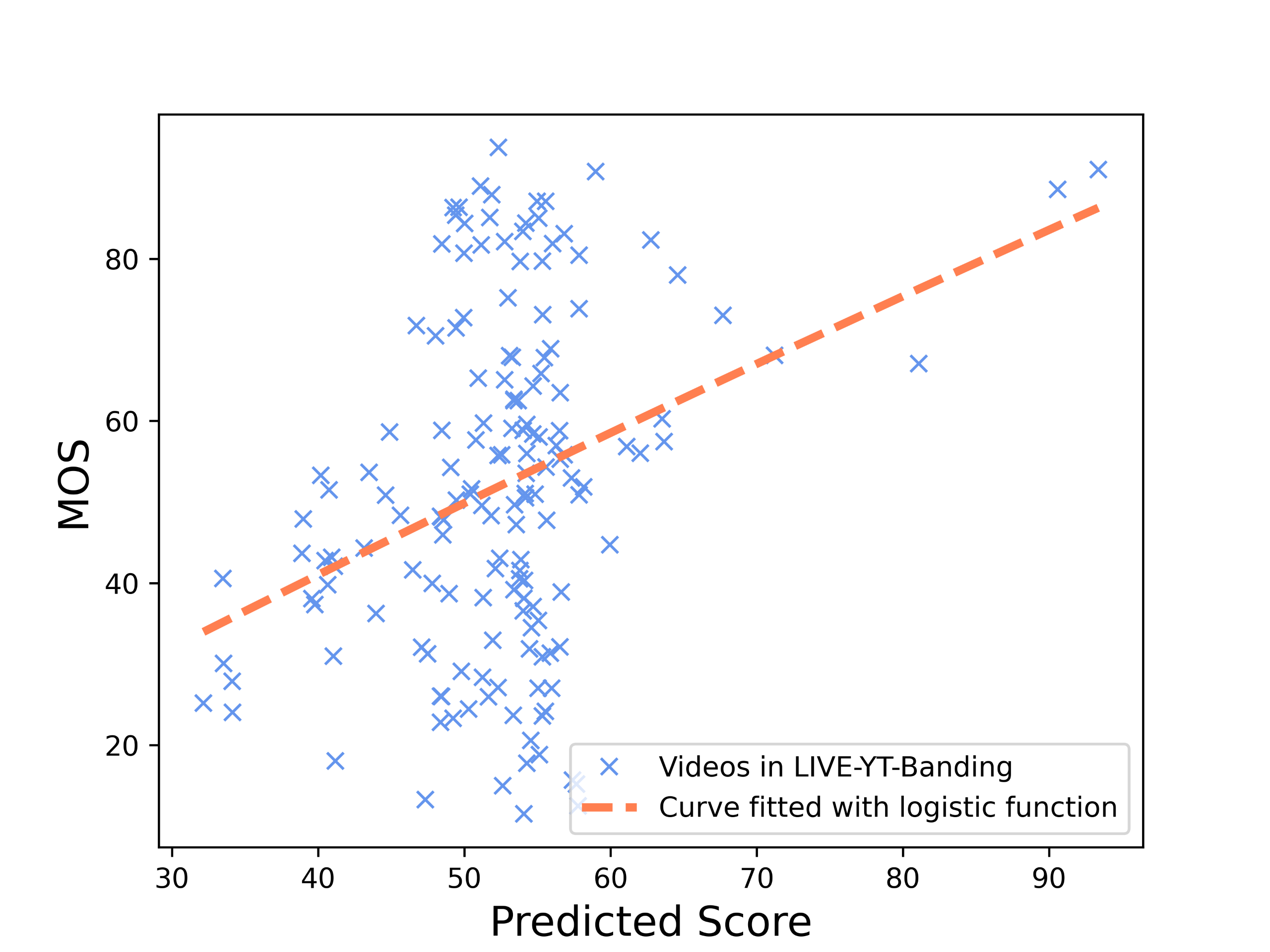}}
\hspace{-3mm} 
\subfigure[FRIQUEE]{\includegraphics[width=2.96cm,height=1.7cm]{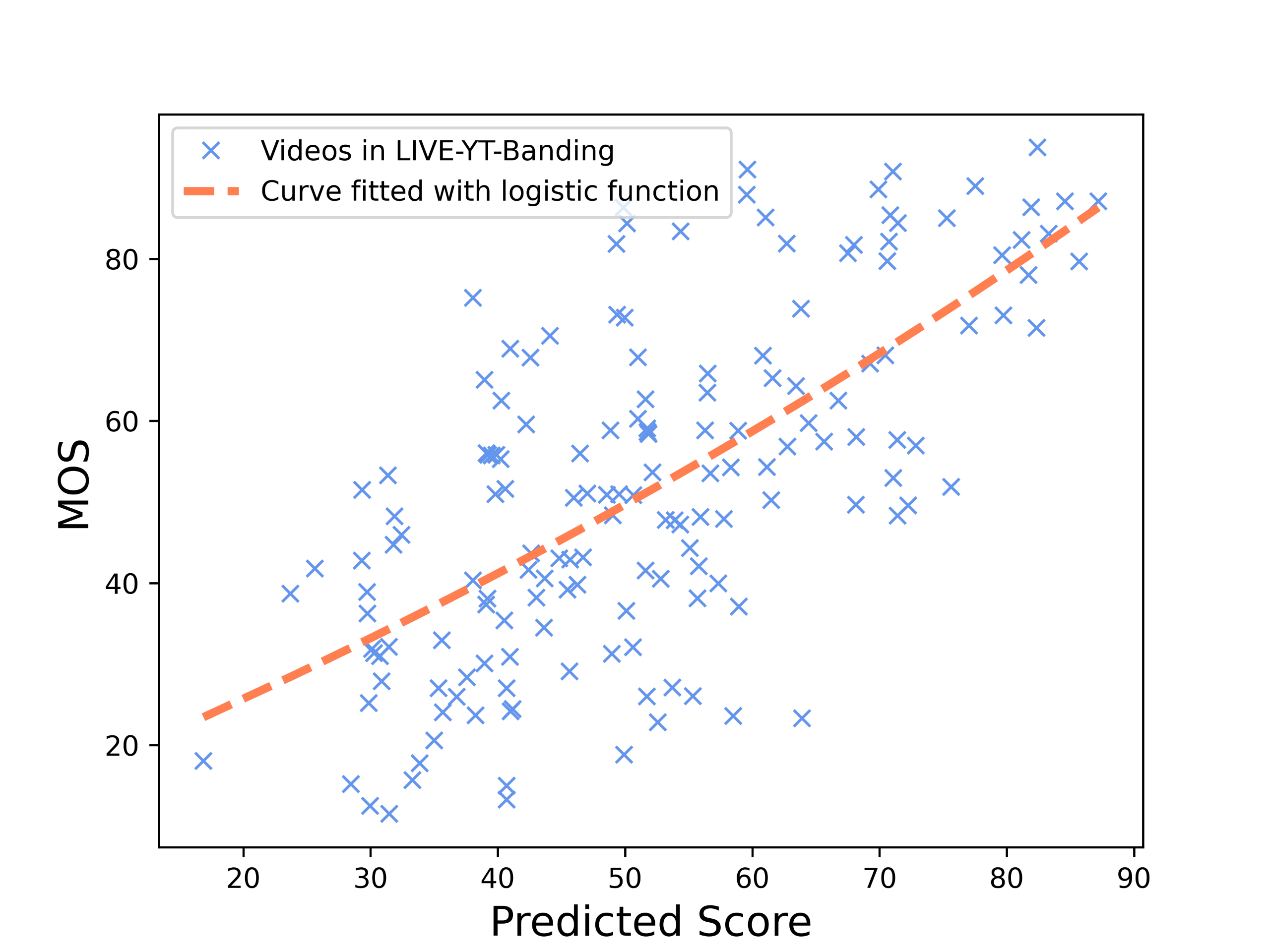}}
\hspace{-3mm} 
\subfigure[HOSA]{\includegraphics[width=2.96cm,height=1.7cm]{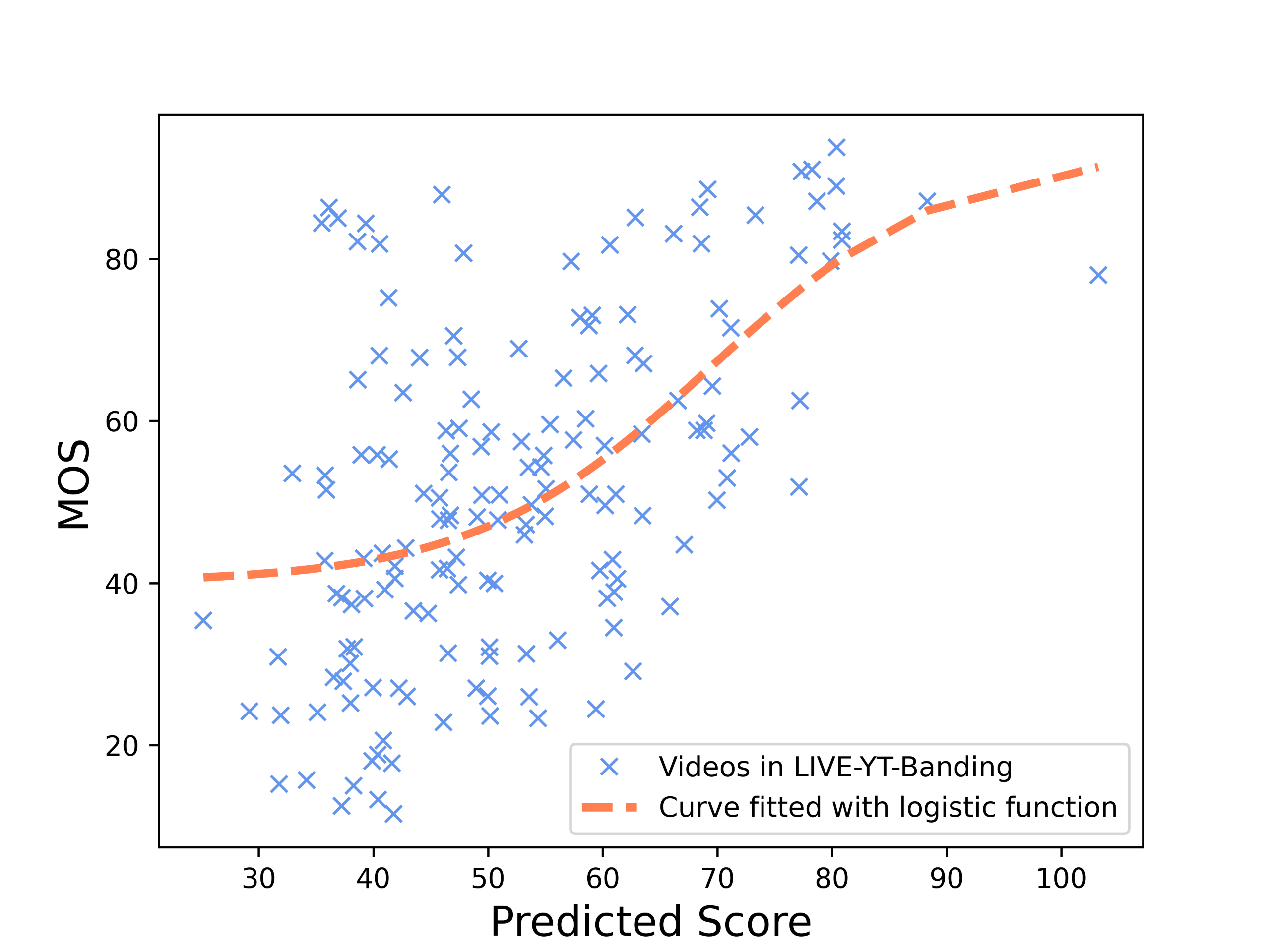}}
\\ [-2.9ex]
\subfigure[CORNIA]{\includegraphics[width=2.96cm,height=1.7cm]{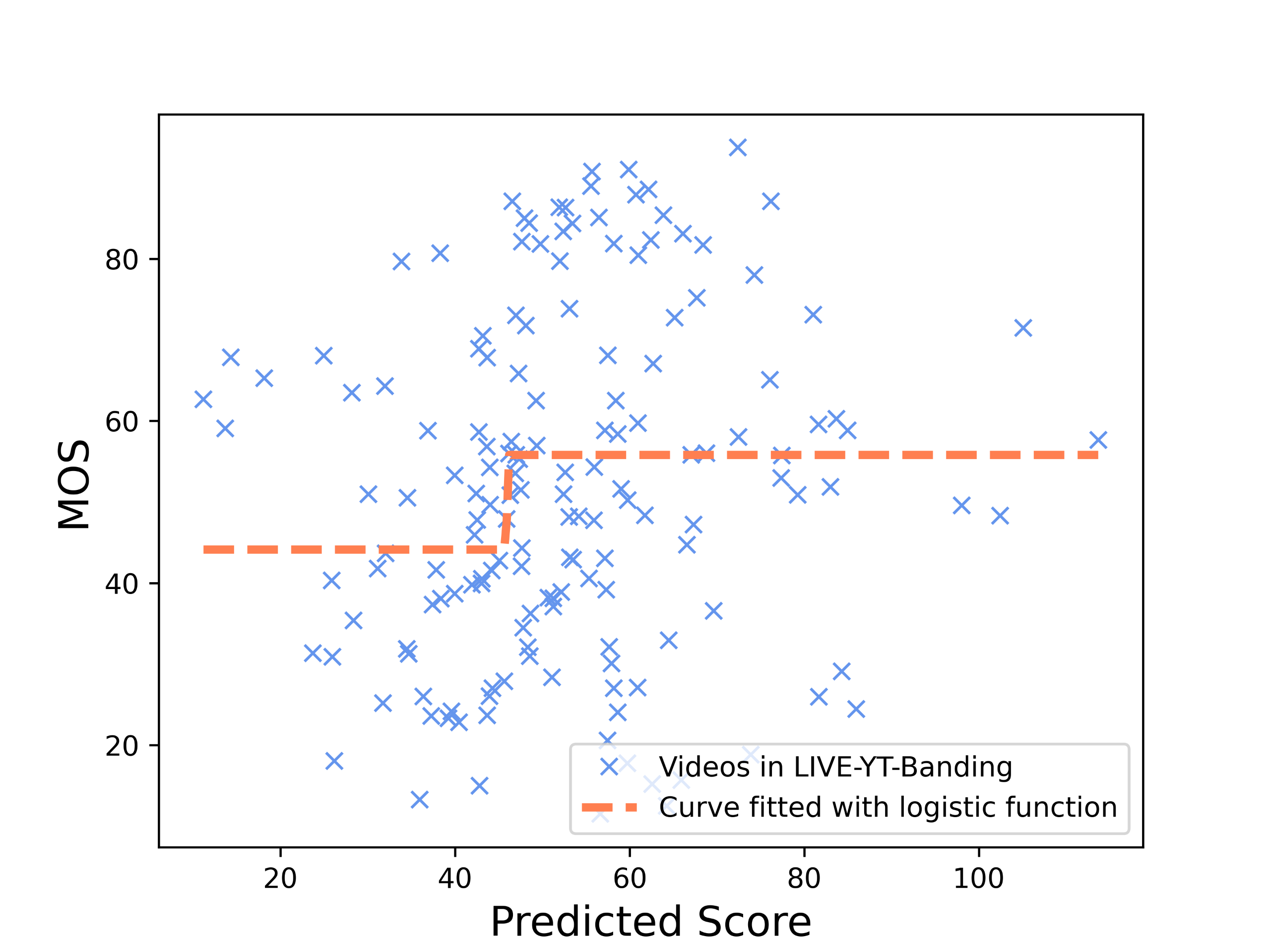}}
\hspace{-3mm} 
\subfigure[VIDEVAL]{\includegraphics[width=2.96cm,height=1.7cm]{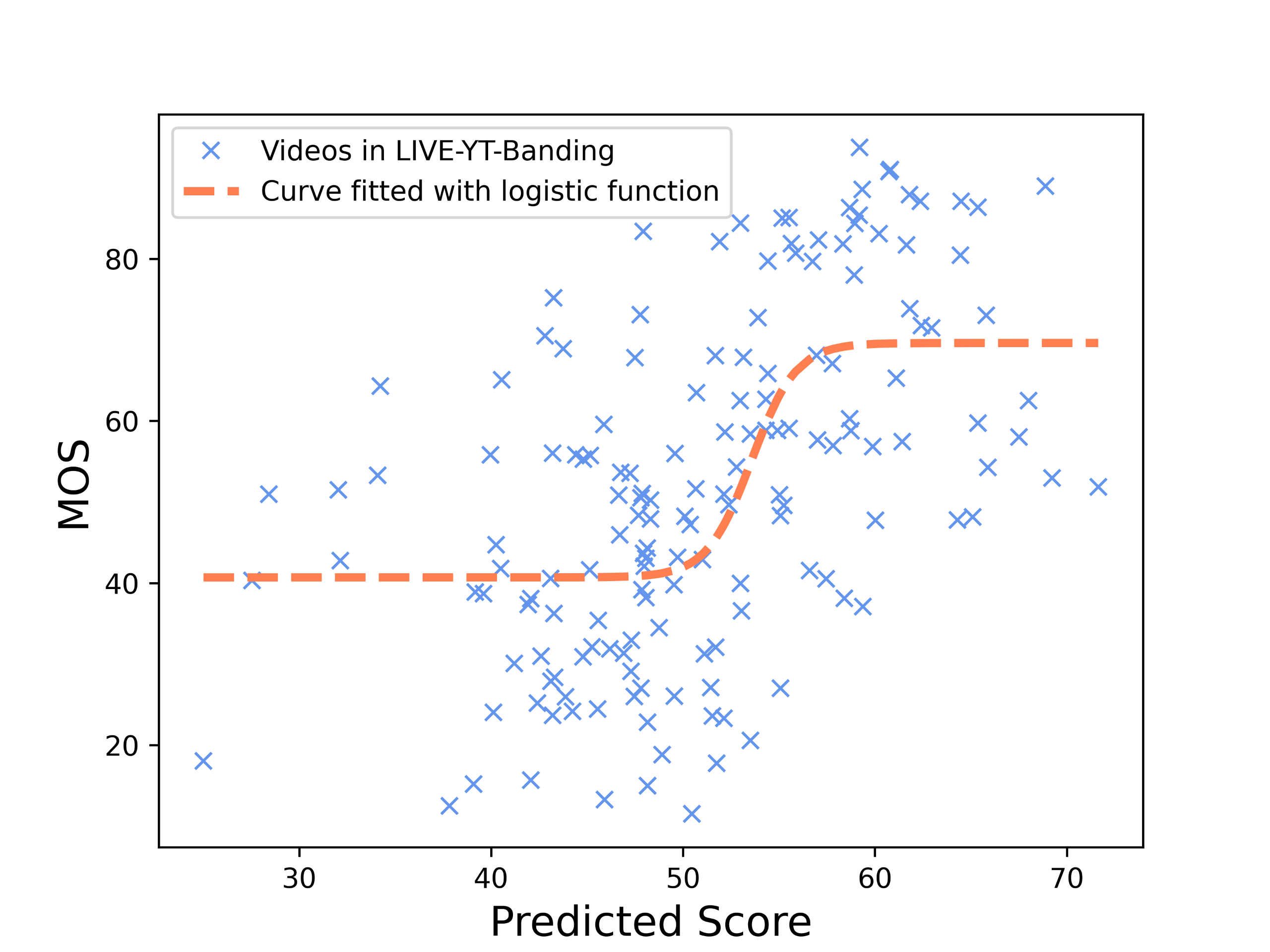}}
\hspace{-3mm} 
\subfigure[TLVQM]{\includegraphics[width=2.96cm,height=1.7cm]{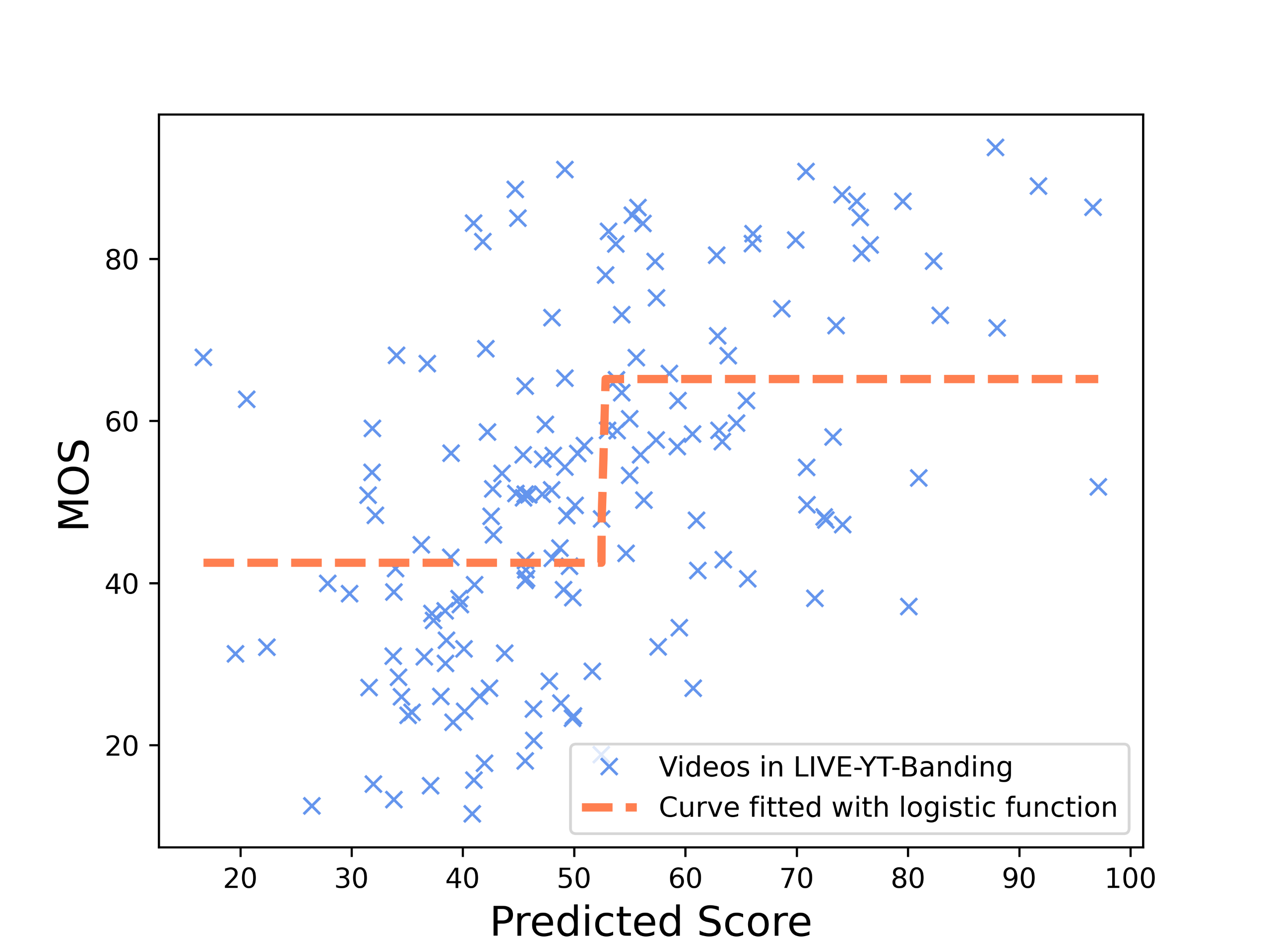}}
\hspace{-3mm} 
\subfigure[FAVER]{\includegraphics[width=2.96cm,height=1.7cm]{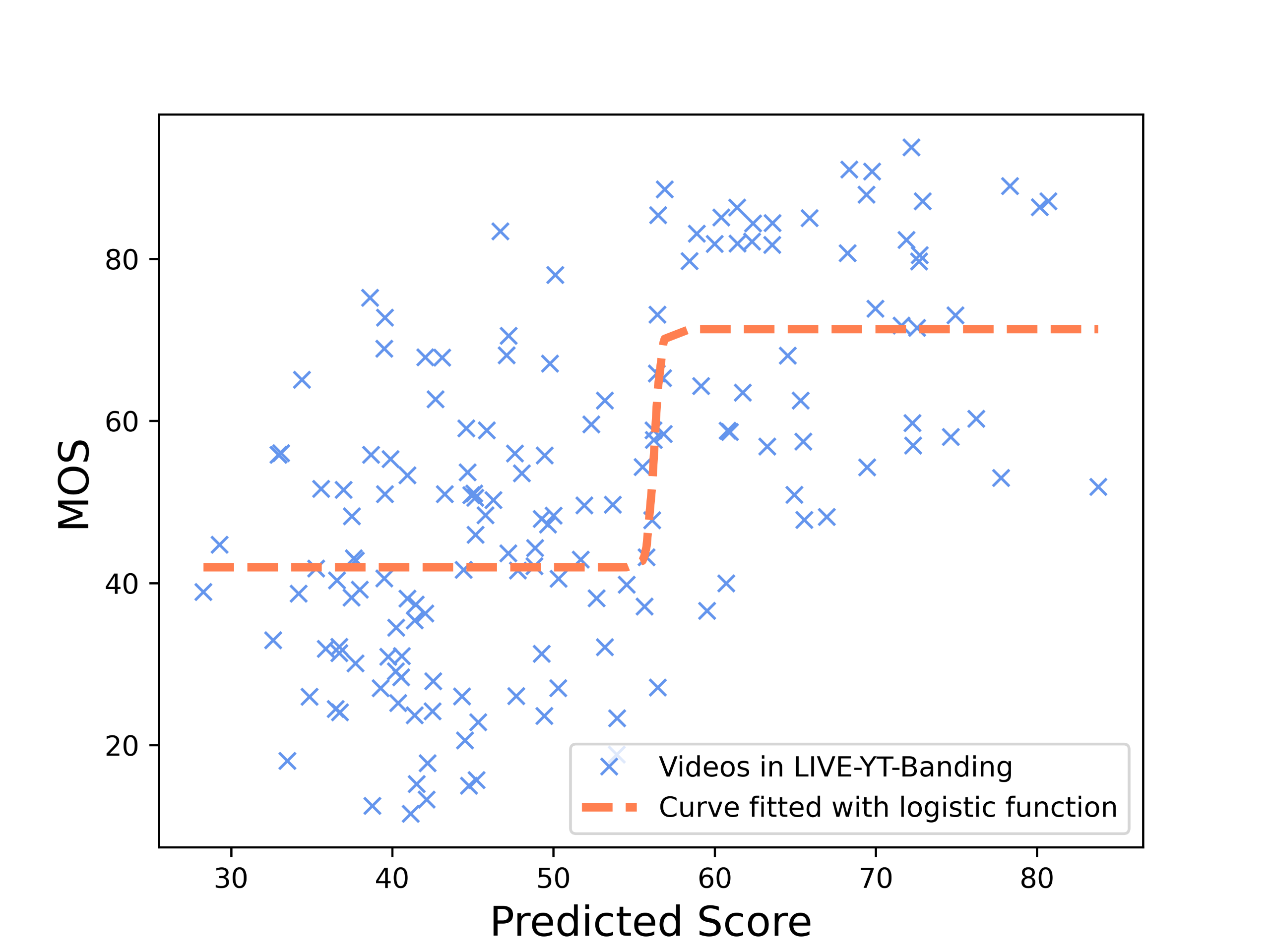}}
\hspace{-3mm} 
\subfigure[RAPIQUE]{\includegraphics[width=2.96cm,height=1.7cm]{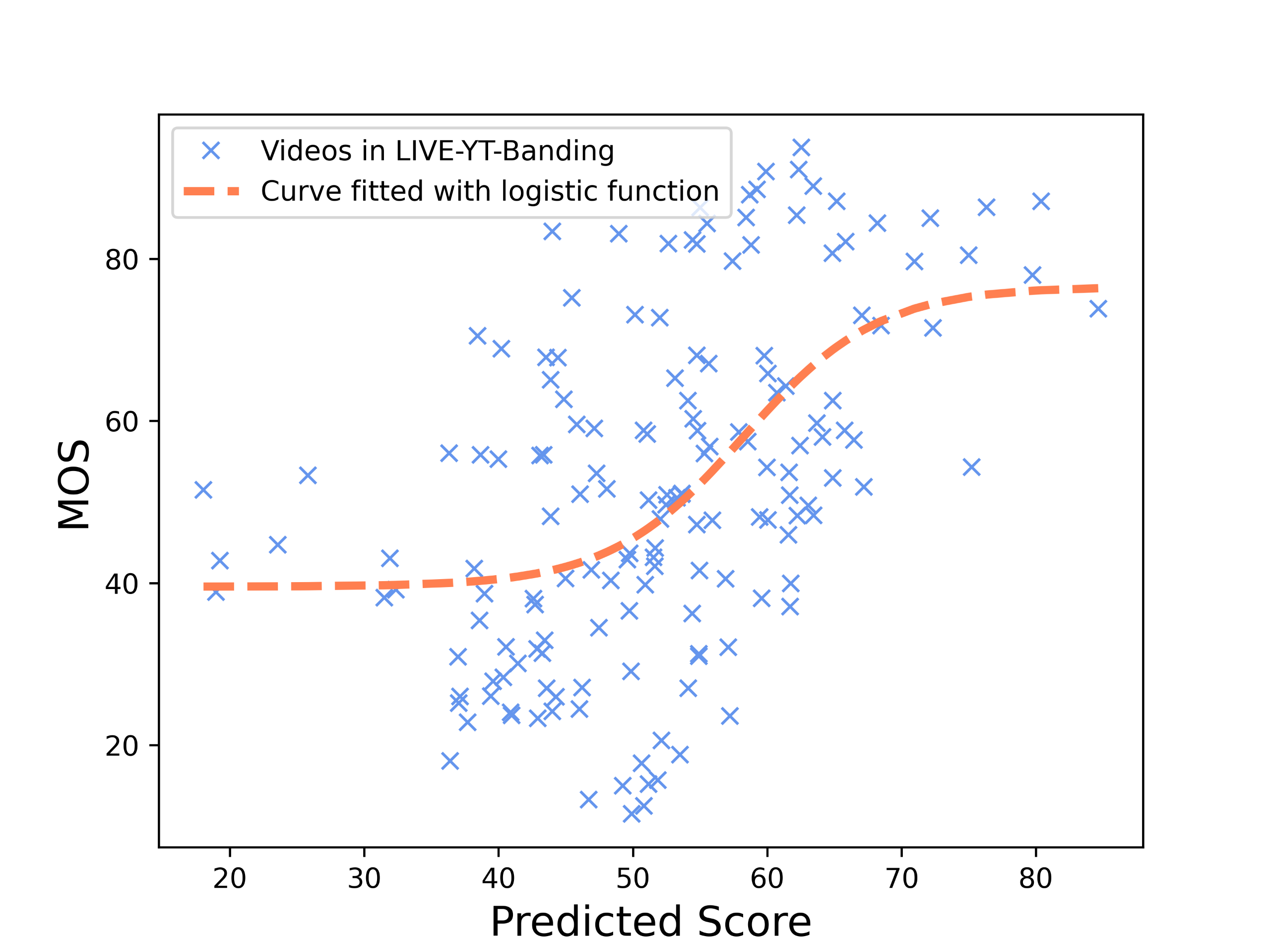}}
\\ [-2.9ex]
\subfigure[VSFA]{\includegraphics[width=2.96cm,height=1.7cm]{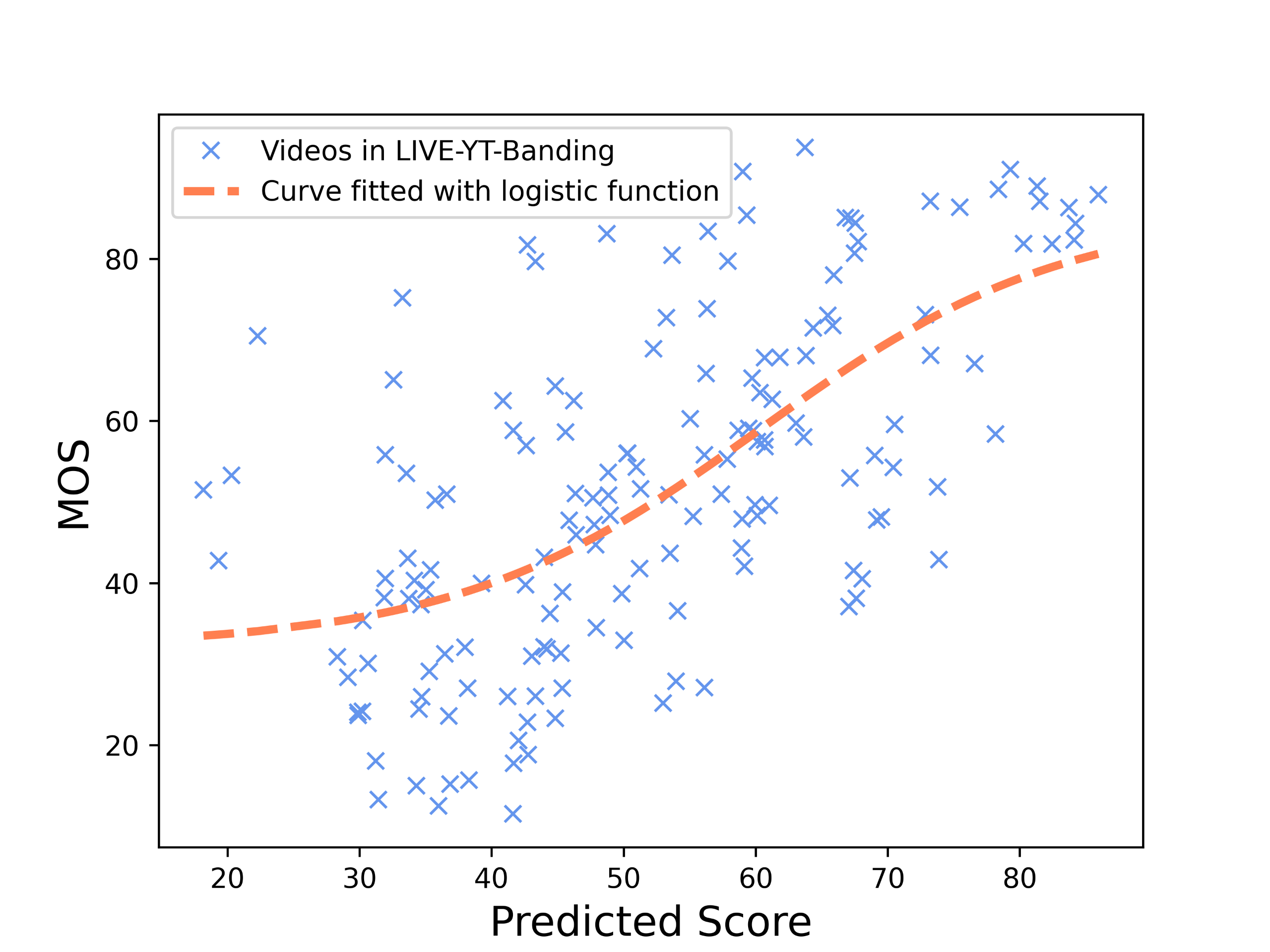}}
\hspace{-3mm} 
\subfigure[$\text{VMAF}_\text{BA}$]{\includegraphics[width=2.96cm,height=1.7cm]{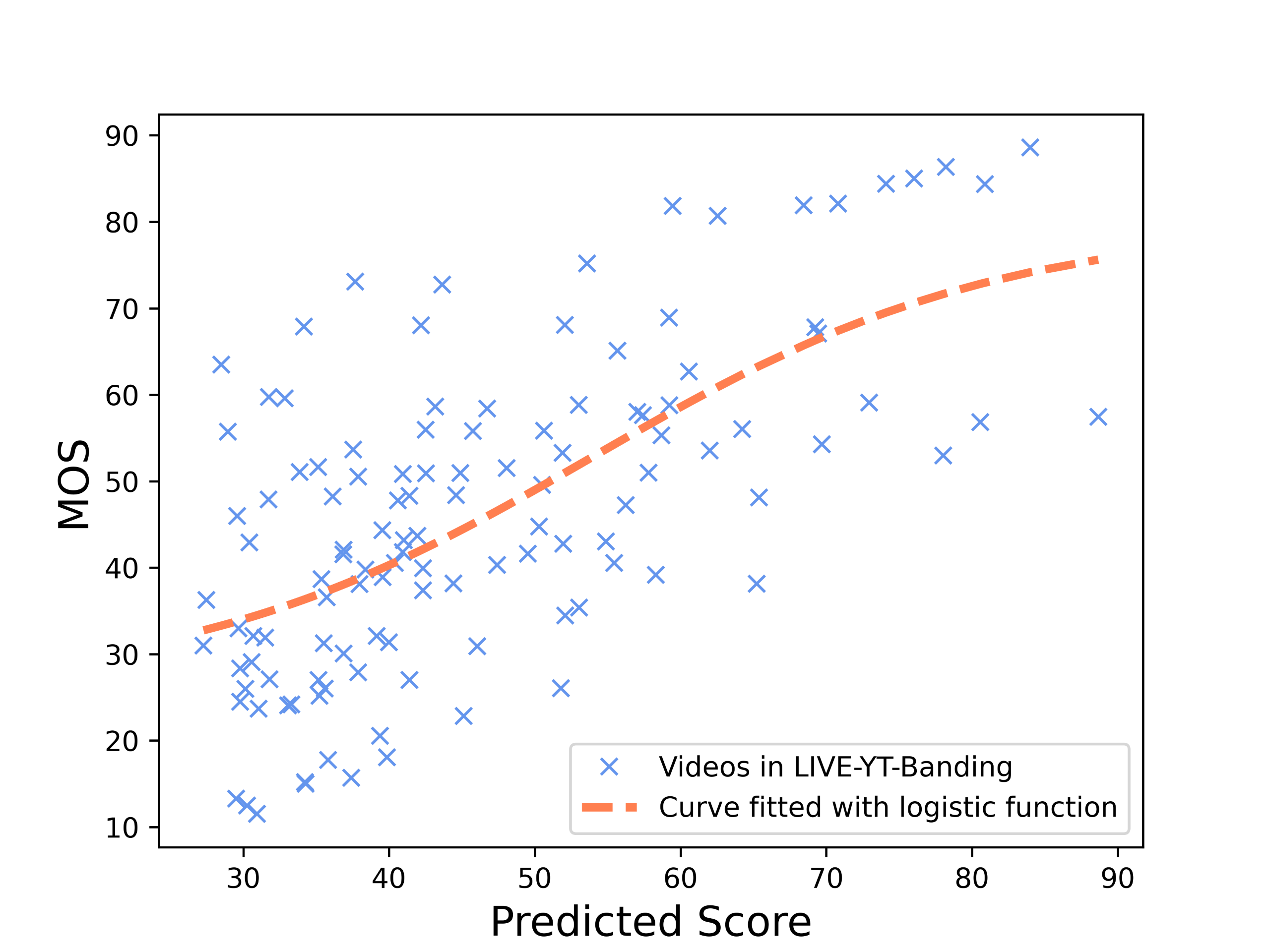}}
\hspace{-3mm} 
\subfigure[DBI]{\includegraphics[width=2.96cm,height=1.7cm]{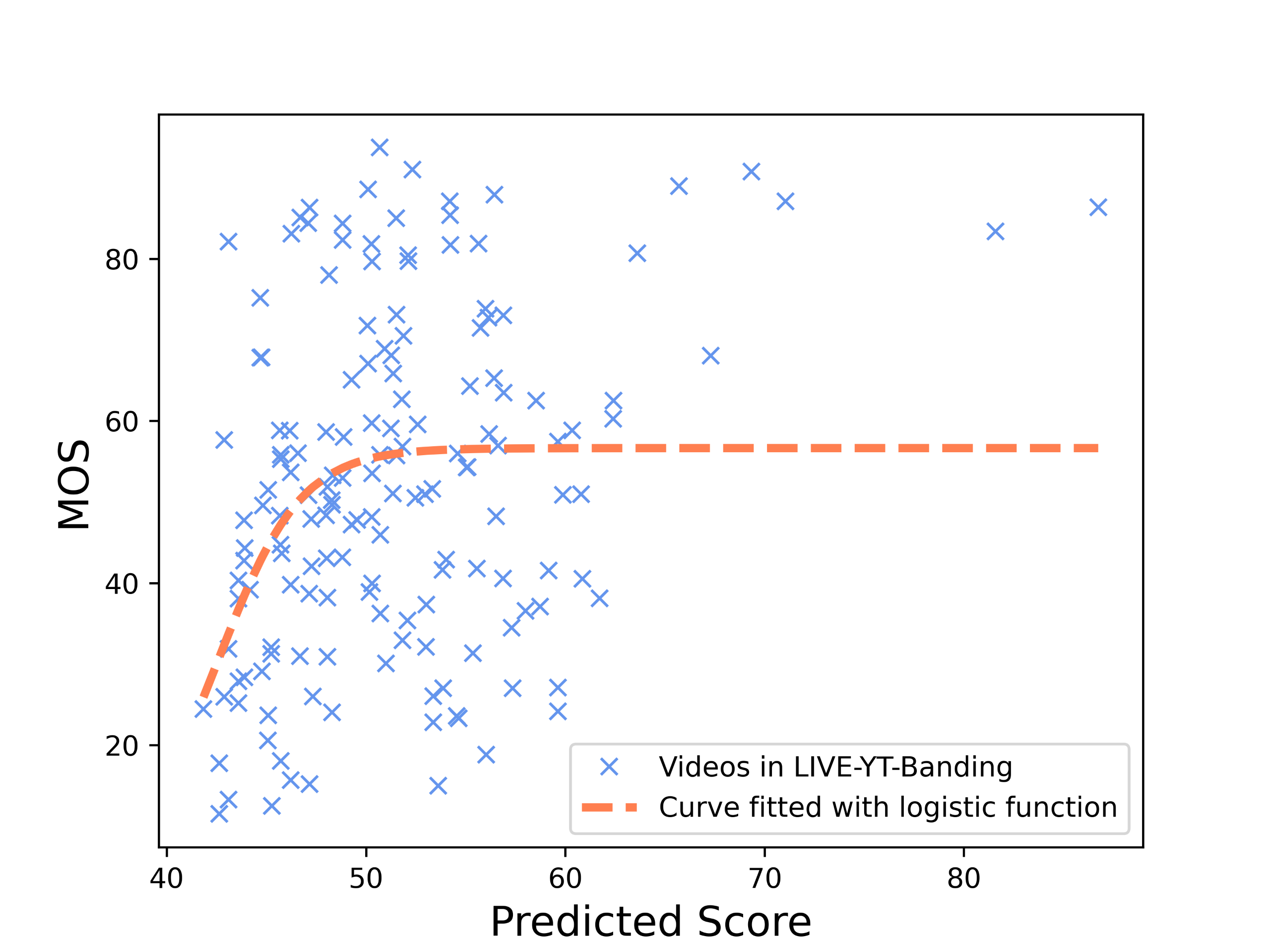}}
\hspace{-3mm} 
\subfigure[BBAND]{\includegraphics[width=2.96cm,height=1.7cm]{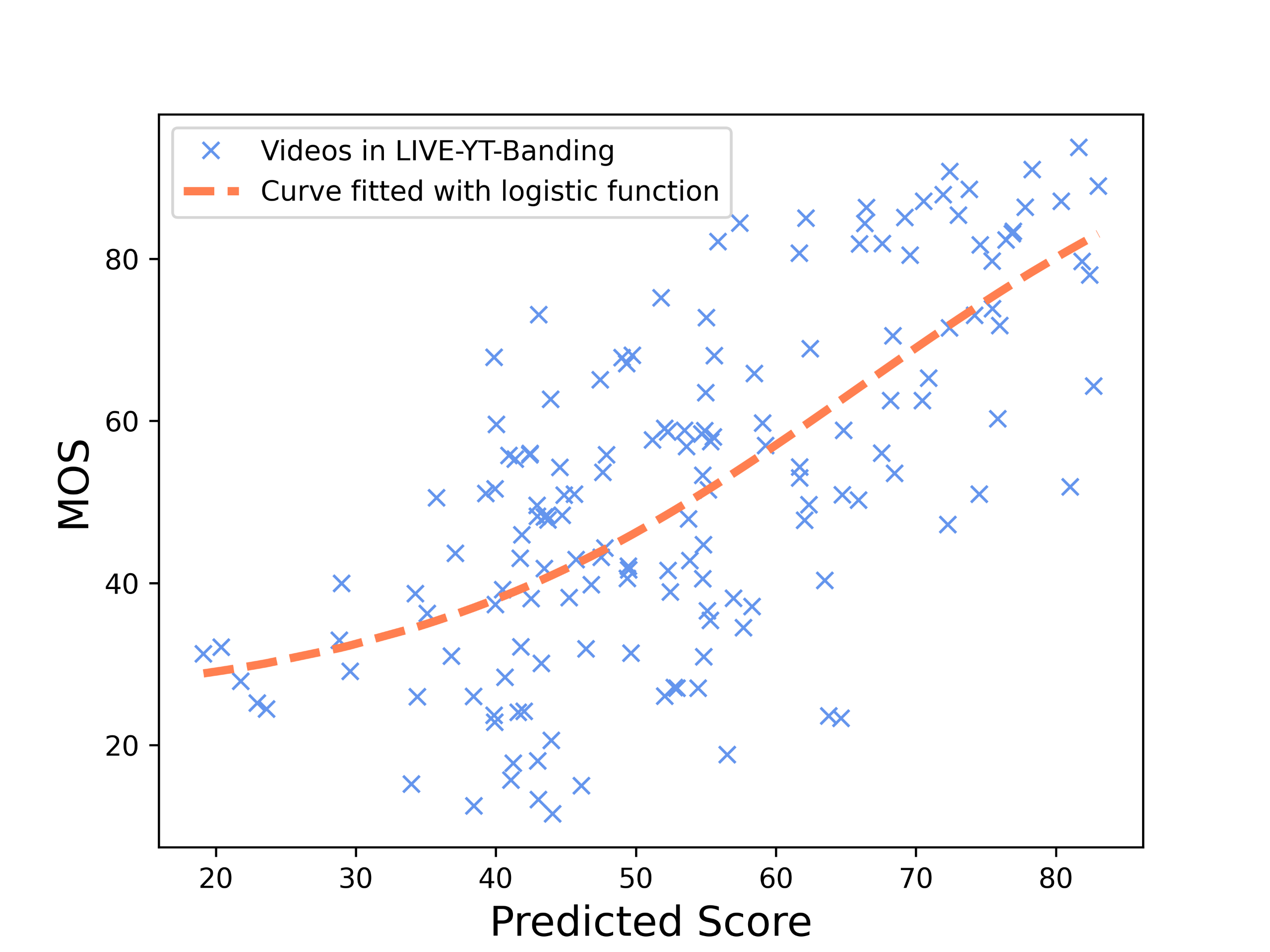}}
\hspace{-3mm} 
\subfigure[CAMBI]{\includegraphics[width=2.96cm,height=1.7cm]{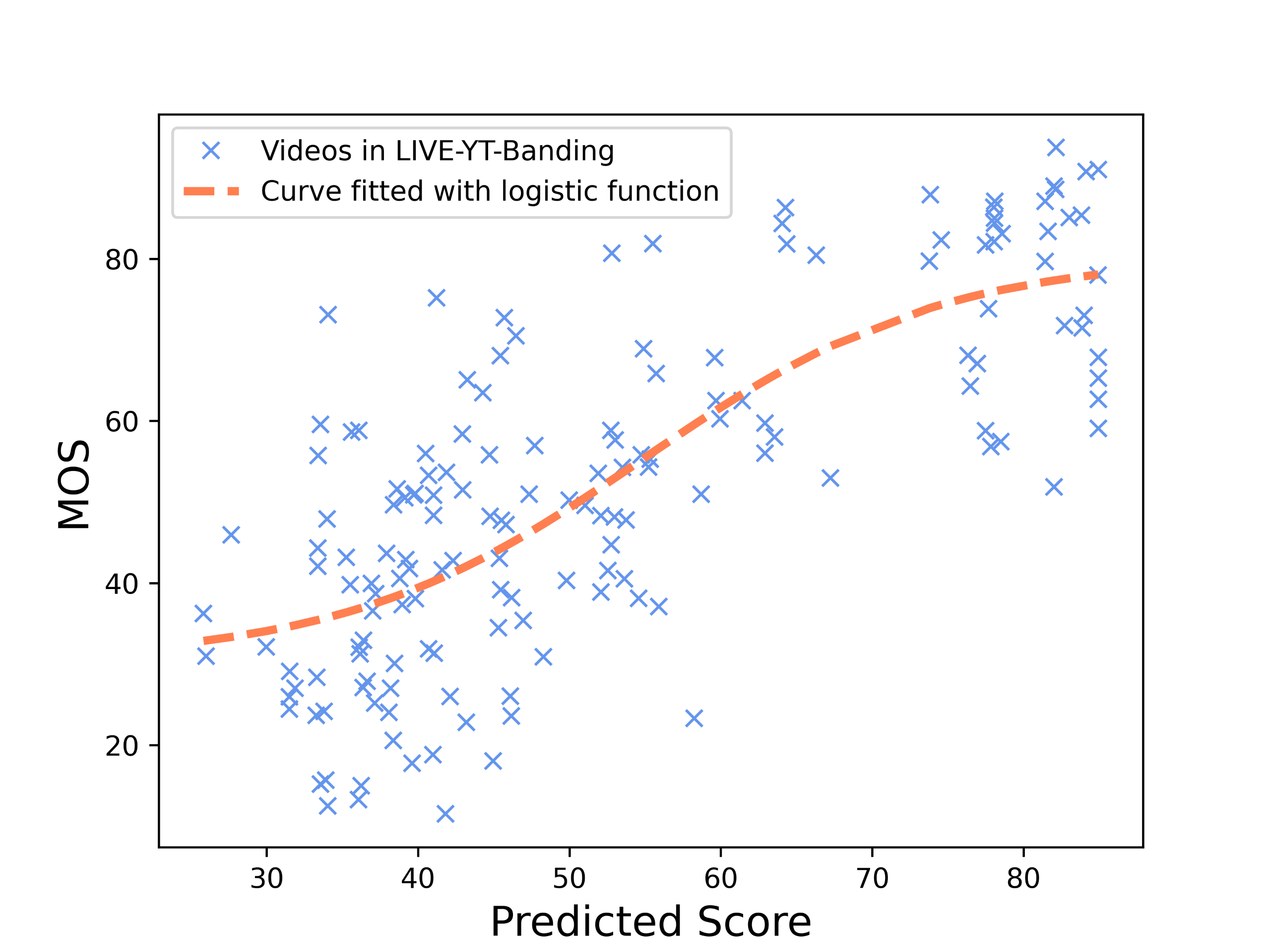}}
\\[-2.9ex]
\subfigure[FAST-VQA]{\includegraphics[width=2.96cm,height=1.7cm]{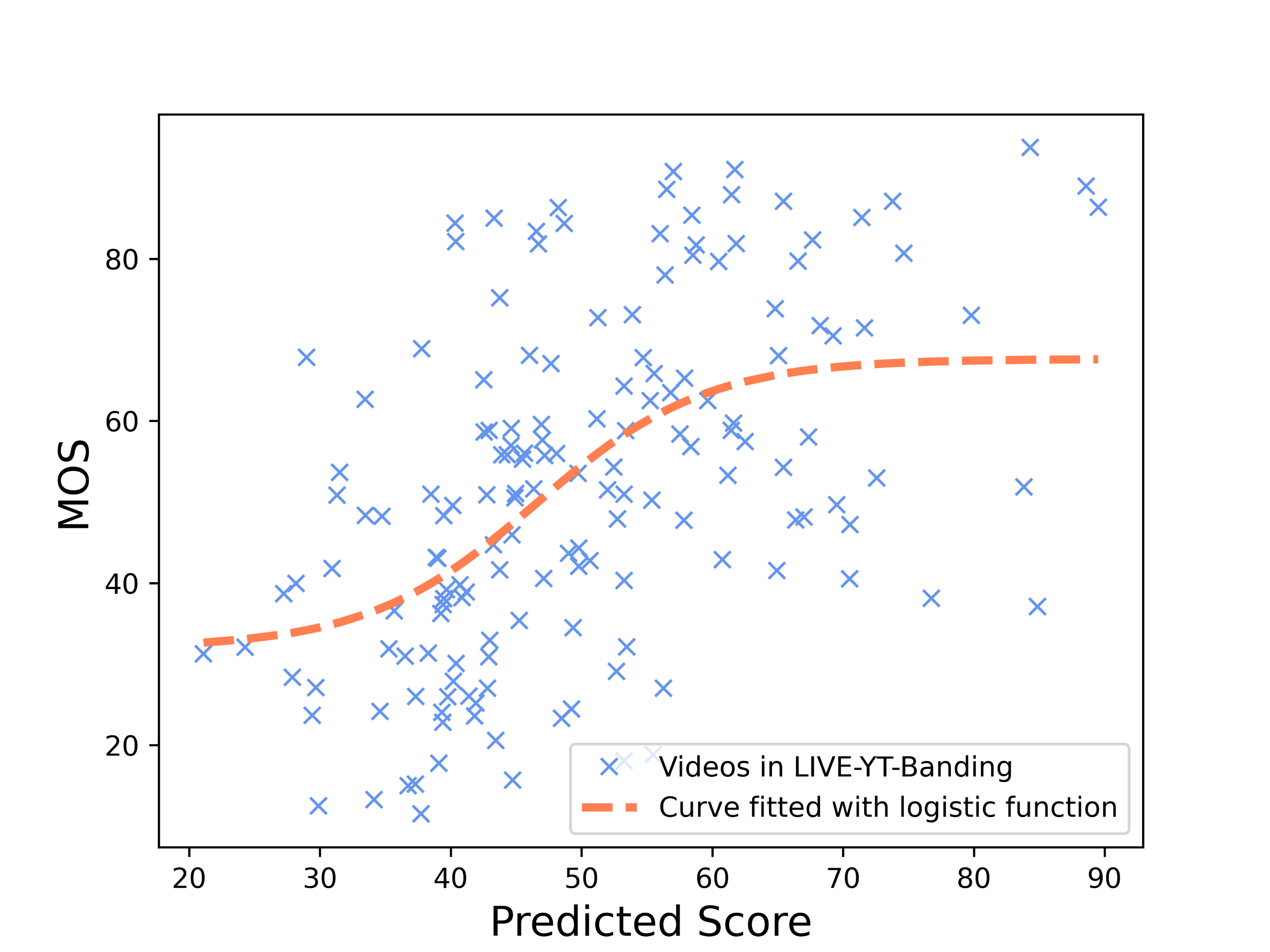}}
\hspace{-3mm} 
\subfigure[FasterVQA]{\includegraphics[width=2.96cm,height=1.7cm]{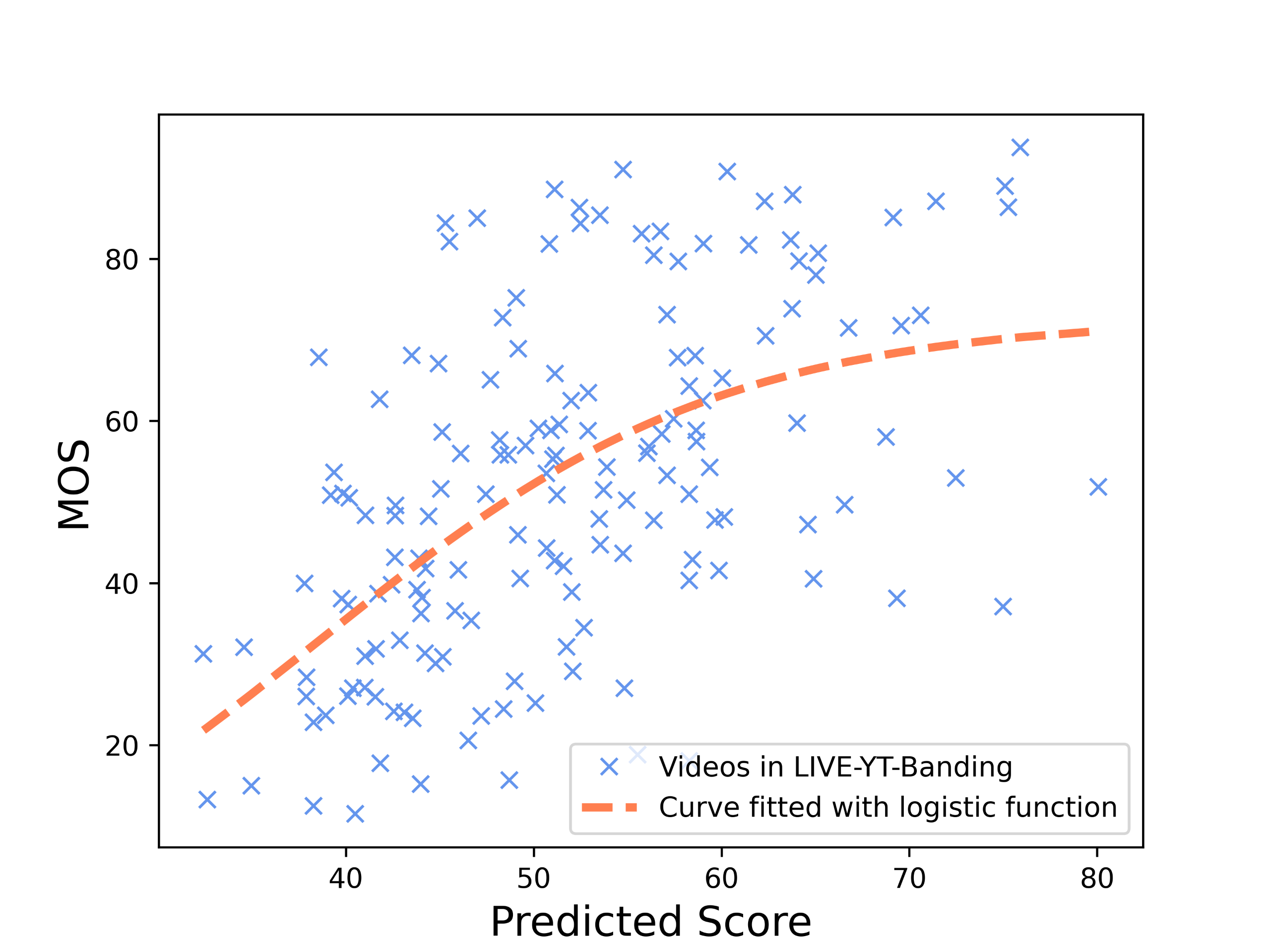}}
\hspace{-3mm} 
\subfigure[DOVER]{\includegraphics[width=2.96cm,height=1.7cm]{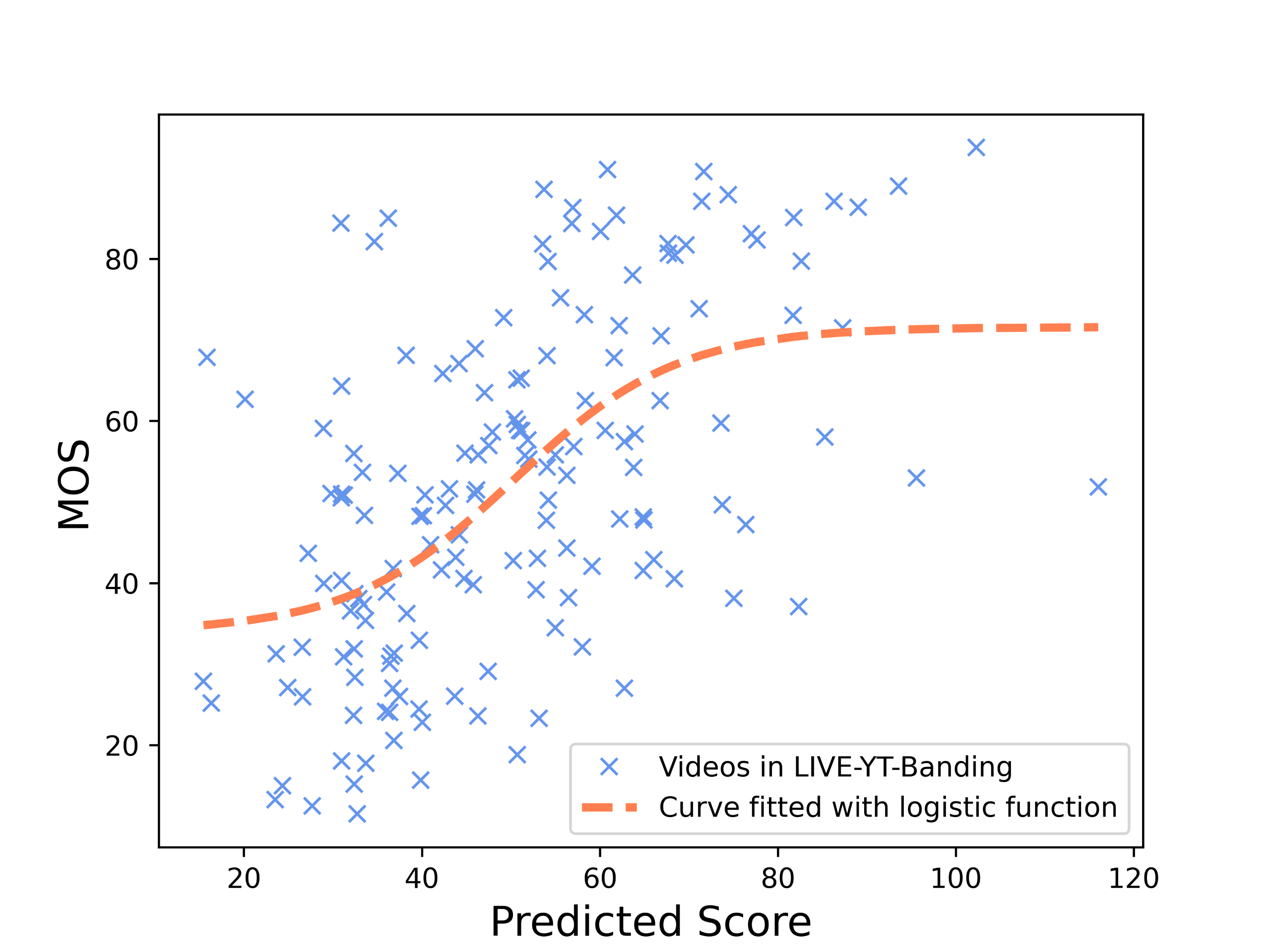}}
\hspace{-3mm} 
\subfigure[SAMA]{\includegraphics[width=2.96cm,height=1.7cm]{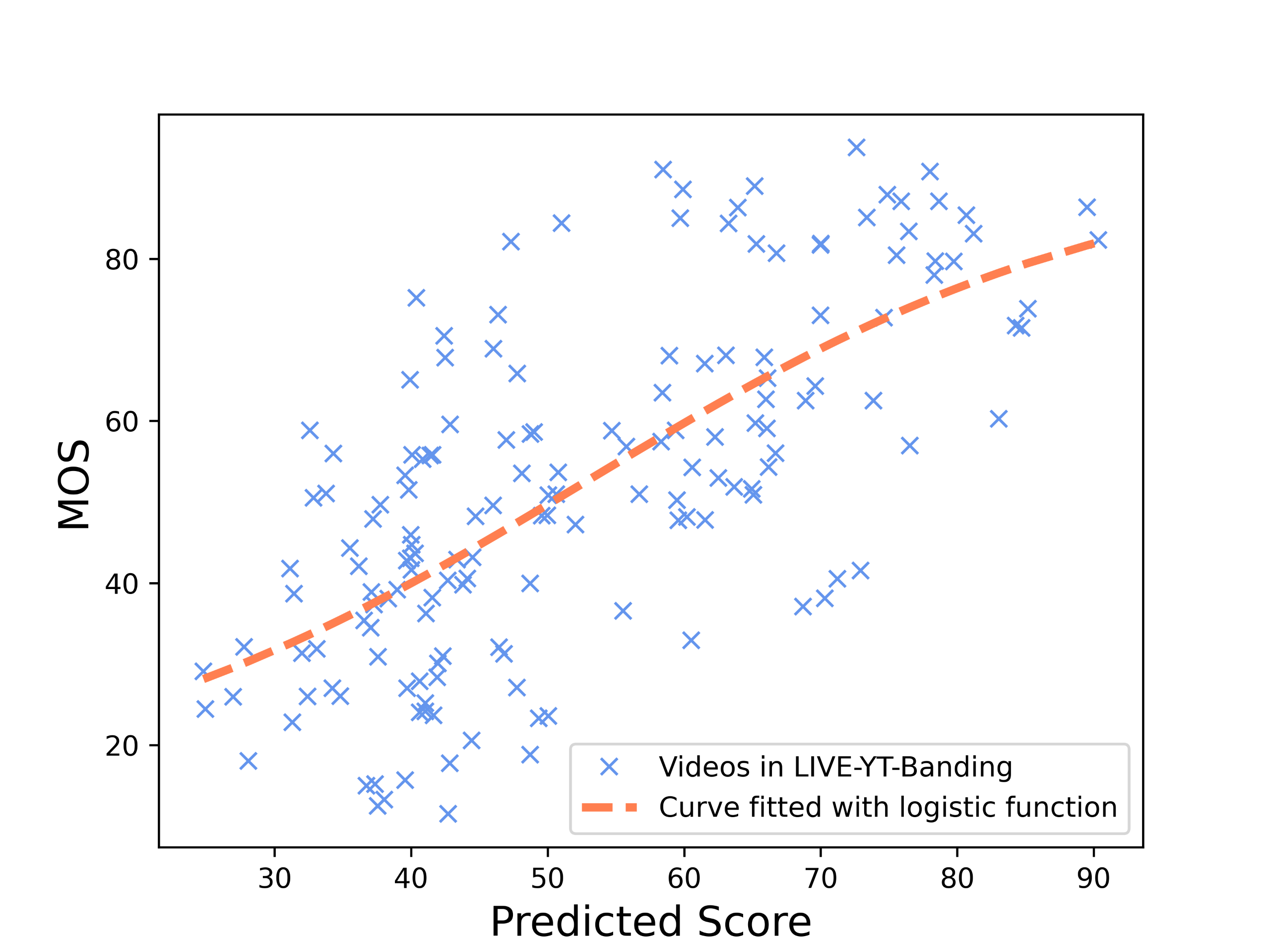}}
\hspace{-3mm} 
\subfigure[ModularBVQA]{\includegraphics[width=2.96cm,height=1.7cm]{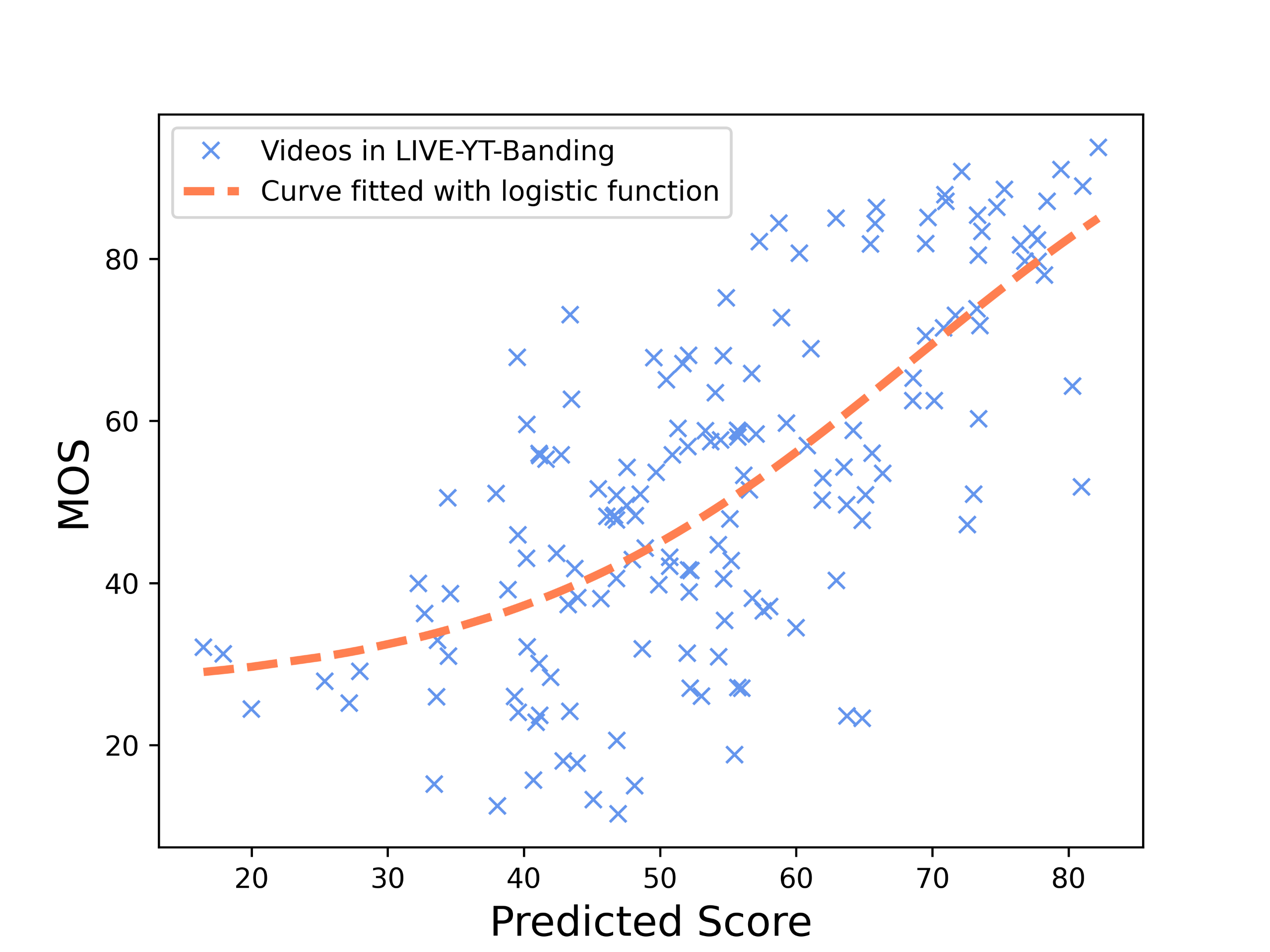}}
\\[-2.9ex]
\renewcommand{\thesubfigure}{(z-1)}
\subfigure[CBAND-VGG16]{\includegraphics[width=2.96cm,height=1.7cm]{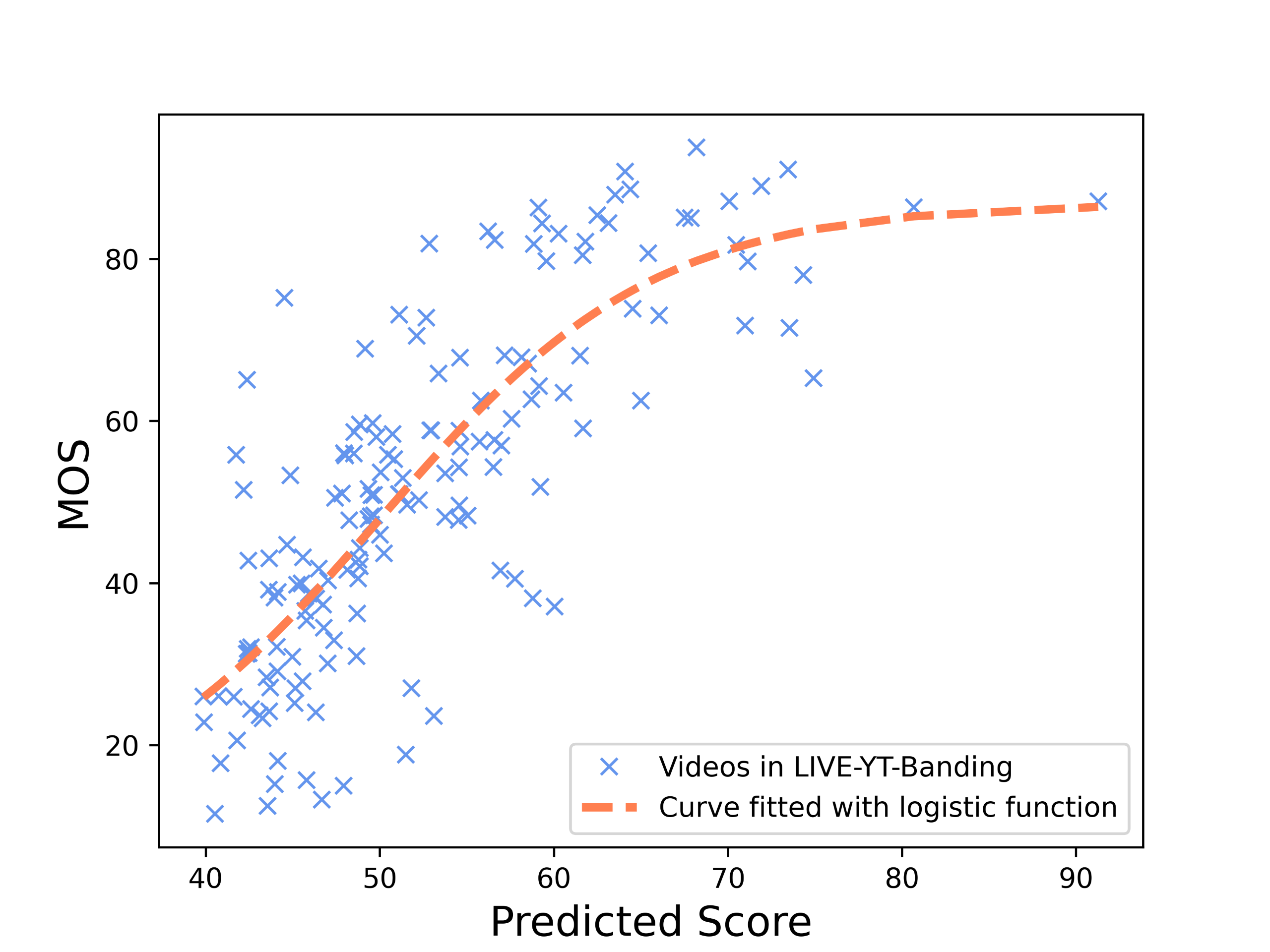}}
\renewcommand{\thesubfigure}{(z-2)}
\subfigure[CBAND-RN50]{\includegraphics[width=2.96cm,height=1.7cm]{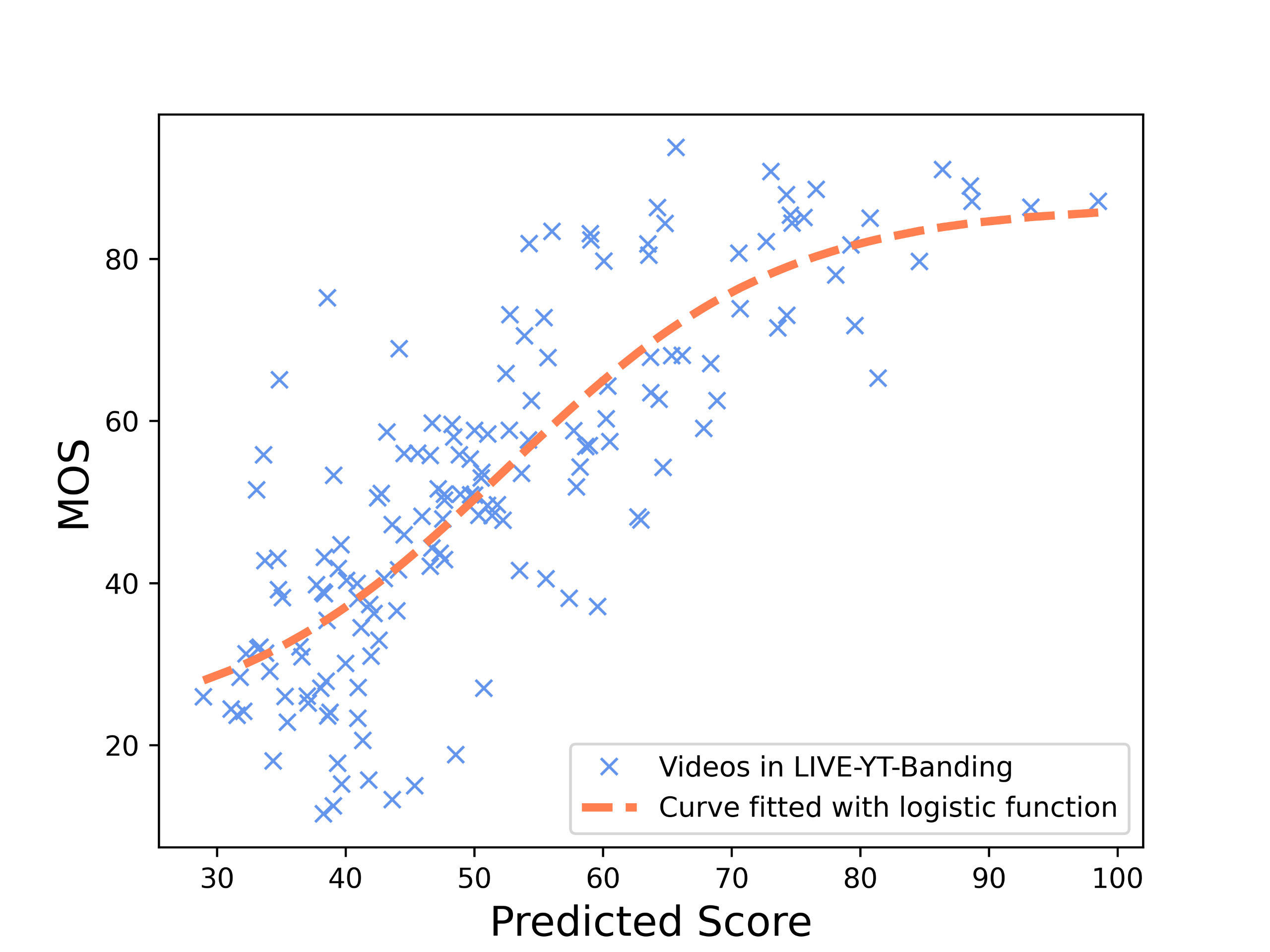}}
\renewcommand{\thesubfigure}{\alph{subfigure}}
\\[-1ex]
\vspace{-3pt}
\caption{Scatter plots and logistic fitted curves of predictions versus MOS on the LIVE-YT-Banding database for all evaluated models.}
\vspace{-15pt}
\label{fig:scatter}
\end{figure*}

\subsection{Main Comparison}

\begin{table}[!t]
\renewcommand{\arraystretch}{1.2}
\centering
\fontsize{6pt}{7pt}\selectfont
\setlength{\tabcolsep}{1.5pt}
\caption{Performance comparison of evaluated models on LIVE-YT-Banding dataset. Italics denote FR methods. Boldfaced entries indicate top performers and underlined entries indicate the second and third performers.}
\vspace{-5pt}
\label{table:performance-all-model}
\begin{tabular}{l|l|cccc}
\toprule
& {\textbf{Model}}                    &  \textbf{SROCC↑}                      &  
\textbf{KROCC↑}    &
\textbf{PLCC↑}                       & \textbf{RMSE↓}                       \\ 
\midrule
\multirow{16}{*}{\begin{tabular}[c]{@{}c@{}}General \\  IQA/VQA \\ methods \end{tabular}} & 
\emph{PNSR}                                      &   0.1265             &    0.0739     &   0.3879    &  14.9923                 \\ &

\emph{SSIM}~\cite{wang2004image}                                           &    0.2103            & 0.1290        &   0.4376    &   14.6145              \\ &

\emph{LPIPS}~\cite{zhang2018unreasonable}                                           &     -0.0587           &  -0.0378       &  0.3114     &   15.6814              \\ &

\emph{VMAF}~\cite{vmaf}          &        0.3394        &   0.2478      &   0.4537    &   14.6145          \\ 
\cline{2-6}

 &
BRISQUE~\cite{mittal2012no}                                     &    0.6593           &    0.4996   &  0.7383   &      13.0672           \\ &

GM-LOG~\cite{xue2014blind}                                           &    0.5844            &   0.4250    &   0.7081   &     13.8046            \\ &

HIGRADE~\cite{kundu2017no}                                         &  0.5880             &   0.4259     &    0.6908  &       13.9232           \\ &

NIQE~\cite{6353522}         &   0.2669            &   0.1983    &   0.4251  & 17.7873                \\ &
FRIQUEE~\cite{FRIQUEE}                            &   0.6142            & 0.4497      &  0.7092   &          13.7676       \\ &
HOSA~\cite{7501619}                                     &    0.4134           &  0.2992     &  0.5556   &        16.1725         \\ &

CORNIA~\cite{6247789}                                           &      0.1169         &   0.0902     &  0.3899   &       18.1365          \\ &

VIDEVAL~\cite{9405420}                                         &     0.5531           &       0.4004 &   0.6314   &      15.0006            \\ &

TLVQM~\cite{korhonen2019two}         &    0.5023           &  0.3619     &    0.5804  &   15.9058              \\ &
FAVER~\cite{zheng2024faver}                            &  0.5836             &  0.4142     & 0.6936    &     13.9952            \\ &
RAPIQUE~\cite{tu2021rapique}                            &   0.5317             &  0.3752     &  0.6318   &   15.0952               \\ &
VSFA~\cite{li2019quality}                            &  0.7284             &    \underline{0.5766}    &    0.7381 & 14.0313              \\ 
& FAST-VQA~\cite{10.1007/978-3-031-20068-7_31} & 0.4578& 0.2782& 0.7453& 15.0513 \\
& FasterVQA~\cite{10264158} & 0.4949 & 0.3266& 0.7356 & 15.2913  \\
& DOVER~\cite{Wu_2023_ICCV} & 0.6924&0.5242 & 0.7360&17.2289 \\
& SAMA~\cite{Liu_Quan_Xiao_Li_Wu_2024} & 0.4736&0.3508 & 0.6437& 15.1323 \\
& ModularBVQA~\cite{Wen_2024_CVPR} &\underline{0.7295} &0.5484 &0.7156 &13.7488  \\
\midrule
\multirow{4}{*}{\begin{tabular}[c]{@{}c@{}}Banding \\  IQA/VQA \\ methods \end{tabular}}& \emph{$\text{VMAF}_{\text{BA}}$}~\cite{krasula2022banding}                                  &    0.6274            &   0.4836      &  0.6888     & \underline{11.2609}       \\ \cline{2-6}
&
DBI~\cite{kapoor2021capturing}                            &   0.3412            &   0.2331    &   0.3721  &      19.3244           \\ &
BBAND~\cite{tu2020bband}                            &   0.6648            &    0.4905    &  0.7213    &     13.5532            \\ &
CAMBI~\cite{tandon2021cambi}                           &   0.7143            &  0.5364     &  \underline{0.7658}    &   13.3496              \\ 
\midrule
\multirow{2}{*}{Proposed} & CBAND-VGG16              &  \underline{0.7797}              &   \underline{0.6048}      &  \underline{0.8006}     &     \underline{11.5938}         \\ & 
CBAND-RN50                &  \textbf{0.8012}              &   \textbf{0.6289}      &   \textbf{0.8287}    &    \textbf{10.6743}   \\                  
\bottomrule
\end{tabular}
\vspace{-15pt}
\end{table}

We report in Table~\ref{table:performance-all-model} the performance outcomes of all the compared general-purpose and banding-specific IQA/VQA models on the LIVE-YT-Banding Dataset.
We tested the two previously defined CBAND variants: CBAND-VGG16 and CBAND-RN50.
From Table~\ref{table:performance-all-model}, we present a detailed analysis of how different categories of IQA/VQA models perform in assessing banding artifacts. This analysis highlights the relative strengths and limitations of various approaches.

\textbf{General FR models} perform poorly in assessing banding artifacts. The best performer, VMAF, achieves only 0.3394 SROCC. This is because banding artifacts appear along sparsely distributed contours, while most regions remain unaffected. FR models compute global perceptual differences without focusing on banding-prone areas, leading to misalignment with human perception. Moreover, banding distortions form structured patterns that amplify visual discomfort, yet FR models rely on local pixel-level comparisons, lacking the ability to capture extended artifacts.

\textbf{General NR models} generally outperform FR models as they extract perceptual features directly from distorted content. Handcrafted NR models like BRISQUE and FRIQUEE show moderate correlation with human perception, reflecting their ability to capture low-level statistical irregularities, but still struggle with banding distortions. Recent learning-based models, such as VSFA,  DOVER, and ModularBVQA, improve performance up to an SROCC of 0.7295, leveraging deep learning techniques, spatiotemporal features, and efficient architectures. But, they remain less specialized for banding artifacts, which exhibit unique perceptual characteristics.



Among \textbf{banding-specific methods}, CAMBI performs best with an SROCC of 0.7143, leveraging heuristic-based banding detection. However, its reliance on handcrafted features places a ceiling on its performance. DBI, designed for still images, fails to generalize to videos, confirming that the statistics of video banding artifacts are fundamentally different from those arising from still-picture bit-depth reduction~\cite{tu2020adaptive}. These results emphasize the need for dedicated video banding datasets and specialized assessment models.



\textbf{The proposed CBAND models} achieve state-of-the-art performance across all key metrics, demonstrating the effectiveness of its banding-aware feature learning. CBAND-RN50 attains an SROCC of 0.8012 and a PLCC of 0.8287, while CBAND-VGG16 closely follows with an SROCC of 0.7797. Notably, CBAND-RN50 surpasses CAMBI by 12.2\% in SROCC and 8.2\% in PLCC, highlighting its ability to better capture the perceptual severity of banding artifacts. Unlike conventional banding-specific models, CBAND leverages early-stage CNN features specifically tailored to banding-sensitive regions, ensuring that fine-grained structural degradations are effectively characterized. Furthermore, CBAND surpasses the best general NR-VQA model (ModularBVQA) by 9.8\% in SROCC and 15.8\% in PLCC, demonstrating that incorporating banding-aware spatial statistics enhances video quality assessment. These results validate that CBAND not only provides a more precise assessment of banding artifacts but also establishes a new benchmark for banding-specific VQA.
Scatter plots and fitted logistic curves of model prediction scores against MOS are shown in Figure~\ref{fig:scatter}. It can be observed that, as compared to other models, predictions produced by CBAND models yielded more consistent prediction performance with a narrower prediction variance. 


\begin{figure*}[!t]
\centering
\footnotesize
\begin{tabular}{c}
\includegraphics[width=0.8\textwidth]{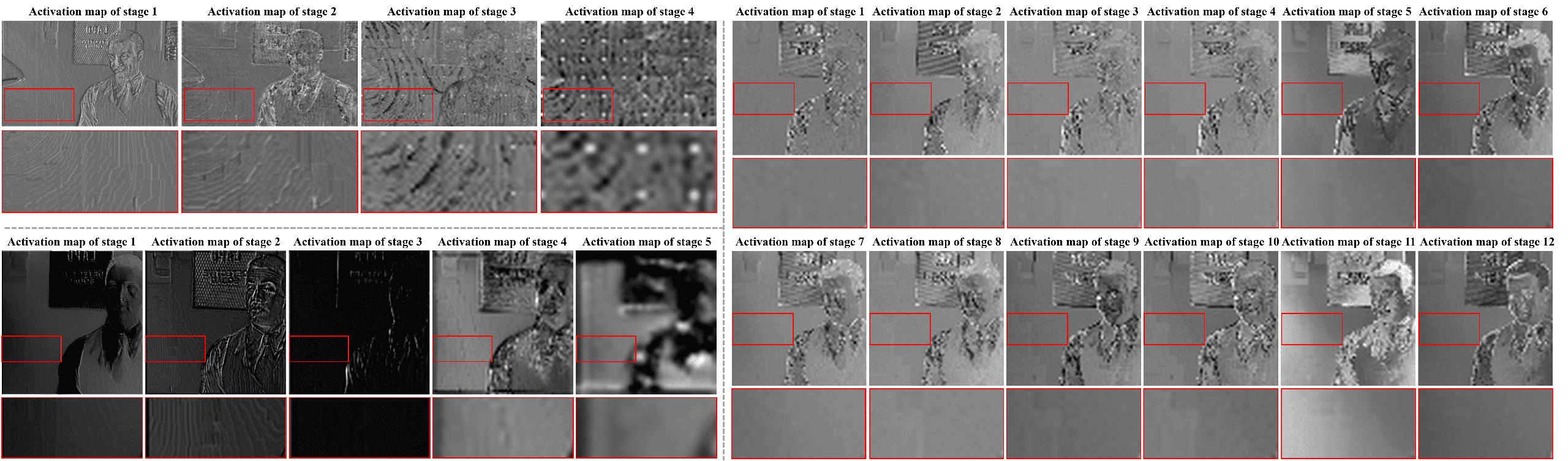} 
\end{tabular}
\vspace{-5pt}
\caption{Visual comparison of the expression of banding artifacts by MambaVision (left top), R2Plus1D (left bottom), and Vision Transformer (right).}
\vspace{-15pt}
\label{fig:different_layers_activation_3d_mamba_vit}
\end{figure*}

\begin{figure}[t]
\centering
\includegraphics[width=0.5\textwidth]{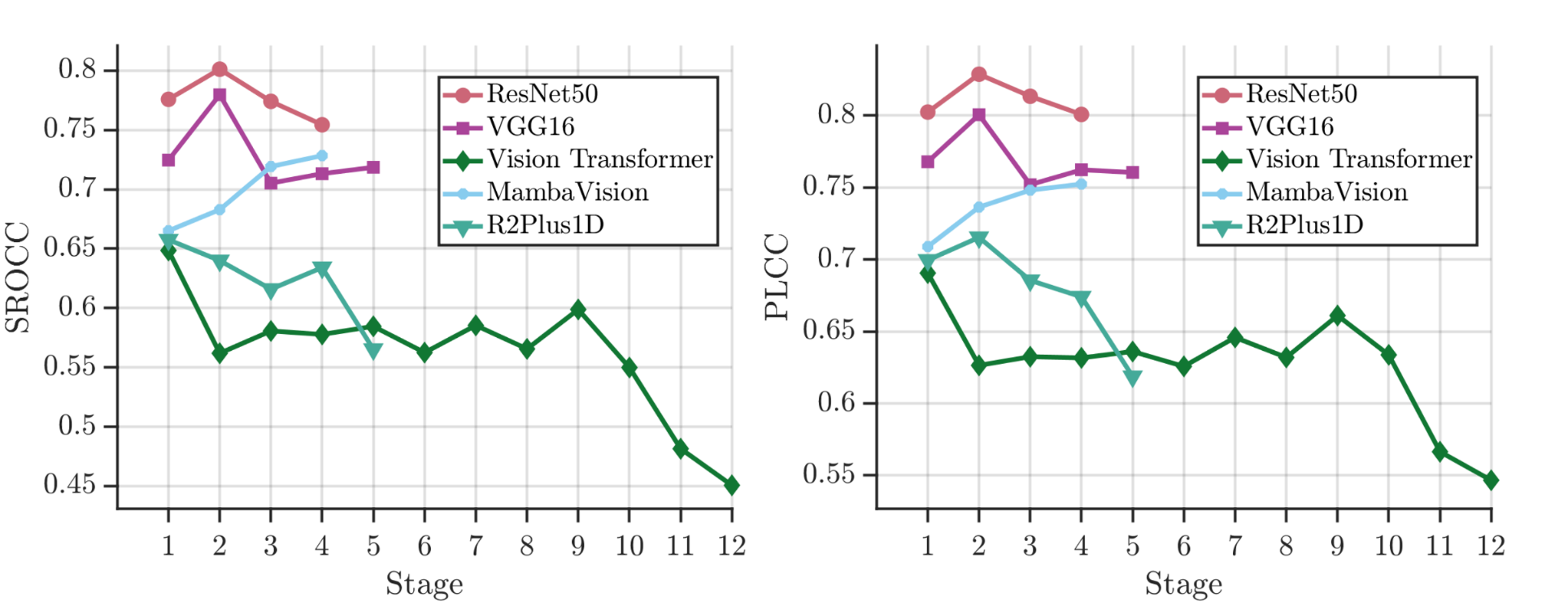}
\vspace{-12pt}
\caption{Performance comparison among different network architectures on LIVE-YT-Banding dataset.}
\label{fig:srocc_plcc_diff_archi}
\vspace{-12pt}
\end{figure}

\section{Experiment on Robustness to Content Variations}
In Section IV-B, we highlighted the nonlinear relationship between MOS and CRF levels in the LIVE-YT-Banding dataset. While banding generally increases with CRF, some videos show diminishing or stable banding at higher compression levels. This phenomenon is crucial for practical applications, as banding visibility does not always scale linearly with compression, emphasizing the need for robust VQA models that can handle such variations. To further investigate this, we conducted an experiment evaluating the robustness of different IQA/VQA models in handling content-dependent variations in banding perception. We categorized the 40 video contents in our dataset into two groups: (1) ``Align" dataset (22 contents) where MOS consistently decreases with increasing CRF; (2) ``Not-Align" dataset (18 contents) where MOS does not follow a strict decreasing trend.

We evaluated all models on both subsets, and the results in Table~\ref{table:performance-robustness-content} reveal several key insights. General-purpose IQA/VQA models perform better on the ``align" dataset than on the ``not-align" dataset, as their predictions are more consistent when distortion visibility progressively increases. Learning-based IQA/VQA models outperform handcrafted methods on the ``not-align" dataset, suggesting that deep neural networks exhibit better adaptability to complex banding perception patterns. Surprisingly, existing banding-specific models such as CAMBI and BBAND show weak performance on the not-align dataset, indicating that they may still be influenced by general compression distortions rather than being purely banding-sensitive. In contrast, the proposed CBAND models achieve the best performance across both datasets, demonstrating their robustness in capturing banding distortions regardless of content variations. This experiment underscores the necessity of robust banding-aware VQA models that generalize across varying content characteristics. CBAND’s ability to maintain high performance despite content-driven variations makes it a reliable solution for real-world applications.

\begin{table}[!t]
\renewcommand{\arraystretch}{1.2}
\centering
\fontsize{6.5pt}{7.5pt}\selectfont
\setlength{\tabcolsep}{2.5pt}
\caption{Robustness comparison of evaluated models. Italics denote FR methods. Boldfaced entries indicate top-1 performers.}
\vspace{-5pt}
\label{table:performance-robustness-content}
\begin{tabular}{l|l|ccccc}
\toprule
& &  \multicolumn{2}{c}{\textbf{Align}} & &  \multicolumn{2}{c}{\textbf{Not Align}} \\ 
\cline{3-4} \cline{6-7} \\[-1.em]
& \textbf{Model}                    &  \textbf{SROCC↑}    &
\textbf{PLCC↑}                         &   &  \textbf{SROCC↑}     &
\textbf{PLCC↑}                                    \\ 
\midrule
\multirow{12}{*}{\begin{tabular}[c]{@{}c@{}}General \\  IQA/VQA \\ methods \end{tabular}}
 &
BRISQUE~\cite{mittal2012no}                                     &     0.4486      &    0.6365  & &  0.0731 & 0.7351  \\  &
GM-LOG~\cite{xue2014blind}                                &    0.3216       &  0.6051    & &  0.2077 & 0.7037  \\ &
HIGRADE~\cite{kundu2017no}                             &   0.3197        &  0.6053    & & 0.2085  &  0.7025 \\ &
 NIQE~\cite{6353522}                                 &  0.1837         &  0.5786    & & 0.1034  & 0.6737 \\ &
FRIQUEE~\cite{FRIQUEE}                             &   0.3883        &  0.6340    & & 0.1779  & 0.6085  \\ &
HOSA~\cite{7501619}                             &  0.3141        &   0.6337   & & 0.4816  &  0.7601 \\ &
CORNIA~\cite{6247789}                           &   0.0308        &  0.4740   & & 0.1089  &  0.7103  \\ &
 VIDEVAL~\cite{9405420}                            &  0.2788         &   0.5900   & &  0.1231 &  0.7319  \\ &
TLVQM~\cite{korhonen2019two}                          &       0.1815    &   0.5917   & & 0.0795 & 0.6951  \\ &
FAVER~\cite{zheng2024faver}                           &    0.4359       &   0.6246   & & 0.1731 &  0.4156 \\ &
RAPIQUE~\cite{tu2021rapique}                          &    0.4070       &  0.5974    & &  0.0899 & 0.7060  \\ &
VSFA~\cite{li2019quality}                              &   0.6994        &  0.7106    & & 0.5995  &  0.6390 \\&
FAST-VQA~\cite{10.1007/978-3-031-20068-7_31}                                  &  0.4344         &  0.5600    & & 0.3176  & 0.6635  \\&
FasterVQA~\cite{10264158}                                &    0.4769       &   0.4558   & & 0.4010  &   0.2823 \\&
DOVER~\cite{Wu_2023_ICCV}                           &     0.6226      &  0.6145    & & 0.5077  & 0.6727  \\&
SAMA~\cite{Liu_Quan_Xiao_Li_Wu_2024}                           &    0.4674       &   0.5239   & &  0.3794 &  0.3000  \\&
ModularBVQA~\cite{Wen_2024_CVPR}                            &    0.6761       &  0.6726    & & 0.5286  &  0.6064 \\

\midrule
\multirow{3}{*}{\begin{tabular}[c]{@{}c@{}}Banding \\  IQA/VQA \\ methods \end{tabular}} &
\emph{$\text{VMAF}_\text{BA}$}~\cite{krasula2022banding}                          &    0.2814       &  0.5604    & &  0.0256 &  0.7229\\
  &
DBI~\cite{kapoor2021capturing}                           &   0.1749        &  0.5544    & & 0.0134  & 0.6899  \\ &
BBAND~\cite{tu2020bband}                              &    0.4267       &  0.6053    & & 0.3831  &  0.7399 \\ &
CAMBI~\cite{tandon2021cambi}                          &    0.3858       &  0.6069    & &  0.2854 &  0.7593 \\
\midrule
\multirow{2}{*}{Proposed}  &
CBAND-VGG16                                              &  0.7401       &   0.7548   & &  0.7108&  \textbf{0.7703} \\ &
CBAND-RN50                            &    \textbf{0.7750}       &  \textbf{0.7728}   & & \textbf{0.7441}  &0.7327  \\
\bottomrule
\end{tabular}
\vspace{-10pt}
\end{table}

\subsection{Evaluation of Alternative Pretrained Architectures for Banding Assessment}
\label{exp:alternative_pretrained_architectures}
To further investigate the compatibility of various pretrained architectures in banding quality assessment, we extend our analysis beyond CBAND’s backbone choices (ResNet50 and VGG16) to include 3D-CNN, Transformer-based, and Mamba-based models. Specifically, we evaluate R2Plus1D~\cite{8578773}, a 3D convolutional network that enhances temporal modeling by factorizing 3D convolutions into separate spatial and temporal components; Vision Transformer (ViT)~\cite{vit_ori}, which treats images as sequences of patch embeddings and applies self-attention to model long-range dependencies; and MambaVision~\cite{hatamizadeh2024mambavisionhybridmambatransformervision}, a recently introduced hybrid Mamba-Transformer backbone that integrates selective state-space modeling for efficient visual representation learning. These architectures represent diverse feature extraction paradigms and provide insights into how different network designs impact banding-aware feature learning.

We first decompose each architecture into multiple stages following their inherent design principles, with the number of stage-wise output channels summarized in Table~\ref{table:channel_num_extended_archi}. To further investigate how each model encodes banding artifacts, we visualized the stage-wise activation maps, as shown in Figure~\ref{fig:different_layers_activation_3d_mamba_vit}. Our observations reveal distinct differences: R2Plus1D captures banding artifacts well in its early stages, but its sensitivity diminishes in later stages, where higher-level motion features dominate. MambaVision maintains strong banding-aware activations across all stages, with deeper stages further concentrating on artifact regions. In contrast, ViT struggles to retain banding information at all stages, with minimal responsiveness in deeper stages. These findings suggest that convolutional and state-space models are more effective in learning spatially structured distortions such as banding.

\begin{table}[t]
\caption{Number of output channels in different architectures.}
\vspace{-5pt}
\label{table:channel_num_extended_archi}
\centering
\fontsize{6pt}{7pt}\selectfont
\setlength{\tabcolsep}{1.5pt}
\begin{tabular}{l|cccccc}
\toprule
\textbf{Model}                               & \textbf{Stage 1}& \textbf{Stage 2}                &\textbf{Stage 3}  & \textbf{Stage 4}        & \textbf{Stage 5} & \textbf{Stage 6-12}          \\ \midrule
Resnet50~\cite{he2016deep} & 256& 512& 1024& 2048& - & - \\
VGG16~\cite{simonyan2014very}  &64 &128 &256 &512 &512 & - \\
R2Plus1D~\cite{8578773}  & 2048&2048 & 2048& 2048& 2048&- \\
MambaVision~\cite{hatamizadeh2024mambavisionhybridmambatransformervision}  &80 & 160& 320& 640& -& - \\
Vision Transformer~\cite{vit_ori}  & 768 &768 &768 &768 &768 &768 \\
\bottomrule
\end{tabular}
\vspace{-15pt}
\end{table}

To quantitatively assess each architecture’s effectiveness, we extract NSS features from each stage’s activation maps and apply an MLP regression, mirroring the CBAND pipeline. The stage-wise SROCC and PLCC results on the LIVE-YT-Banding dataset are depicted in Figure~\ref{fig:srocc_plcc_diff_archi}. The results align with our previous qualitative observations. ViT exhibits poor performance as well as a steady decline in correlation as depth increases, due to its self-attention mechanism distributing focus across broader spatial regions without explicitly preserving localized distortions. R2Plus1D performs well in its early stages but deteriorates in deeper layers, as it transitions toward motion-oriented feature extraction. MambaVision retains relatively strong performance across multiple stages, with later stages enhancing banding-sensitive representations, indicating the state-space modeling in MambaVision reinforcing its effectiveness in structured artifact detection. Despite the promising results of MambaVision, CBAND-RN50 and CBAND-VGG16 still achieve the best performance across all metrics. This superior performance can be attributed to their explicit focus on banding-sensitive spatial statistics. By leveraging early-stage CNN features, CBAND effectively captures fine-grained banding distortions while preserving strong correlations with human perception. Even though, our evaluation results highlight the potential of Mamba-based architectures for perceptual video quality tasks.



\subsection{Generalization of CBAND on UGC VQA Datasets}
To assess the generalization of CBAND beyond banding artifacts, we evaluate its performance on LSVQ~\cite{Ying_2021_CVPR} and KoNViD-1k~\cite{7965673}, which contain diverse UGC distortions. As shown in Table~\ref{table:performance-ugc-dataset}, existing banding-specific models (BBAND, CAMBI) fail on UGC datasets, confirming their limited applicability beyond banding. In contrast, CBAND-RN50 and CBAND-VGG16 achieve competitive results, demonstrating their broader VQA potential.

Since CBAND extracts early-stage CNN features, we analyze all stages of ResNet50 and VGG16 to identify those most responsive to UGC distortions. Results show that stage 4 of ResNet-50 (CBAND-RN50-S4) and stage 5 of VGG16 (CBAND-VGG16-S5) yield superior performance, surpassing the original CBAND variants. These findings highlight CBAND’s adaptability and effectiveness in assessing diverse video distortions beyond banding.

\begin{table}[!t]
\renewcommand{\arraystretch}{1.2}
\centering
\fontsize{6pt}{7pt}\selectfont
\setlength{\tabcolsep}{1.5pt}
\caption{Performance comparison of evaluated models on UGC datasets. Metrics are SROCC/PLCC. Boldfaced entries indicate top-1 performers.}
\vspace{-5pt}
\label{table:performance-ugc-dataset}
\begin{tabular}{l|l|ccc}
\toprule
& {\textbf{Model}}   & \textbf{LSVQ-test}&  \textbf{LSVQ-1080P} &  \textbf{KoNViD-1k}   \\ 
\midrule
\multirow{12}{*}{\begin{tabular}[c]{@{}c@{}}General \\  IQA/VQA \\ methods \end{tabular}}
 &
BRISQUE~\cite{mittal2012no}                                      & 0.569/0.576  &  0.479/0.531     & 0.661/0.662   \\  
&VIDEVAL~\cite{9405420}                                                                   &   0.795/0.783  &  0.545/0.554    & 0.785/0.781     \\ &
TLVQM~\cite{korhonen2019two}                                                                  & 0.772/0.774&0.589/0.616      & 0.768/0.765      \\ 
& VSFA~\cite{li2019quality}                                                                & 0.801/0.796& 0.675/0.704     & 0.784/0.795\\ &
FAST-VQA~\cite{10.1007/978-3-031-20068-7_31}                                                               &  0.876/0.877  & 0.779/0.814     & 0.854/0.850     \\ &
FasterVQA~\cite{10264158}                                                                   & 0.873/0.874  &  0.772/0.811    &  0.827/0.828    \\ &
DOVER~\cite{Wu_2023_ICCV}                                                                 &  0.888/0.889  &  0.795/0.830  & 0.875/0.881  \\ &
SAMA~\cite{Liu_Quan_Xiao_Li_Wu_2024}                                                                   &   0.883/0.884     & 0.782/0.822& \textbf{0.880}/0.877    \\ &
ModularBVQA~\cite{Wen_2024_CVPR}                                                                &    \textbf{0.895/0.895}     &\textbf{0.809/0.844} &  0.878/\textbf{0.884}     \\

\midrule
\multirow{2}{*}{\begin{tabular}[c]{@{}c@{}}Banding \\VQA\end{tabular}}
 &
BBAND~\cite{tu2020bband}                                  &  0.1502/0.1686     & 0.0156/0.0941& 0.141/0.231     \\ &
CAMBI~\cite{tandon2021cambi}                                       &  0.101/0.123    & 0.225/0.276& 0.355/0.396 \\ 
\midrule
\multirow{4}{*}{Proposed} &
CBAND-VGG16                             &    0.711/0.706  &  0.602/0.640&0.718/0.720     \\&
CBAND-VGG16-S5                             &   0.808/0.803  &  0.678/0.720& 0.800/0.797    \\ &
CBAND-RN50                                                                &  0.773/0.770  &  0.656/0.705 &0.816/0.813     \\  &
CBAND-RN50-S4                                                               & 0.822/0.816  & 0.717/0.754& 0.777/0.783    \\ 
\bottomrule
\end{tabular}
\vspace{-12pt}
\end{table}

Moreover, to demonstrate that CBAND can indeed serve effectively as a modular enhancement for existing general-purpose IQA/VQA models, we conducted experiments involving both handcrafted and deep learning-based IQA/VQA methods on UGC VQA datasets which exhibit diverse and complex distortions. Specifically, we selected three traditional baseline methods with relatively lower performance—BRISQUE, VIDEVAL, and TLVQM—and one state-of-the-art, deep-learning-based model—ModularBVQA—to evaluate the benefit of integrating CBAND-RN50. The integration strategies were carefully designed as follows:
\begin{itemize}
    \item For the handcrafted methods (BRISQUE, VIDEVAL, TLVQM), we concatenated the original handcrafted features with CBAND's banding-aware NSS features and applied their original Support Vector Regression (SVR) method to predict quality.
    \item For ModularBVQA, which contains base, spatial, and temporal branches generating individual quality scores, we integrated the CBAND-RN50-generated score into these original scores and fine-tuned the overall quality fusion module to output an enhanced final quality prediction.
\end{itemize}

The experimental results (Table~\ref{table:performance-boosted-ugc-dataset}) clearly demonstrate CBAND's effectiveness as a modular enhancement across general IQA/VQA methods and diverse datasets. Integrating CBAND-RN50 substantially improved the initially limited handcrafted metrics, achieving notable relative gains (12\%-63\% on LSVQ datasets; 13\%-29\% on KoNViD-1k), especially evident in BRISQUE's 60\%+ improvement. Even the top-performing ModularBVQA benefited consistently, showing incremental gains (~2\%-4\%). These results confirm CBAND’s capability to significantly enhance both handcrafted and deep-learning-based models in general visual quality assessment, highlighting its broader practical relevance.

\begin{table}[!t]
\renewcommand{\arraystretch}{1.2}
\centering
\fontsize{6pt}{7pt}\selectfont
\setlength{\tabcolsep}{1.5pt}
\caption{Performance comparison of general-purpose IQA/VQA models boosted by CBAND-RN50 on UGC datasets. Metrics are SROCC/PLCC.}
\vspace{-5pt}
\label{table:performance-boosted-ugc-dataset}
\begin{tabular}{l|ccc}
\toprule
{\textbf{Model}}   & \textbf{LSVQ-test}&  \textbf{LSVQ-1080P} &  \textbf{KoNViD-1k}   \\ 
\midrule
BRISQUE~\cite{mittal2012no}                                      & 
0.841$_\text{\textcolor{red}{+47.8\%}}$/0.839$_\text{\textcolor{red}{+45.6\%}}$  &  0.779$_\text{\textcolor{red}{+62.6\%}}$
/0.815$_\text{\textcolor{red}{+53.4\%}}$     & 0.852$_\text{\textcolor{red}{+28.9\%}}$/0.846$_\text{\textcolor{red}{+27.8\%}}$   \\  
VIDEVAL~\cite{9405420}                                                                   &   0.893$_\text{\textcolor{red}{+12.3\%}}$/0.907$_\text{\textcolor{red}{+15.8\%}}$  &  0.784$_\text{\textcolor{red}{+43.8\%}}$/0.803$_\text{\textcolor{red}{+44.9\%}}$    & 0.893$_\text{\textcolor{red}{+13.8\%}}$/0.883$_\text{\textcolor{red}{+13.1\%}}$     \\ 
TLVQM~\cite{korhonen2019two}                                                                  & 0.891$_\text{\textcolor{red}{+15.4\%}}$/0.893$_\text{\textcolor{red}{+15.3\%}}$ & 0.818$_\text{\textcolor{red}{+38.8\%}}$/0.834$_\text{\textcolor{red}{+35.3\%}}$      & 0.884$_\text{\textcolor{red}{+15.1\%}}$/0.877$_\text{\textcolor{red}{+14.6\%}}$      \\ 
ModularBVQA~\cite{Wen_2024_CVPR}                                                                &    0.922$_\text{\textcolor{red}{+3.0\%}}$/0.934$_\text{\textcolor{red}{+4.3\%}}$     &0.837$_\text{\textcolor{red}{+3.5\%}}$/0.861$_\text{\textcolor{red}{+2.0\%}}$ &  0.910$_\text{\textcolor{red}{+3.6\%}}$/0.916$_\text{\textcolor{red}{+3.6\%}}$        \\ 
\bottomrule
\end{tabular}
\vspace{-12pt}
\end{table}

\begin{table}[!t]
\caption{Ablation study of NSS features.}
\vspace{-6pt}
\label{table:ablation_nss}
\centering
\fontsize{6pt}{7pt}\selectfont
\setlength{\tabcolsep}{1.5pt}
\begin{tabular}{l|cccccc}
\toprule
\textbf{Model}                               & \textbf{Mean}& \textbf{Standard deviation}              & \textbf{SROCC↑}  & \textbf{KROCC↑} & \textbf{PLCC↑}  & \textbf{RMSE↓}               \\ \midrule
\multirow{3}{*}{CBAND-VGG16}                   & \textcolor[RGB]{18,220,168}{\ding{52}}   &   & 0.7407          &  0.5738               & 0.7705   &  12.2889                                 \\  
                   &   &  \textcolor[RGB]{18,220,168}{\ding{52}}          &  0.6556  &   0.4896              &    0.6993 & 13.9556        \\   
                    & \textcolor[RGB]{18,220,168}{\ding{52}}&  \textcolor[RGB]{18,220,168}{\ding{52}}          & \textbf{0.7797} & \textbf{0.6048}   & \textbf{0.8006}                &  \textbf{11.5938}            \\    \midrule
\multirow{3}{*}{CBAND-RN50}               & \textcolor[RGB]{18,220,168}{\ding{52}}   &             & 0.7811 &  0.6125  &  0.8217               & 10.9548          \\  
                    &  &   \textcolor[RGB]{18,220,168}{\ding{52}}          &0.7372 & 0.5689   &      0.7919           & 11.6753               \\ 
                         & \textcolor[RGB]{18,220,168}{\ding{52}}  &   \textcolor[RGB]{18,220,168}{\ding{52}}          &  \textbf{0.8012}  &   \textbf{0.6289}              &   \textbf{0.8287}  & \textbf{10.6743}     \\ 
                         
\bottomrule
\end{tabular}
\vspace{-12pt}
\end{table}

\begin{table}[!t]
\centering
\fontsize{6pt}{7pt}\selectfont
\setlength{\tabcolsep}{1.5pt}
\caption{Ablation study on stages of activation maps.}
\vspace{-6pt}
\label{table:ablation_study_stages}
    \begin{tabular}{c|ccccccccc}
    \toprule
  &  \multicolumn{4}{c}{\textbf{CBAND-VGG16}} & & \multicolumn{4}{c}{\textbf{CBAND-RN50}}   \\
  \cline{2-5} \cline{7-10} \\[-1.em]
 \textbf{Stage} &  \textbf{SROCC↑}& \textbf{KROCC↑} & \textbf{PLCC↑}& \textbf{RMSE↓} & & \textbf{SROCC↑} & \textbf{KROCC↑} & \textbf{PLCC↑} & \textbf{RMSE↓} \\ \midrule
 1& 0.7246& 0.5544 & 0.7679&12.4516 &  &0.7758 & 0.5984& 0.8024 & 11.5989\\
2&  \textbf{0.7797}              &   \textbf{0.6048}      & \textbf{0.8006}     &     \textbf{11.5938} & & \textbf{0.8012}              &   \textbf{0.6289}      &   \textbf{0.8287}    &    \textbf{10.6743}  \\
 3&0.7052 & 0.5382 & 0.7520& 12.5872& & 0.7741& 0.5974& 0.8134& 11.2256\\
4& 0.7132& 0.5430 & 0.7623& 12.4862& & 0.7544 & 0.5797& 0.8008& 11.6385\\
 5& 0.7186 &0.5508 & 0.7605& 12.2797& & -&- &- &- \\
    
    \bottomrule
    \end{tabular}
    \vspace{-20pt}
\end{table}

\subsection{Ablation Study on NSS Features and Activation Maps }

To validate the effectiveness of each feature component in CBAND, we conducted comprehensive ablation studies on the NSS features. CBAND extracts two types of NSS features from each activation map: the mean and standard deviation of spatial activations. To evaluate their individual contributions, we trained and tested CBAND using only the mean, only the standard deviation, and both combined. The results in Table~\ref{table:ablation_nss} indicate that the mean feature alone contributes more significantly to performance than the standard deviation. However, combining both features leads to the highest overall performance, demonstrating that jointly leveraging mean and standard deviation improves CBAND’s ability to capture banding artifacts effectively.

To further investigate the role of different network stages in capturing banding distortions, we conducted an ablation study evaluating CBAND using feature representations extracted from different stages of ResNet50 and VGG16. As shown in Table~\ref{table:ablation_study_stages}, the performance varies across stages, with earlier layers generally being more effective at capturing banding-sensitive information. Notably, stage 2 achieves the best predictive performance for both ResNet50 and VGG16, aligning with our prior visualization analysis in Section~\ref{ssec:banding_aware_activation_maps}. This justifies our final design choice of using stage 2 as the foundation of CBAND.

\subsection{Effects of Temporal Sampling Rate}
The temporal redundancy of adjacent video frames is a fundamental property that drives the development of video compression techniques.
Likewise, we can also leverage the temporal redundancy of frame-wise perceptual quality to optimize inference efficiency, without significantly compromising model performance.
To explore temporal sampling effects, we varied the temporal sampling rate of the frame-level banding-aware feature extraction in our CBAND implementations. 
Performance comparisons between our models using different temporal sampling rates are reported in Table~\ref{table:sampling-rate}.
Specifically, the outcomes obtained using five different sampling strides, every 30 frames, 20 frames, 10 frames, 5 frames, and every frame, are compared for both CBAND-RN50 and CBAND-VGG16.
Table~\ref{table:sampling-rate} reports that variations in the temporal sampling rate have different impacts on the performances of the two CBAND models.
For CBAND-RN50, the performance remained relatively robust across the temporal sampling rates.
This suggests that CBAND-RN50 can be made even more computationally efficient by skipping redundant frames without suffering performance drops.
Conversely, gradual decreases in performance of CBAND-VGG16 were observed as the temporal sampling rate was decreased.
However, as shown in Table~\ref{table:performance-all-model}, the performance of CBAND-VGG16 using the slowest sampling rate (every 30 frames) still surpassed that of all the compared other models.

\begin{table}[!t]
\caption{Effects of temporal sampling rate (stride) on the CBAND models. The boldfaced entries indicate the top-1 performers.}
\vspace{-5pt}
\label{table:sampling-rate}
\centering
\fontsize{6pt}{7pt}\selectfont
\setlength{\tabcolsep}{1.5pt}
\begin{tabular}{l|ccccc}
\toprule
\textbf{Model}                               & \begin{tabular}[c]{@{}c@{}}\textbf{Sampling} \\ \textbf{Stride} \end{tabular}& \textbf{SROCC↑}                 & \textbf{KROCC↑}  & \textbf{PLCC↑}          & \textbf{RMSE↓}           \\ \midrule
\multirow{5}{*}{CBAND-VGG16}                   & 1 second    &  0.7388           & 0.5664                &  0.7811 & 12.1084                   \\  
                    &
                   20 frames &    0.7398             & 0.5658  & 0.7779         &    12.2214     \\ 
                          &          10 frames & 0.7441                       &     0.5722       & 0.7861 & 11.9930        \\ 
                          &
                 5 frame  &        0.7440         &  0.5708        & 0.7778 &  12.1839    \\ &
                 every frame  &   \textbf{0.7797}              &   \textbf{0.6048}      &  \textbf{0.8006}     &     \textbf{11.5938}      \\ \midrule
\multirow{5}{*}{CBAND-RN50}                    & 1 second    &    0.7958         &   0.6211              & 0.8301 & 10.6846                   \\  
                    &
                   20 frames &   0.7973              & 0.6232  &  0.8220        &  10.8980       \\ 
                          &          10 frames &   \textbf{0.8015}                     &    0.6282        & 0.8232 & 10.8142        \\ 
                          &
                 5 frame  &     0.7994            &  0.6246        & \textbf{0.8302} & \textbf{10.6280}     \\ &
                 every frame  &    0.8012              &   \textbf{0.6289}      &   0.8287    &    10.6743  \\
\bottomrule
\end{tabular}
\begin{tablenotes}
    \item[1] Sampling stride 1 second: sampled once per second when extracting NSS features. Sampling stride 20 (or 10, or 5) frames: extracted NSS features every 20 (or 10, or 5) frames.
    \end{tablenotes}
\vspace{-17pt}
\end{table}

\begin{figure*}[!t]
\centering
\footnotesize
\includegraphics[width=0.9\linewidth]{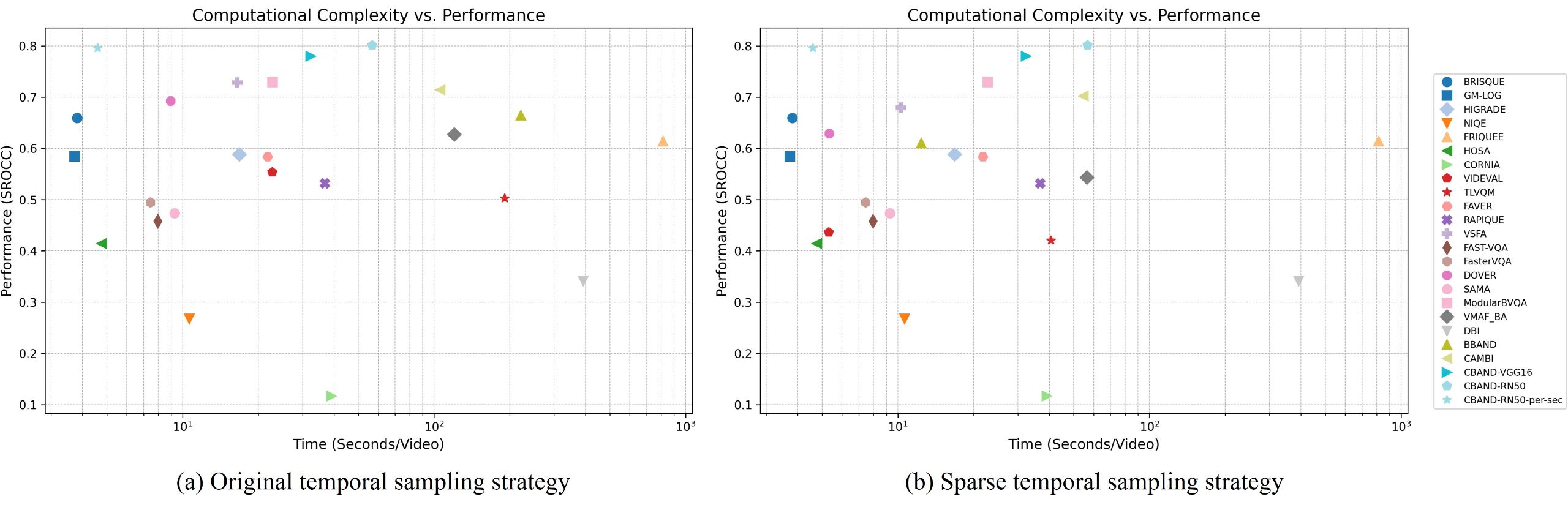} 
\vspace{-10pt}
\caption{Computational complexity comparison between original and sparse sampling strategies.}
\vspace{-10pt}
\label{fig:computational_comp}
\end{figure*}

\begin{figure}[t]
\centering
\footnotesize
\includegraphics[width=0.9\linewidth]{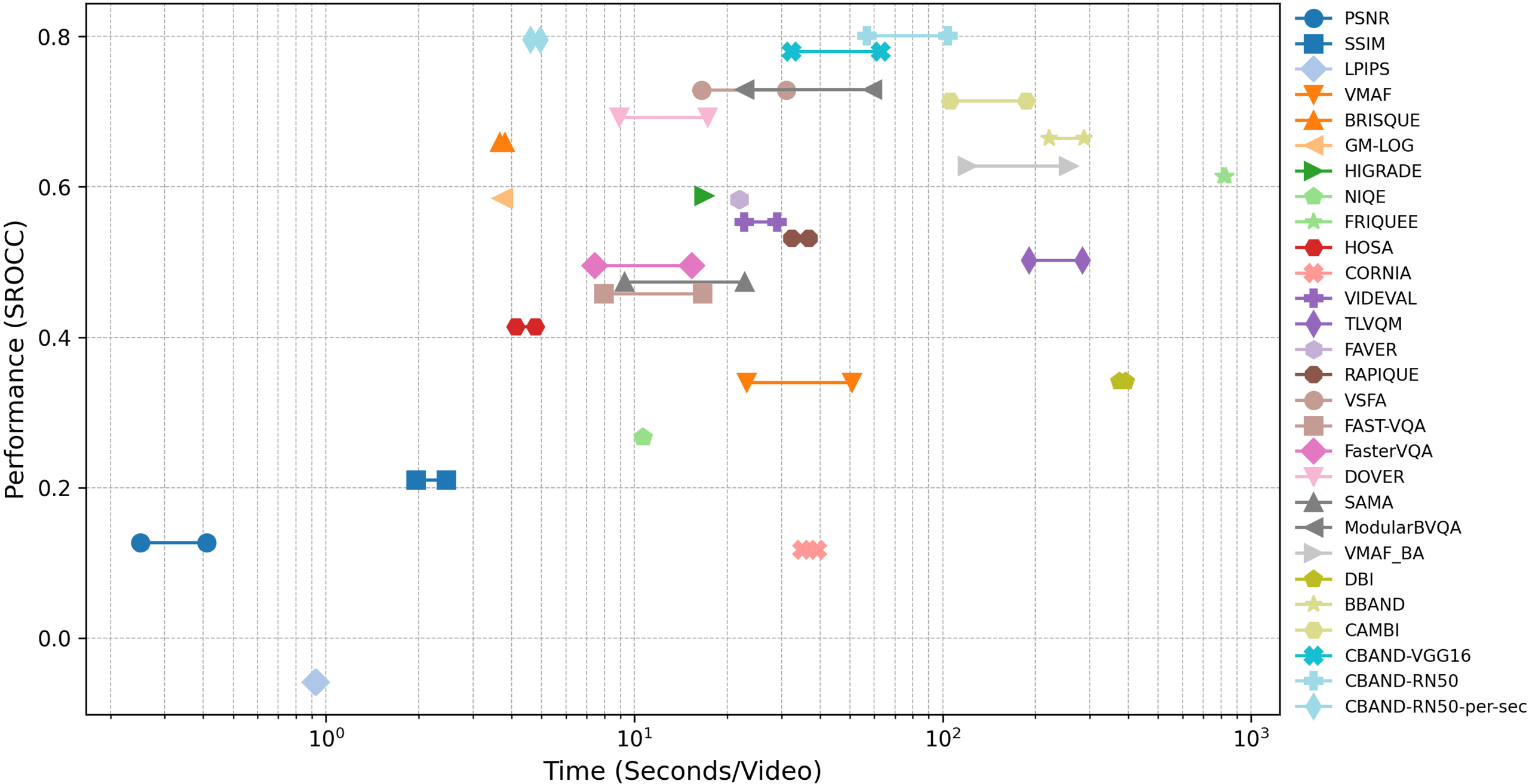} 
\vspace{-10pt}
\caption{Computational complexity vs. performance on banding VQA.}
\vspace{-12pt}
\label{fig:srocc_complexity}
\end{figure}

\subsection{Computational Complexity}
The computational complexity of a video quality model plays a critical role in its potential impact on real-world video streaming platforms.
Thus, we carried out a computational complexity analysis on the compared models.
For fair comparison, all of the experiments were evaluated on the same server equipped with 32 Intel Xeon E5-2620 v4 CPU processors and 4 NVIDIA TITAN RTX Graphics Cards.
Specifically, PSNR, SSIM, and LPIPS were computed using the pyiqa~\cite{10478301} package in Python.
VMAF, CMABI, and $\text{VMAF}_\text{BA}$ were computed in ffmpeg~\cite{team2013ffmpeg}, and VSFA and DBI were implemented using their original releases.
The remaining models were implemented using their original releases in MATLAB.
We ran the models as is--- accounting for the video sampling designs---and recorded the time costs on 1080p \@30fps and 60 fps videos of 7 seconds duration, respectively.
Note that the CBAND-RN50 model maintained its superior performance even at the slowest temporal sampling rate; thus, we evaluated the slowest sampled version as well, which is denoted as CBAND-RN50-per-sec.
%
Figure~\ref{fig:srocc_complexity} shows the time costs of all the compared models. As may be observed, the CBAND models are extremely efficient as compared to other top-performing models like ModularBVQA and CAMBI.
When evaluated on 1080p \@30fps videos, slowest sampled model CBAND-RN50-per-sec was \textbf{22.82x} and \textbf{4.96x} faster, respectively, than the SOTA banding prediction model CAMBI and the SOTA general-purpose model ModularBVQA.
When evaluated on 1080p \@60fps videos, CBAND-RN50-per-sec was \textbf{37.64x} and \textbf{11.92x} faster than CAMBI and ModularBVQA, respectively. Note that CBAND achieves high efficiency by directly using the second stage of a pretrained ResNet-50 for feature extraction, while ModularBVQA adopts the first two stages of ResNet-18 for a spatial rectifier and relies on deeper ViT layers, increasing computational complexity. 

Moreover, a comprehensive additional experiment comparing both computational complexity and sensitivity to temporal sampling strategies across various IQA/VQA methods is provided here. We systematically analyzed the temporal sampling feasibility for all evaluated IQA/VQA methods (as clearly summarized in Table~\ref{table:temporal-sampling-analysis-all-model}), and conducted additional rigorous experiments to evaluate their frame sensitivity on 7-second videos with a frame rate of 30 fps following original computation complexity experiment. This new comparative experiment is carefully designed to ensure fairness: all methods were sampled as sparsely as possible without fundamentally compromising their original temporal fusion mechanisms or algorithmic integrity. Specifically:

\begin{itemize}
    \item For IQA methods (e.g., BRISQUE, NIQE, etc.), our original evaluations were already conducted at a sparse rate of one frame per second. Therefore, no further sparsification was possible without compromising their fundamental frame-wise scoring principle.
    \item For VQA methods, we thoroughly examined each model’s temporal sampling strategy. For simpler temporal-insensitive methods (e.g., BBAND, $\text{VMAF}_{\text{BA}}$, and CAMBI), we were able to reduce the sampling rate significantly (one frame per second), matching CBAND's sparse setting. For methods with moderate temporal modeling requirements (e.g., VSFA, VIDEVAL, TLVQM, and DOVER), we carefully adjusted to the sparsest possible sampling without substantially compromising their temporal quality modeling principles. However, methods strictly designed with specific frame-length requirements or sophisticated temporal modeling (e.g., FAVER, RAPIQUE, ModularBVQA) were either minimally sparsified or maintained in their original optimal configurations to preserve their inherent algorithmic integrity. Finally, for methods such as FAST-VQA and FasterVQA, we have utilized their most efficient official versions in previous evaluation, maintaining their reported performance and efficiency.
\end{itemize}

Our results, as shown in Figure~\ref{fig:computational_comp}, clearly demonstrate that the insensitivity to frame reduction varies significantly across existing VQA methods. For example, CAMBI exhibited robustness to reduced frame sampling, with SROCC slightly decreasing from 0.7143 to 0.7021. In contrast, other VQA methods demonstrated marked sensitivity to frame reduction, showing obvious performance degradation when temporal sampling was sparsified (e.g., BBAND, $\text{VMAF}_{\text{BA}}$, DOVER, and VIDEVAL), due primarily to differences in their inherent temporal fusion designs. Thus, the assumption that the ability to skip redundant frames might simply be inherent to the banding assessment task is not universally valid.

\begin{table*}[!t]
\renewcommand{\arraystretch}{1.2}
\centering
\fontsize{6.5pt}{7.5pt}\selectfont
\setlength{\tabcolsep}{4.5pt}
\caption{Systematical analysis of temporal sampling strategy of evaluated IQA/VQA methods.}
\vspace{-5pt}
\label{table:temporal-sampling-analysis-all-model}
\begin{tabular}{l|c|c|c}
\toprule
 {\textbf{Model}}                 &  \textbf{Original temporal sampling}                      &  \textbf{If possible for sparser sampling?} &
\textbf{Evaluated sparser sampling}                         \\ 
\midrule

BRISQUE~\cite{mittal2012no}                                  &    1 frame per second         &  No. Already sparsest  &   -     \\ \cline{1-4} 

GM-LOG~\cite{xue2014blind}                                           &    1 frame per second            &  No. Already sparsest  &   -       \\ \cline{1-4} 

HIGRADE~\cite{kundu2017no}                                        &    1 frame per second         &  No. Already sparsest  &   -            \\ \cline{1-4} 

NIQE~\cite{6353522}         &    1 frame per second           &  No. Already sparsest  &   -            \\ \cline{1-4} 
FRIQUEE~\cite{FRIQUEE}                            &    1 frame per second       &  No. Already sparsest  &   -        \\ \cline{1-4} 
HOSA~\cite{7501619}                                     &    1 frame per second         &  No. Already sparsest  &   -       \\ \cline{1-4} 

CORNIA~\cite{6247789}                                         &    1 frame per second           &  No. Already sparsest  &   -      \\ \cline{1-4} 

VIDEVAL~\cite{9405420}                                         &    Every two frames per second           & \begin{tabular}[c]{@{}c@{}}  Yes. The original sampling rate is 15 frames per second for\\ a 7-second video at 30 fps. The sampling rate can be \\reduced to as low as 4 frames per second, as the algorithm \\requires multiple frame-wise features within each second to \\compute meaningful averages and standard deviations, which \\are necessary to generate reliable second-wise features. To \\ensure statistical validity, a minimum of 4 frames per second\\ was therefore evaluated.   \end{tabular}& 4 frames per second.    \\ \cline{1-4} 

TLVQM~\cite{korhonen2019two}         &    Every two frames per second        & \begin{tabular}[c]{@{}c@{}}  Yes. The original sampling rate is 15 frames per second for\\ a 7-second video at 30 fps. The sampling rate can be \\reduced to as low as 4 frames per second, as the algorithm \\requires multiple frame-wise features within each second to \\compute meaningful averages and standard deviations, which \\are necessary to generate reliable second-wise features. To \\ensure statistical validity, a minimum of 4 frames per second\\ was therefore evaluated.   \end{tabular}& 4 frames per second.   \\  \cline{1-4} 
FAVER~\cite{zheng2024faver}                           &   \begin{tabular}[c]{@{}c@{}} Spatial: 1 frame per second  \\  Motion: 1-time calculation \\per second \\ Temporal: 1-time calculation \\per second \end{tabular}       &  \begin{tabular}[c]{@{}c@{}} No. For spatial branch, the original sampling \\ is already sparsest. For temporal and motion branch, \\ only one time calculation per second is required by \\the algorithm. \end{tabular} &  -   \\ \cline{1-4} 
RAPIQUE~\cite{tu2021rapique}                            &   \begin{tabular}[c]{@{}c@{}} Spatial: 2 frames per second  \\ CNN branch: 1 frame per second \\Temporal: 1-time calculation \\per second  \end{tabular}           & \begin{tabular}[c]{@{}c@{}} No. For the spatial branch, the model inherently requires\\ calculations on two frames per second—one for the current \\frame result and one for computing the inter-frame distance.\\ For the temporal branch, the model design explicitly \\mandates calculations using eight consecutive frames per \\second. Regarding the CNN branch, the original sampling \\rate is already at its sparsest possible setting. Therefore, \\further reduction of the sampling rate is not feasible \\without compromising the integrity of the method.  \end{tabular} &  -        \\ \cline{1-4}  
VSFA~\cite{li2019quality}                           &     Every frame per second         & \begin{tabular}[c]{@{}c@{}} Yes. The sampling rate can be reduced, but not below \\24 frames per second. This minimum sampling threshold is \\necessary because the model inherently relies on capturing \\temporal hysteresis effects, which require at least 24 frames \\per second as dictated by the optimal parameter $\tau$. \end{tabular}       &  24 frames per second      \\ \cline{1-4} 
 FAST-VQA~\cite{10.1007/978-3-031-20068-7_31}  &  4 clips with 16 frames each& \begin{tabular}[c]{@{}c@{}}  No. The previously evaluated model is already the\\ official efficient version (FAST-VQA-M), which has been \\specifically designed to utilize sparse sampling. \end{tabular}& - \\ \cline{1-4} 
 FasterVQA~\cite{10264158} & 4 clips with 4 frames each & \begin{tabular}[c]{@{}c@{}}  No. The previously evaluated model is already the\\ official efficient version (FasterVQA-MT), which has been \\specifically designed to utilize sparse sampling. \end{tabular}  & - \\ \cline{1-4} 
 DOVER~\cite{Wu_2023_ICCV}  & \begin{tabular}[c]{@{}c@{}} Aesthetic: 32 frames per video \\ Technical: 3 clips of 32 \\frames each \end{tabular}& \begin{tabular}[c]{@{}c@{}} Yes. For the aesthetic branch, the sampling rate can be\\ reduced from 32 frames per video down to 7 frames per \\video, effectively corresponding to approximately 1 frame per \\second. For the technical branch, the original three clips can be \\reduced to a single clip; however, the sequence of 32 consecutive \\frames within this clip cannot be further reduced due to the \\inherent requirements of its temporal quality modeling design. \end{tabular}& \begin{tabular}[c]{@{}c@{}}  Aesthetic: 7 frames per video \\ Technical: 1 clip of 32 frames \end{tabular}\\ \cline{1-4} 
 SAMA~\cite{Liu_Quan_Xiao_Li_Wu_2024} &  32 frames per video & \begin{tabular}[c]{@{}c@{}} No. The original sampling rate corresponds to 4 frames per \\second for a 7-second video. Due to inherent constraints \\imposed by the VideoSwin model, this input frame length \\cannot be reduced or adjusted to a sparser sampling strategy. \end{tabular}& - \\ \cline{1-4} 
ModularBVQA~\cite{Wen_2024_CVPR}  & \begin{tabular}[c]{@{}c@{}} Base: 7 frames per video\\ Spatial: 7 frames per video\\ Temporal: 7-time calculation \\ per video \end{tabular}&  \begin{tabular}[c]{@{}c@{}} No. For the base and spatial branches, the original sampling \\already corresponds to one frame per second for a 7-second video. \\Similarly, the temporal branch requires exactly one calculation per \\second. Due to the inherent algorithmic requirements, these \\temporal calculations rely on consecutive frames to effectively \\model temporal quality. Therefore, it is not feasible to further \\reduce or adjust the sampling to a sparser strategy. \end{tabular}& - \\ 
\midrule
 \emph{$\text{VMAF}_{\text{BA}}$}~\cite{krasula2022banding}                                  &    \begin{tabular}[c]{@{}c@{}} VMAF: every frame per second \\ CAMBI: 1 frame per 0.5 second\end{tabular}          &   Yes & 1 frame per second        \\ \cline{1-4} 

DBI~\cite{kapoor2021capturing}                            &    1 frame per second        &  No. Already sparsest  &   -    \\ \cline{1-4}  
BBAND~\cite{tu2020bband}                            &  Every frame per second          &   Yes    & 1 frame per second   \\ \cline{1-4}  
CAMBI~\cite{tandon2021cambi}                           &   1 frame per 0.5 second         &   Yes  &  1 frame per second      \\ 
\midrule
CBAND-RN50-per-sec             &   1 frame per second         &  No. Already sparsest  &   -      \\ 
\bottomrule
\end{tabular}
\vspace{-15pt}
\end{table*}

Crucially, even under equally sparse sampling conditions (approximately one frame per second), our CBAND-RN50-per-sec model maintained significantly higher performance than other methods as seen in Figure~\ref{fig:computational_comp} (b), clearly highlighting its intrinsic strength. Therefore, the per-sec comparison is entirely fair: our rigorous experimental design explicitly considered and respected the fundamental temporal modeling constraints of each method, thereby confirming that CBAND’s efficiency and robustness are genuine advantages stemming from its carefully crafted banding-aware neural architecture rather than task-specific frame insensitivity.

In summary, these thorough additional analyses decisively support the fairness and validity of our comparisons and reinforce that CBAND uniquely achieves superior performance and efficiency in the banding quality assessment task compared to state-of-the-art alternatives.

\begin{table}[!t]
\caption{Performance comparisons on the video debanding task. Boldfaced entries indicate top-1 performers.}
\vspace{-8pt}
\label{table:debanding_loss}
\centering
\fontsize{6pt}{7pt}\selectfont
\setlength{\tabcolsep}{4pt}
\begin{tabular}{l|ccccc}
\toprule
\textbf{Model}                               & \textbf{CBAND-RN50↑} & \textbf{CAMBI↓}                 & \textbf{BBAND↓}  & \textbf{PSNR↑}          & \textbf{LPIPS↓}           \\ \midrule
Groundtruth & 0.9151 & 5.5982 & 0.5944 & - & -\\
Input & 0.7727 & 10.8570 & 1.1623 & 44.1347 & 0.0188\\
NAFNet & 0.7105 & 9.4406 & 0.8942 & \textbf{44.3730} & 0.0228\\
NAFNet$^\text{ModularBVQA-S}$ & 0.7987 & 4.9402 & 0.8335 & 43.4008 & 0.0215 \\
NAFNet$^\text{CBAND-RN50}$ & \textbf{0.8628} & \textbf{3.3186} & \textbf{0.6044} & 43.9367 & \textbf{0.0186} \\
\bottomrule
\end{tabular}
\vspace{-12pt}
\end{table}

\begin{figure}[!t]
\centering
\footnotesize
\includegraphics[width=0.9\linewidth]{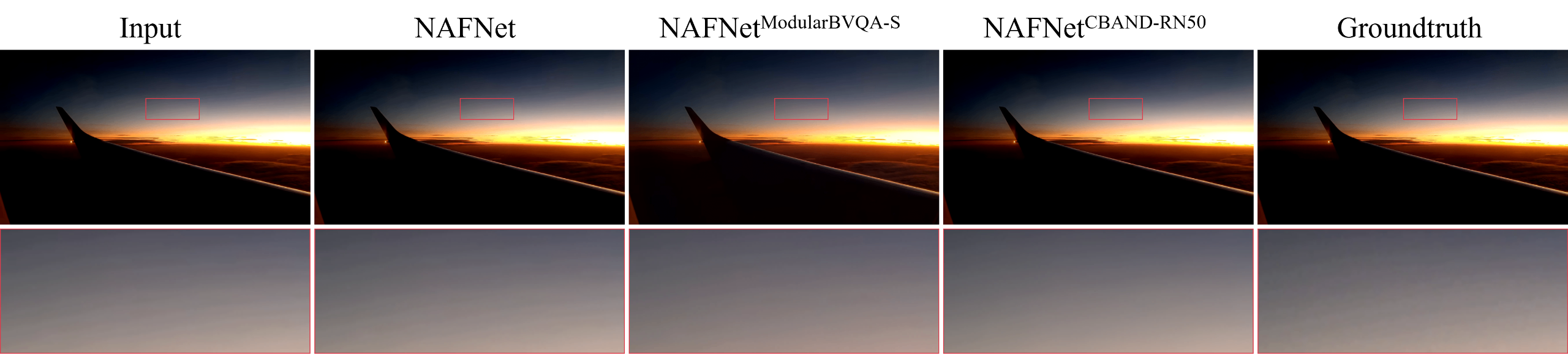} 
\vspace{-10pt}
\caption{Qualitative comparison of debanding results (better zoom in).}
\vspace{-17pt}
\label{fig:visual_comp_restoration}
\end{figure}


\subsection{Application to the Video Frame Debanding Task}
To validate CBAND's practical usability, we employed it as a differentiable loss function for video quality enhancement tasks. We also include the spatial branch of the state-of-the-art method ModularBVQA (denoted as ModularBVQA-S) for a clearer comparison, as it achieves comparable performance to the full ModularBVQA on the LIVE-YT-Banding dataset. All experimental settings for CBAND and ModularBVQA-S are carefully maintained to be identical to ensure fair and rigorous evaluation.
We utilized a state-of-the-art image restoration model, NAFNet~\cite{chen2022simple}, which achieves superior performance on image denoise, deblur, and super-resolution (SR) tasks, as our testbed.
We then built a paired frame dataset by extracting video frames from the LIVE-YT-Banding Dataset at a sampling rate of one frame per second. The reference videos (crf=0) served as ground truth, while the most distorted banding videos (crf=37) served as the degraded inputs.
We randomly split the banding frame dataset into 80\% training and 20\% test subsets. We trained NAFNet from scratch for 150 epochs using two optimization strategies: 1) training using the original MSE loss and 2) training linearly combined MSE and CBAND-RN50/ModularBVQA-S loss (CBAND/ModularBVQA-S loss weight=0.001, MSE weight=1-0.001). We measured the performances of the banding removal models using three top-performing banding-aware VQA metrics: CBAND-RN50, CAMBI, and BBAND. We also deployed the general quality prediction models PSNR and LPIPS, the latter popularly used to measure SR outcomes.

As shown in Table~\ref{table:debanding_loss}, integrating CBAND-RN50 as an additional loss function delivered significant performance improvement on video banding artifact removal, outperforming the baseline NAFNet by \textbf{21.43\%}, \textbf{64.85\%}, \textbf{32.41\%} in terms of CBAND, CAMBI, and BBAND, respectively. It may also be observed that NAFNet, optimized with CBAND-RN50, yielded better perceptual quality as measured by LPIPS, while delivering slightly worse PSNR results.
In short, the exceptional perceptual quality benefit obtained by CBAND-RN50 came at the cost of increased pixel-wise distortion. This result both aligns with known distortion-perception tradeoffs~\cite{blau2018perception, zhang2021universal}, and also with the known limitations of PSNR~\cite{4775883}. Moreover, NAFNet guided by CBAND-RN50 significantly outperformed the ModularBVQA-S-guided model across all perceptual quality metrics. Additionally, the computational complexity assessment shows that CBAND is significantly more efficient, with shorter inference time per video frame (0.6611s for CBAND vs. 0.7890s for ModularBVQA-S), making it more practical for real-world applications. We also provide the qualitative comparison of restored frames in Figure~\ref{fig:visual_comp_restoration}. Compared with ModularBVQA-S, the CBAND-driven model consistently yields smoother gradients and fewer banding artifacts, thereby achieving superior perceptual quality.

\begin{figure}[!t]
\centering
\footnotesize
\includegraphics[width=0.9\linewidth]{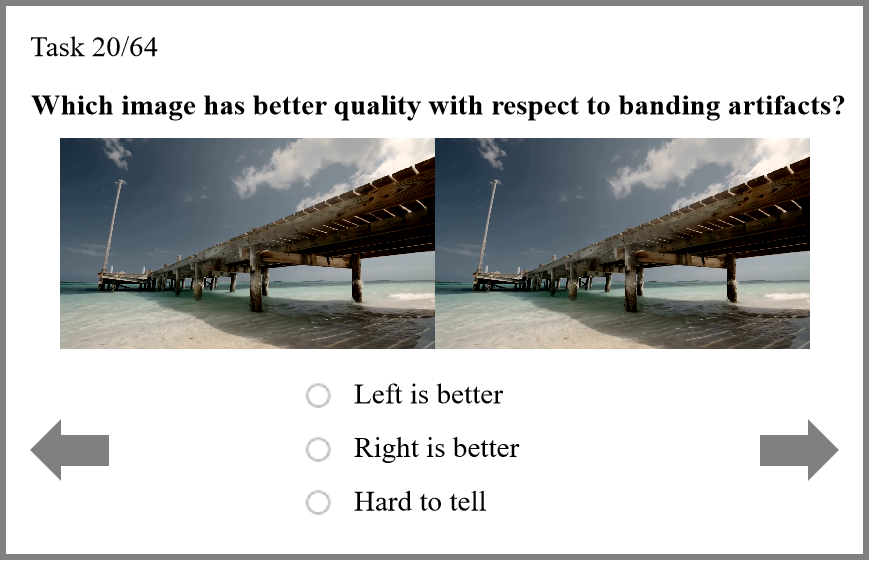} 
\vspace{-10pt}
\caption{User study interface.}
\vspace{-12pt}
\label{fig:user_study_interface}
\end{figure}

To further validate perceptual improvements, we conducted a user study. A total of 16 participants (8 male, 8 female, ages 21–27) evaluated 64 pairs of video frames, each containing a banding-corrupted image restored using two different training strategies. The user interface shown in Figure~\ref{fig:user_study_interface}, displayed the two frames side by side in a randomized order, ensuring counterbalancing of left-right positions. Participants were asked to choose which image exhibited better visual quality or indicate if the difference was imperceptible. Table~\ref{table:user_study_results} summarizes the aggregated results from all the individual decisions. The CBAND-optimized NAFNet was preferred in 87.40\% of the cases, which strongly confirms that CBAND-optimized NAFNet yields significantly better perceptual quality for banding removal.

\begin{table}[!t]
\caption{User study results.}
\vspace{-8pt}
\label{table:user_study_results}
\centering
\fontsize{6.5pt}{7.5pt}\selectfont
\setlength{\tabcolsep}{4.5pt}
\begin{tabular}{l|cc}
\toprule
\textbf{Preference}                               & \textbf{Count} & \textbf{Percentage (\%)}                 \\ \midrule
NAFNet$^\text{ModularBVQA-S}$ preferred & 895 & 87.40\% \\
NAFNet preferred & 74 & 7.23\% \\
Indistinguishable (tie) & 55& 5.37\% \\
\bottomrule
\end{tabular}
\vspace{-12pt}
\end{table}


\section{Conclusion and Discussion}
\label{sec:conclusion}
Current developments in video quality assessment (VQA) typically follow two distinct trends~\cite{zheng2024video}: 1) General-purpose methods~\cite{Wu_2023_ICCV,Wen_2024_CVPR,he2024cover,10.5555/3692070.3694286,zheng2022blind,li2021unified,zheng2022completely} provide versatile solutions suitable for diverse practical scenarios; 2) Specialized models~\cite{wang2016perceptual,tandon2021cambi,xue2023large,zheng2024faver,10438477,zheng2022no,ZHAO2020103024,10054147,10325414,10031193,MUKHERJEE2016426} offer detailed, perceptually precise assessments critical for targeted applications, particularly where subtle yet impactful artifacts significantly affect viewer experience. Our study explicitly aligns with the latter, rigorously investigating banding artifacts prevalent in compressed high-definition video content. Addressing this subtle distortion not only enhances quality prediction accuracy in targeted streaming scenarios but also complements general-purpose VQA systems as standalone banding predictors or as modular components within broader ensemble frameworks.
Specifically, we conducted a comprehensive subjective and objective study of banding artifacts arising from video compression.
We built the first-of-a-kind open-source banding VQA database to date, dubbed the LIVE-YT-Banding Database.
We benchmarked many FR and NR video quality prediction models on it, including both general-purpose and banding-specific models.
We also created a novel banding video quality paradigm by modeling banding distortions at the neural level, which we call CBAND.
Experiments conducted on the LIVE-YT-Banding database show that the CBAND models significantly outperform state-of-the-art algorithms, with orders-of-magnitude faster inference speed.
Additionally, we demonstrated the usefulness of CBAND as a supervising objective on the perceptual video debanding problem.
Furthermore, we explored the modular capability of CBAND, demonstrating substantial performance gains when integrated into existing general-purpose VQA models. Based on this finding, we recommend that researchers further explore hybrid strategies, combining general-purpose and specialized distortion-specific methods to leverage their complementary strengths.

While this work establishes a strong foundation for banding-aware VQA, several areas remain open for future exploration. 
The scale of our dataset, while carefully curated, is smaller than large-scale general VQA datasets, and expanding it with more diverse resolutions, frame rates, and compression settings could further improve generalization. 
Moreover, CBAND primarily focuses on spatial banding artifacts without explicitly modeling temporal dynamics. Integrating motion-aware features in future work could further enhance banding assessment in dynamic scenes. 
Our evaluation also highlights the potential of Mamba-based architectures for perceptual video quality tasks, suggesting further research into efficient, adaptive models for banding-aware assessment.
We envision that our work will facilitate future research efforts on banded VQA and perceptually optimized video compression, leading to higher-quality, more efficient streaming video systems.
\section*{Acknowledgment}
The authors would like to thank volunteers who participated in the subjective study. The human study was conducted under the approval of The University of Texas at Austin Institutional Review Board (IRB) under protocol 2007-11-0066. The authors also thank Eric Klassen, Dominik Millarc (MILLARC CGI), Mitch Martinez, Jan Curtis (\@Northern\_Nights (Flickr)) for generously providing the high-quality source videos.



\bibliographystyle{IEEEtran}
\bibliography{IEEEfull}

\end{document}